\documentclass[preprint,12pt]{elsarticle}

\usepackage{amssymb}
\usepackage{amsfonts}
\usepackage{amsthm,bm}
\usepackage{amsmath}
\usepackage{algorithm,algorithmicx}
\usepackage{algpseudocode}
\usepackage[hidelinks]{hyperref}
\usepackage[font=tiny]{subcaption}
\usepackage[font=small,skip=0pt]{caption}
\usepackage{array}
\usepackage{textcomp}
\usepackage{multirow}
\usepackage{url}
\usepackage{verbatim}
\usepackage{graphicx}
\usepackage{booktabs}
\usepackage{color}
\usepackage{pgf-pie}
\usepackage{xcolor}
\usepackage{thmtools, thm-restate}
\usepackage{tabularx}
\usepackage{ragged2e}
\usepackage{array}
\usepackage{amsmath}
\usepackage{caption}

\usepackage{enumitem}
\usepackage[table]{xcolor}

\usepackage{tikz}
\usepackage{tabularx}
\usepackage{enumitem}
\usetikzlibrary{positioning,fit,arrows.meta,backgrounds,calc}

\tikzset{
    >=Stealth,
    node distance = 1.2cm and 1.2cm, 
    basebox/.style={rectangle,draw,thick,fill=white,align=center,inner sep=6pt},
    statebox/.style={basebox,minimum width=1.3cm,minimum height=1.6cm},
    nnlayer/.style={rectangle,minimum width=0.4cm,minimum height=2cm,inner sep=0pt},
    modulebg/.style={rectangle,rounded corners=6pt,thick,draw,inner sep=15pt},
    moduletitle/.style={font=\bfseries\large,align=center},
    formulatxt/.style={align=center,font=\normalsize,inner sep=0pt},
    stamp/.style={red,thick,draw,rounded corners,font=\bfseries,rotate=12,align=center,inner sep=4pt}
}

\newcommand{\goalcycle}{%
\begin{tikzpicture}[
  font=\small\sffamily,
  every node/.style={
    draw,
    rounded corners,
    align=center,
    minimum height=0.7cm,
    minimum width=2.2cm
  },
  >=Stealth
]
  \node (G) at (0,1.2) {high-level valued goal\\ $G$};
  \node (st) at (0,0) {goal-relevant state \\$s_t$};
  \node (etheta) at (4.2,0.6) {emotional preference\\ $\mathbf{e}_\theta(s_t,{\color{gray}G})=\mathbf{w}_t$};
  \node (piQ) at (8.6,0.6) {strategy selection\\ $\pi_{\mathbf{Q}}(s_t,\mathbf{w}_t)$};

  \node (beh) at (8.6,-1.2) {behavior $a_t$};
  \node (snext) at (0,-1.2) {goal-relevant state\\ $s_{t+1}$};

  \draw[->] (G.east) -- (etheta.west);
  \draw[->] (st.east) -- (etheta.west);
  \draw[->] (etheta.east) -- (piQ.west);
  \draw[->] (piQ.south) -- (beh.north);
  \draw[->] (beh.west) -- (snext.east);
\end{tikzpicture}
}

\definecolor{c1} {HTML}{4472C4}
\definecolor{c2} {HTML}{ED7D31}
\definecolor{c3} {HTML}{A5A5A5}
\definecolor{c4} {HTML}{FFC000}
\definecolor{c5} {HTML}{70AD47}
\definecolor{c6} {HTML}{5B9BD5}
\definecolor{c7} {HTML}{FF6F61}
\definecolor{c8} {HTML}{C154C0}
\definecolor{c9} {HTML}{8B5A2B}
\definecolor{c10}{HTML}{2E8B57}
\definecolor{c11}{HTML}{DC143C}
\definecolor{c12}{HTML}{00CED1} 
\definecolor{c13}{HTML}{9400D3} 
\definecolor{c14}{HTML}{FF4500} 
\definecolor{c15}{HTML}{9ACD32} 
\definecolor{c16}{HTML}{8A2BE2} 
\definecolor{c17}{HTML}{D2691E} 
\definecolor{c18}{HTML}{00FA9A} 
\definecolor{c19}{HTML}{FF1493} 
\definecolor{c20}{HTML}{1E90FF} 
\definecolor{c21}{HTML}{B22222} 

\theoremstyle{plain}
\newtheorem{theorem}{Theorem}[section]
\newtheorem{lemma}[theorem]{Lemma}
\newtheorem{corollary}[theorem]{Corollary}

\theoremstyle{definition}
\newtheorem{definition}[theorem]{Definition}

\theoremstyle{remark}
\newtheorem{remark}{Remark}

\journal{}

\begin{document}

\begin{frontmatter}



\title{Emotional Preferences as Goal-Priority Regulation}


\author{Shiqi Liu} 
\ead{shiqi.liu647@foxmail.com}
\author{Yihua Tan\corref{cor1}}
\ead{yhtan@hust.edu.cn}
\author{Hu Fu}
\ead{fuhu@hust.edu.cn}
\author{Guanyu Qi}
\ead{gyqi@hust.edu.cn}
\affiliation{organization={School of Artificial Intelligence and Automation, Huazhong University of Science and Technology},
            addressline={Luoyu Street},
            city={Wuhan},
            postcode={430074},
            state={Hubei},
            country={China}}
\cortext[cor1]{Corresponding Author}

\begin{abstract}
A core question in autonomous decision-making for artificial agents is whether the relative priorities of competing lower-level objectives can be determined by emotional preferences autonomously generated by higher-level goals, rather than being externally prespecified. Under changing external environments and evolving internal states, emotions play an important functional role in regulating the relative priorities of competing goals. Inspired by the goal-directed theory of emotion, this paper studies how such preference regulation can be computationally realized through reinforcement learning. To this end, we first propose a conception of emergent emotional preference: a high-level goal autonomously induces state-dependent preferences over competing lower-level objectives. This conception is built upon a framework consisting of a pretrained multi-objective reinforcement learning (MORL) inner controller and an outer preference generator. The inner controller provides a repertoire of preference-conditioned goal-directed behaviors, while the outer preference generator learns a mapping from the current state to objective preferences through reinforcement learning on a high-level goal. We operationalize emotional preference as a state-dependent regulation of relative goal priorities that emerges through optimization, rather than as predefined emotion labels or a complete model of human emotions. Furthermore, we characterize the policy space induced by preference regulation and derive an upper bound on the optimality gap in terms of the representation error of the inner behavioral repertoire. We show that the gap vanishes when the optimal policy can be represented by the available preference-conditioned policies, meaning that the policy space contains the optimal policy at the representational level. Experiments in self-constructed basic and advanced multi-objective exploration environments (Grid Fruit Tree Battery Exploration) show that the learned preference function exhibits contextual priority switching, graded trade-offs, and temporal persistence, and outperforms the evaluated fixed-preference and handcrafted-preference strategies. These results reveal a computational mechanism by which high-level goal optimization can induce state-dependent emotional preference and dynamically reorganize competing goal-directed behaviors.
\end{abstract}

%


\end{frontmatter}

\section{Introduction}

Emotion plays a central role in human goal-directed behavior by regulating which goals and action strategies receive priority under changing internal and environmental conditions. When multiple goals compete, an individual does not necessarily maintain a globally fixed trade-off among them. Instead, the relative priority of behavioral goals may change with the current situation, physiological needs, and long-term objectives. This perspective is particularly relevant to computational theories that view emotional episodes as components of goal-directed processes rather than as isolated stimulus-response mechanisms. In the goal-directed theory (GDT) proposed by Moors\cite{moors2026emotions}, behavior and affect arise from interacting and competing goal-directed cycles, each involving discrepancy detection and the selection of strategies or behaviors that can reduce the relevant discrepancy. Emotional processes are therefore closely related to the regulation of behavior under highly valued goals.

Some psychological research points out that emotions dynamically generate behavioral tendencies toward lower-level subgoals (such as acquiring energy, avoiding threats, and accumulating achievements) based on an individual's high-level goals (such as survival, autonomy, control, connectedness, happiness, and identity)~\cite{moors2022demystifying,moors2019demystifying,pineda2026some}. This perspective raises an important question for artificial agents: \textbf{can the relative priority of competing objectives be learned autonomously from a high-level goal, rather than being specified externally in advance?} The question is particularly natural in multi-objective reinforcement learning (MORL), where an agent typically receives a preference vector $\mathbf{w}$ that specifies the relative importance of different reward dimensions. Existing MORL methods can learn policies that generalize across different preferences, but the preference itself is usually treated as an input supplied by a user, fixed globally, or generated according to an externally defined dynamic process. Thus, these methods primarily address the question of \textit{how to act under a given preference}, while leaving open the complementary question of \textit{when should one objective become more important or take priority over another.}

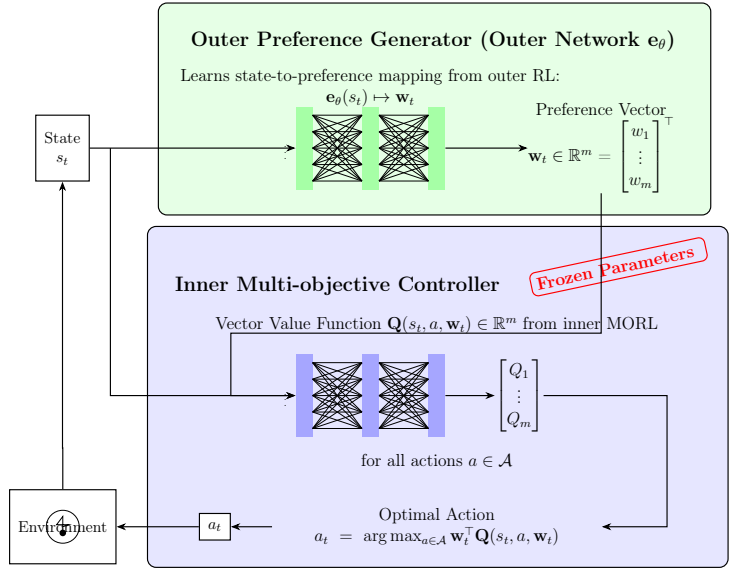
\begin{figure}[htb]
  \centering
  \resizebox{0.7\textwidth}{!}{%
\begin{tikzpicture}[font=\small]



\node[statebox,anchor=east] (state_outer) at (0, 8) {State \\ $s_t$ };

\node[nnlayer,fill=green!35,right= 5cm of state_outer] (outer_nn1) {};
\node[nnlayer,fill=green!35,right=of outer_nn1] (outer_nn2) {};
\node[nnlayer,fill=green!35,right=of outer_nn2] (outer_nn3) {};

\node[moduletitle,above=of outer_nn3] (outer_title)
{Outer Preference Generator (Outer Network $\mathbf{e}_\theta$) };

\foreach \y in {-0.8,-0.4,0,0.4,0.8} {
    \foreach \yy in {-0.8,-0.4,0,0.4,0.8} {
        \draw[thin] ($(outer_nn1.east)+(0,\y)$) -- ($(outer_nn2.west)+(0,\yy)$);
        \draw[thin] ($(outer_nn2.east)+(0,\y)$) -- ($(outer_nn3.west)+(0,\yy)$);
    }
}
\draw[->,thick] (state_outer.east) -- (outer_nn1.west);
\node[left=2pt of outer_nn1.west,font=\tiny] {$\vdots$};

\node[formulatxt,anchor=west,above= 0cm of outer_nn2] (outer_explain) {
    Learns state-to-preference mapping from outer RL: \\
    $\mathbf{e}_\theta(s_t) \mapsto \mathbf{w}_t$
};

\node[formulatxt,anchor=west,right=2cm of outer_nn3] (w_vec) {
    Preference Vector \\
    $\mathbf{w}_t \in \mathbb{R}^m$ = $\begin{bmatrix} w_1 \\ \vdots \\ w_m \end{bmatrix}^\top$ 
};


\draw[->,thick] (outer_nn3.east) -- (w_vec.west);

\begin{scope}[on background layer]
\node[modulebg,fill=green!10,fit=(outer_title) (outer_nn1) (outer_explain) (w_vec)] (outer_module) {};
\end{scope}


\node[nnlayer,fill=blue!35,below=4cm of outer_nn1] (inner_nn1) {};
\node[nnlayer,fill=blue!35,right=of inner_nn1] (inner_nn2) {};
\node[nnlayer,fill=blue!35,right=of inner_nn2] (inner_nn3) {};
\foreach \y in {-0.8,-0.4,0,0.4,0.8} {
    \foreach \yy in {-0.8,-0.4,0,0.4,0.8} {
        \draw[thin] ($(inner_nn1.east)+(0,\y)$) -- ($(inner_nn2.west)+(0,\yy)$);
        \draw[thin] ($(inner_nn2.east)+(0,\y)$) -- ($(inner_nn3.west)+(0,\yy)$);
    }
}

\node[moduletitle,above=1.3cm of inner_nn2, xshift=-0.8cm] (inner_title)   {Inner Multi-objective Controller};

\node[stamp,anchor=west, right=0.6cm of inner_title] (frozen_stamp)   {Frozen Parameters };

\node[left=2pt of inner_nn1.west,font=\tiny] {$\vdots$};
\node[right=2pt of inner_nn2.east,font=\tiny] {$\vdots$};


\node[formulatxt,above=0.5cm of inner_nn3] (vvf_title) 
{Vector Value Function $\mathbf{Q}(s_t,a,\mathbf{w}_t) \in \mathbb{R}^m$ from inner MORL};
\node[formulatxt,below=0.4cm of inner_nn3] (vvf_note) {for all actions $a \in \mathcal{A}$};

\node[formulatxt,right=1.2cm of inner_nn3] (Q_vec) {$\begin{bmatrix} Q_1 \\ \vdots \\ Q_m \end{bmatrix}$};

\draw[->,thick] (inner_nn3.east) -- (Q_vec.west);
\node[formulatxt,below=1cm of vvf_note,text width=8cm] (action_select) {
    Optimal Action \\
    $a_t = \arg\max_{a\in\mathcal{A}} \mathbf{w}_t^\top \mathbf{Q}(s_t,a,\mathbf{w}_t)$ 
};

\node[basebox,left=1cm of action_select] (a_out) {$a_t$};
\node[basebox,left=2cm of a_out,minimum width=1.5cm,minimum height=1.8cm] (env) {Environment};
\draw (env.center) circle (0.4cm);
\draw (env.center) -- ++(0,0.25) -- ++(-0.15,-0.2) -- ++(0.3,0) -- cycle;
\fill (env.center) ++(0,-0.15) circle (0.08cm);

\draw[->,thick] (action_select.west) -- (a_out.east);
\draw[->,thick] (a_out.west) -- (env.east);
\draw[->,thick] (Q_vec.east) -- ++(3,0)  |-(action_select.east);
\begin{scope}[on background layer]
\node[modulebg,fill=blue!10,fit=(inner_title) (frozen_stamp)(inner_nn1)  (Q_vec) (action_select) (a_out)  (vvf_note)] (inner_module) {};
\end{scope}


\draw[->,thick] (state_outer.east) ++(0.5,0) |- (inner_nn1.west);
\draw[->,thick] (env.north) -- ++(0,1)-| (state_outer.south);
\draw[->,thick] (w_vec.south) -- ++(0,-3.4) -- ++(-9,0) |-(inner_nn1.west);
\end{tikzpicture}
  }
  \caption{The overall framework for emergent emotional preference, comprising inner MORL and outer emotional preference learning.}
  \label{fig:emergent emotional preference}
\end{figure}

We study this problem as \textbf{autonomous preference generation}. Our key idea is to formulate preference generation as an outer reinforcement learning problem operating on top of a pretrained preference-conditioned inner MORL controller. The inner controller learns a family of behaviors under different objective trade-offs, while a separate outer network learns a state-dependent mapping $ \mathbf{e}_{\theta}: \mathcal{S} \to \Delta^{m-1} $ where $\mathcal{S}$ denotes the state space and $\Delta^{m-1}$ denotes the probability simplex over $m$ objectives. At state $s_t$, the outer network produces the preference $\mathbf{w}_t = \mathbf{e}_\theta(s_t),$ and the frozen inner MORL controller $\mathbf{Q}$ executes an action according to this preference, $a_t = \arg\max_{a\in\mathcal{A}} \mathbf{w}_t^\top \mathbf{Q}(s_t,a,\mathbf{w}_t)$. (see Fig.~\ref{fig:emergent emotional preference}). The outer network is optimized solely with respect to a high-level task reward, such as long-term survival. Hence, the preference function is not manually programmed as a rule such as “when energy is low, prioritize energy.” Instead, the state-to-preference mapping is formed through task optimization.

We call the resulting state-dependent mechanism an \textbf{emergent emotional preference}. Importantly, we use this term in an operational and computational sense. We do not claim that the learned network constitutes a complete model of human emotion, nor that subjective affect, bodily responses, or phenomenological experience emerge from the proposed architecture. Rather, we define an emotional preference as a learned state-dependent regulation of the relative priority of competing goal-directed objectives. This interpretation is grounded in the goal-directed theory of emotion: the preference vector can be understood as a computational variable that regulates the relative valuation of competing strategies within a goal-directed cycle. Under this interpretation, the outer high-level objective specifies the longer-term valued goal, the current state provides information about goal-relevant discrepancies, and the learned preference determines which lower-level objective and corresponding strategy should receive priority.

This formulation provides a computational bridge between MORL and the goal-directed perspective on emotion. The inner MORL controller represents alternative goal-directed strategies, while the outer preference generator regulates which strategy is prioritized in the current context. For example, in our exploration environment, long-term survival may require prioritizing energy acquisition initially, whereas achievement-oriented behavior may become advantageous when energy can no longer be replenished. The resulting preference shifts are therefore not externally prescribed; they emerge from optimization of the high-level goal.

The proposed framework also gives rise to an important theoretical question: \textbf{how close can such an outer preference-selection mechanism come to an unrestricted optimal policy?} Because the outer controller can only select behaviors available in the pretrained MORL policy family, its feasible policy space is generally smaller than the space of all policies. We therefore characterize the resulting optimality gap in terms of the representation capacity of the inner policy set. In particular, we derive an upper bound on the gap between the optimal unrestricted policy and the best policy representable through preference-conditioned inner policies, showing that the gap is governed by the representation error of the inner policy family and is amplified by long-horizon discounted optimization.

We evaluate the framework in synthetic multi-objective exploration environments with competing achievement, energy, and safety objectives. Across the experiments, the learned preference generator produces state-dependent and temporally persistent preference patterns and outperforms the tested fixed-preference policies in the survival task. More importantly, the preference trajectories reveal interpretable transitions between competing objectives, concretely demonstrating how a high-level objective can context-dependently regulate the priorities of lower-level objectives.

Our work makes the following contributions:

\begin{itemize}
\item \textbf{Autonomous emotional preference generation.} We formulate the generation of state-dependent preferences in MORL as an outer reinforcement learning problem. Instead of treating objective weights as externally specified inputs, our framework learns a mapping from states to objective preferences under a high-level survival objective.

\item \textbf{A computational formulation of emergent emotional preference.} We provide an operational definition of emotional preference as state-dependent regulation of competing goal priorities and relate this mechanism explicitly to the goal-directed theory of emotion.

\item \textbf{A decoupled architecture for preference regulation and skill execution.} We decouple when or what to prefer from how to act: a frozen preference-conditioned MORL controller provides a repertoire of lower-level goal-directed behaviors, while an outer preference generator learns to regulate their relative priorities according to the current state and high-level objective.

\item \textbf{Theoretical characterization of representation-limited optimality.} We characterize the policy class induced by an outer preference generator over a pretrained MORL policy family in a simplified discrete setting, and derive an optimality-gap bound in terms of policy representation error. The analysis identifies the representational capacity of the pretrained policy repertoire as a fundamental determinant of the performance achievable by outer preference learning.

\item \textbf{Empirical analysis of contextual preference dynamics.}  Across synthetic multi-objective exploration environments, we show that outer optimization produces state-dependent preference switching,  including transitions among energy, achievement, and safety priorities, and temporally persistent preference patterns. These preferences lead the agent to activate different objective-conditioned behaviors in different contexts, while outperforming the tested fixed-preference policies on the survival task.

\item \textbf{A computational bridge between emotion and MORL.} The framework connects the goal-directed psychological theory of emotion with a concrete reinforcement-learning mechanism for selecting among competing objective-conditioned behaviors.
\end{itemize}

\section{Related Work}

\subsection{Goal-Directed Theories of Emotion and Computational Preference Regulation}
Classical computational approaches to emotion have often attempted to reproduce specific emotional states, appraisal dimensions, or affective responses. Such approaches differ considerably in their assumptions about how emotion relates to decision-making. Some models treat emotion as an appraisal process that evaluates environmental events, whereas others introduce emotion as an auxiliary signal that modulates motivation or action selection. A common challenge across these approaches is to specify how an internal affective state changes the relative priority of competing behavioral goals.

The \textbf{goal-directed theory (GDT) of emotion} provides a particularly relevant perspective for this problem. Moors characterizes behavior and affect in terms of goal-directed cycles involving two central stages: discrepancy detection, in which a current, anticipated, or imagined state is compared with a desired state, and strategy or behavior selection, in which an option capable of reducing the discrepancy is selected\cite{moors2022demystifying,moors2026emotions}. The theory further proposes that multiple goal-directed cycles may operate concurrently and compete for control of behavior.\cite{moors2022demystifying,moors2026emotions} Under this view, what are ordinarily described as emotional episodes are closely related to the same goal-directed machinery involved in instrumental behavior, with emotional cycles tending to involve goals of particularly high value.\cite{moors2022demystifying,moors2026emotions}

This perspective is important for computational modeling because it does not require an emotion mechanism to be an isolated, domain-specific module. Instead, emotion can be understood through the functional role that a process plays within a goal-directed system. In particular, a process that changes the relative priority of competing strategies according to goal relevance can be interpreted as part of an emotional goal-directed cycle. Moors therefore emphasizes a continuum between instrumental and emotional goal-directed processes rather than a strict mechanistic separation between them.

Our framework adopts this functional interpretation. We do not attempt to model the full set of phenomena associated with human emotion. Instead, we focus on one computationally explicit component: the regulation of relative priority among competing goal-directed objectives. Let $\mathbf{r}(s,a) = (r_1(s,a), r_2(s,a), \dots, r_m(s,a))^\top$ denote a vector of lower-level objectives, and let $\mathbf{w}_t \in \Delta^{m-1}$ denote their current relative priorities. The resulting scalarized utility is $U(s_t,a_t,\mathbf{w}_t) = \mathbf{w}_t^\top \mathbf{Q}(s_t,a_t,\mathbf{w}_t).$ In our framework,  $\mathbf{w}_t$ is not fixed or externally supplied. Instead, $\mathbf{w}_t = \mathbf{e}_\theta(s_t),$ is optimized with respect to a high-level outer goal. Consequently, the outer network learns to regulate which lower-level objective receives behavioral priority as a function of the current state.

This mechanism can be interpreted as a computational realization of the preference-regulation component of a goal-directed emotional cycle. The correspondence is functional rather than representational: $\mathbf{w}_t$ is not intended to correspond to a phenomenological emotion label such as fear or happiness. Rather, it represents the relative priority assigned to competing goal-directed strategies. This distinction is important because our objective is to investigate how emotional preferences can emerge as an optimization process, rather than to claim that a complete human-like emotional state has emerged.

Recent discussions of GDT also emphasize that the same underlying goal-directed machinery can support both instrumental and emotional behavior. This observation is particularly compatible with our architecture: the inner MORL controller provides the general decision-making machinery, while the outer learning process determines when different objective dimensions should be prioritized. Therefore, rather than introducing a dedicated, hard-coded emotion module, our model asks whether the relative priorities of competing low-level objectives can be determined by emotional preferences autonomously generated from the high-level goal, rather than being externally specified by humans beforehand.

This interpretation also distinguishes our work from computational approaches in which emotion is merely added as an intrinsic reward. In those settings, an affective signal typically modifies the scalar reward optimized by the agent. In contrast, our emotional preference directly changes the relative priority structure of multiple objectives, thereby modulating which pretrained goal-directed strategy is selected. The distinction is therefore between learning an additional reward signal and learning a state-dependent mechanism for regulating competing goal priorities.

\begin{table}[htb]
\centering
\caption{Correspondence between the Goal-Directed Theory of Emotion and Our Computational Framework}
\label{tab:emotion-mapping}
\footnotesize
\renewcommand{\arraystretch}{1.35}
\setlength{\tabcolsep}{8pt}
\begin{tabularx}{\textwidth}{@{}>{\RaggedRight\arraybackslash}p{0.30\textwidth}>{\RaggedRight\arraybackslash}X@{}}
\toprule
\textbf{GDT construct / mechanism} &
\textbf{Computational interpretation in our framework} \\
\midrule

Valued goal / Goal at stake
& The high-level task objective $G$, e.g., long-term survival \\

Actual or anticipated stimulus / state
& The current or anticipated agent--environment state $s_t$ \\

Stimulus--goal discrepancy
& A mismatch between the current situation and states favorable to the high-level goal, e.g., low energy or elevated safety risk \\

Discrepancy-reduction strategies
& Preference-conditioned MORL policies representing alternative lower-level goal-directed behaviors \\

Strategy selection based on expected utility
& Preference-conditioned strategy execution through the inner MORL controller \\

Goal value / Goal competition
& The relative priority among competing lower-level objectives, represented by the preference vector $\mathbf{w}_t \in \Delta^{m-1}$ \\

Action selection by expected utility
& $a_t = \arg\max_{a} \; \mathbf{w}_t^\top Q(s_t, a, \mathbf{w}_t)$ \\

Goal-directed cycle
& High-level valued goal + goal-relevant state $\rightarrow$ preference regulation $\rightarrow$ strategy selection $\rightarrow$ behavior/outcome $\rightarrow$ feedback \\

\bottomrule
\end{tabularx}
\end{table}

\subsection{Other Computational Models of Emotion and Decision-Making}

\textbf{Evolutionary Emotion Theory\cite[Evolutionary Theories]{moors2022demystifying}:} Emotions are adaptive affective programs shaped by long-term natural selection in species. Their core function is to help individuals rapidly cope with survival-related environmental challenges. They possess innate, hardwired processing pathways, in which specific stimuli can automatically trigger corresponding response tendencies.
In this architecture, the outer reinforcement learning optimizes the outer network with the goal of maximizing survival duration, essentially simulating at the computational level the process by which natural selection shapes emotion mechanisms. This aligns with the core claim that emotions are products of survival adaptation. The design in which the outer network receives state inputs, outputs stable preferences, and the preference-to-action policy is frozen corresponds to the processing logic of ``stimulus-triggered $\to$ automatic response'' in affective programs, reflecting the innate and low-flexibility characteristics of the emotion pathway.

\textbf{Network Theory of Emotion\cite[Network Theories]{moors2022demystifying}:} Based on a connectionist perspective, network theory of emotion posits that emotions are not unitary mental entities but dynamic networks composed of multiple components linked through association. The connection strengths between network nodes can be continuously shaped by individual experience, and activation spreads along the connection pathways, supporting the learning, generalization, and dynamic change of emotions.
The outer network in this architecture is implemented as a neural network, directly corresponding to the core assumptions of this theory: the process whereby environmental states activate the network and generate behavioral preferences through weighted computation mirrors the spreading activation mechanism in associative networks; the iterative optimization of network weights by the outer reinforcement learning corresponds to the long-term shaping of emotional connections by individual experience, corroborating the plasticity of the outer network.

\textbf{Stimulus Evaluation Theory\cite[Stimulus Evaluation Theories]{moors2022demystifying}:} Cognitive evaluation of environmental stimuli is the core process of emotion generation. The individual’s appraisal of the meaning of events (e.g., threat, controllability) determines the type, intensity, and subsequent behavioral tendencies of the emotion, with rapid automatic evaluation providing the basis for prioritized emotional responses.
In this architecture, the processing of the outer neural network is essentially an automatic stimulus evaluation: the network receives the current environmental state, completes the appraisal of survival significance through its internal weights, and outputs corresponding behavioral preferences, consistent with the logic that ``evaluation drives emotion and determines behavioral tendencies.'' The preferences output by the network correspond to action readiness states in the theory, and the subsequent frozen policy generates concrete actions based on these preferences, completing the full pathway from cognitive evaluation to overt behavior.

In recent years, computational models have begun to explore the interaction between emotion and decision-making. ~\cite{kirtay2016sequential} proposed a sequential decision-making framework based on emotion emergence, in which a robot minimizes neural processing costs through RL, resulting in externally observable ``liking'' behaviors that are interpretable to observers---such emotional expressions emerge purely from an internal cost-minimization principle. 

\cite{zhang2024modeling} proposed a computational model integrating the component process model with RL. By formalizing four appraisal checks---suddenness, goal relevance, goal conduciveness, and power---the model enables an agent to dynamically predict emotional experiences in goal-directed tasks, thereby establishing a formal link between reward processing and cognitive appraisal. It should be noted that, although this model can predict emotion intensity, the emotion itself does not influence the agent's behavioral choices or policy updates. Its emotion appraisal relies on a predefined MDP structure and manually designed transition probabilities and rewards. These models primarily focus on appraisal and emotion prediction, whereas our framework explicitly treats relative objective priority as a learned control variable.

\subsection{MORL}

MORL aims to handle decision-making problems with multiple competing objectives, where the core challenge lies in how to trade off conflicts among different goals. Existing methods fall mainly into two categories: single-policy methods and multi-policy methods~\cite{roijers2013survey}. A similar taxonomy includes decision-prior and optimization-prior methods~\cite{peitz2025multi}. We present them below following the single-policy and multi-policy classification.

Single-policy methods~\cite{roijers2013survey} apply to scenarios with known weights, i.e., before planning or learning begins, the decision-maker has already specified the importance weights of each objective. The goal of such methods is to directly find an optimal policy that maximizes the scalarized expected return under the given weights. When the scalarization function is linear, a MOMDP can be equivalently transformed into a single-objective MDP, allowing standard RL or dynamic programming algorithms such as Q-learning and SARSA to be applied directly.

Multi-policy methods~\cite{roijers2013survey} apply to scenarios with unknown weights or decision support, where weights are unavailable during planning or learning, or user preferences are difficult to quantify. The goal of such methods is to return a coverage set, ensuring that for any possible weight vector there exists at least one policy in the set that is optimal. For linear scalarization, it suffices to compute the convex coverage set, i.e., the convex hull formed by deterministic stationary policies. The Envelope Q-Learning proposed by~\cite{yang2019generalized} introduces preferences into the value function and updates through a convex envelope to learn optimal policies under all possible preferences, achieving few-shot adaptation capability. \cite{hu2024pa2d} utilizes Pareto ascent direction to select scalarization weights and selectively optimizes multiple policies under an evolutionary framework to approximate the Pareto front. \cite{shu2024learning} proposed using a single hypernetwork to learn a continuous representation of the Pareto set, which can directly generate policy networks for different user preferences, significantly improving resource efficiency.

However, the preferences considered by these methods are usually globally fixed and cannot change as the environmental state changes. 

MORL determines how to execute a given preference; our outer process learns when and what to prefer. 

$$ s \rightarrow \underbrace{\mathbf{w}}_{\text{preference generation}} \rightarrow \underbrace{\pi_\mathbf{w}}_{\text{preference execution}} \quad $$

$$\boxed{\text{preference generation}\neq\text{preference execution}.}$$

\subsection{Dynamic Preferences}

Traditional MORL assumes that preferences are static and known a priori, or specified online by the user. Yet this assumption often fails in real-world scenarios---when hungry, a human prioritizes finding food; when in danger, safety takes precedence. Such contextualized preference changes are precisely an expression of emotion at work.

In recent years, researchers have begun to pay attention to the problem of preferences that change dynamically with the environment. \cite{buet2023robust} proposed a robust MORL method for dynamic preferences, achieving joint exploration of states and preferences by constructing an augmented state space composed of states and preferences. They treat static preferences as a special case of dynamic preferences, proving the generality of the framework. However, the preference dynamics are still externally specified rather than internally generated by the agent.

Meta-MORL methods, represented by Preference Controllable RL (PCRL)~\cite{yang2025preference}, train a meta-policy that can accurately execute various given preferences through goal-aligned preference regularization. However, the implicit premise of this ``instruction execution'' paradigm is that preferences are defined externally and input into the system; the agent itself lacks the ability to autonomously generate goal trade-offs in the absence of external instructions. It is at this point that our work diverges from existing Meta-MORL paths: we are concerned not with ``how to execute given preferences'', but with ``how preferences themselves emerge from survival tasks''.

\cite{cao2026learning} proposed using variational inference to dynamically infer and adjust current preferences based on environmental changes. However, the objectives and priors guiding preference generation in this method are not clearly defined, and the absence of publicly available code makes replication difficult.

It should be noted that the innovative focus of this framework lies in the preference generation(priority regulation) mechanism, rather than the preference execution mechanism (the inner multi-objective policy). The inner module can adopt any MORL algorithm with preference generalization capability (e.g., Envelope Q-Learning~\cite{yang2019generalized} or the meta-policy of PCRL~\cite{yang2025preference}), because we argue that when the outer task requires long-term survival, the outer network will give rise to contextualized preference-generation behavior regardless of which preference-conditioned policy is used internally. In particular, we choose Envelope Q-Learning as the inner algorithm.

\subsection{Emotion as Intrinsic Motivation in RL}

The close connection between emotion and motivation has led researchers to explore the possibility of using emotion as an intrinsic reward signal. Studies have shown that emotional factors such as curiosity, happiness, and sense of control can act as intrinsic motivation, driving agents to explore unknown states and adjust learning preferences and behavioral patterns~\cite{lu2016using}. Such methods accelerate the learning process of classical RL by introducing emotion as an intrinsic reward, achieving good results in scenarios such as maze navigation.

\cite{sequeira2015emergence} used genetic programming to evolve multiple appraisal signals (e.g., advantage, novelty, predictability) within intrinsically motivated RL agents. These signals significantly improved performance across multiple tasks and showed correspondences with psychological emotion appraisal dimensions such as valence, novelty, and coping potential. A notable shortcoming of this method is that it does not consider the complexity of decision-making in multi-objective contexts. When extended to MORL problems, this method may struggle to generate dynamic preferences with context adaptability, thus limiting its applicability in more complex and realistic tasks.

In the context of responsible RL, \cite{keerthana2025towards} proposed an emotional intelligence and responsible RL framework that integrates emotion and contextual understanding into sequential decision-making. The framework formalizes the personalization problem as a constrained Markov decision process, using a multi-objective reward function to balance short-term behavioral engagement and long-term user well-being. Although their multi-objective weights might be obtained through meta-learning optimization, they remain globally fixed.

\subsection{Skill-Based and Hierarchical RL}
Skill-based RL aims to discover a set of distinguishable latent skills, typically through unsupervised mutual information maximization or diversity rewards, using a policy ensemble trained with uniform latent variables~\cite{bai2024constrained,liu2025balancing,cho2025amped}. Formally, such methods learn a conditional policy $\pi(a|s,z)$ and a discriminator $q(z|s)$, and force different $z$ to correspond to different visited state distributions by maximizing \(\max I(Z;S)\)~\cite{eysenbach2018diversity} or similar objectives.  However, the learned skills lack explicit objective semantics---they are defined purely by state coverage distinctiveness, not by objective-relevant trade-offs. In contrast, our inner policies are conditioned on preference weights \(\mathbf{w}\) that directly encode interpretable multi-objective priority trade-offs, providing clear physical meaning.

Hierarchical RL (HRL) decomposes long-horizon tasks into subtasks, with higher-level policies selecting options or sub-policies and lower-level policies executing primitive actions~\cite{pateria2021hierarchical,sutton1999between,dietterich2000hierarchical}. Classical frameworks include the Options framework~\cite{sutton1999between}, where an option $\omega$ is defined by an initiation set $I_\omega$, a policy $\pi_\omega$, and a termination condition $\beta_\omega$. MAXQ~\cite{dietterich2000hierarchical} decomposes the value function into sub-MDP components. The Option-Critic architecture~\cite{bacon2017option} learns options end-to-end via policy gradients.

Despite these advances, existing HRL methods differ from our framework in several crucial aspects. First, the high-level policy in HRL selects subgoals or options that directly influence low-level actions, whereas our outer network outputs a preference vector $\mathbf{w}$ that re-balances the multiple objectives, leaving the inner policy to execute actions under that preference. This provides a built-in interpretability: $\mathbf{w}$ directly indicates the current emphasis on each objective. Second, most HRL methods require joint or alternating training of high- and low-level policies, which often suffers from local optima and non-stationarity. Our approach adopts a two-stage decoupled training.  Third, while options or subgoals are typically black-box policies or state targets, our preference vector operates in the simplex of objective weights, offering a lightweight and interpretable modulation mechanism. Fourth, our theoretical analysis explicitly characterizes when and why the emergent emotional preferences can approach the performance of an ideal policy.

\section{Method}\label{sec:method}
\subsection{Problem Formulation: From High-Level Goals to Emotional Preferences}
We consider a hierarchical multi-objective decision-making problem in which an agent must pursue a high-level goal while simultaneously balancing multiple lower-level objectives. The key question is not only how to act under a given preference, but also \textbf{how the relative priority of competing objectives should be determined as the situation changes}.

Let
$$\mathcal{M}_{in} = \langle\mathcal{S}, \mathcal{A}, P, \mathbf{r}, \Omega, f_{\mathbf{\Omega}}, \gamma_{in}\rangle$$
denote an inner multi-objective Markov decision process (MOMDP), where
$\mathbf{r}(s,a) = (r_1(s,a), r_2(s,a), \dots, r_m(s,a))^\top$ is the vector-valued reward and
$\Omega = \Delta^{m-1} = \left\{ \mathbf{w} \in \mathbb{R}^m \mid w_i \ge 0, \sum_{i=1}^m w_i = 1 \right\}$
is the preference space. For a given preference $\mathbf{w}\in \Omega$, we use linear scalarization, $f_{\mathbf{w}}(\mathbf{r}) = \mathbf{w}^\top \mathbf{r}$.

The inner MORL problem therefore learns a family of policies indexed by preferences. Rather than fixing $\mathbf{w}$ globally, our framework introduces an outer decision process that learns \textbf{when and how the relative priority among the lower-level objectives should change}.

\subsubsection{High-level goal and lower-level objectives}

Let \(G\) denote a high-level task objective, with scalar reward \(r_G(s,a)\).
In the experiments of this paper, \(G\) corresponds to long-term survival and \(r_G(s,a)=r_{\mathrm{surv}}(s)\).

The high-level goal is not itself one of the lower-level MORL objectives. Instead, it provides the long-horizon criterion according to which the agent learns to regulate the priorities among the lower-level objectives.

At time \(t\), the agent observes state \(s_t\) and generates a preference
\[
\mathbf{w}_t = \mathbf{e}_{\theta}(s_t), \quad \text{where } \mathbf{e}_{\theta}: \mathcal{S} \to \Omega.
\]
This preference determines the relative priority assigned to the competing lower-level objectives. The inner MORL controller then selects an action according to this preference,
\[
a_t = \pi_{\mathbf{Q}}(s_t, \mathbf{w}_t) = \arg\max_{a \in A} \mathbf{w}_t^\top \mathbf{Q}(s_t, a, \mathbf{w}_t).
\]
The resulting transition is
\[
s_{t+1} \sim P(\cdot \mid s_t, a_t).
\]
Hence, the complete decision cycle can be written as
\[
s_t \to \mathbf{w}_t = \mathbf{e}_{\theta}(s_t) \to a_t = \pi_\mathbf{Q}(s_t, \mathbf{w}_t) \to s_{t+1}.
\]
The novelty of this formulation lies in making the preference itself a learned state-dependent variable rather than an externally prescribed constant.

\subsection{Emotional Preference as Goal-Directed Priority Regulation}
We use the term \textbf{emotional preference} in an operational computational sense. It does not denote a complete model of human emotion, nor does it attempt to reproduce subjective experience or all physiological and cognitive components associated with human affect. Instead, we focus on a specific functional role of emotion: regulating the relative priority of competing goal-directed objectives under a high-value goal.

\begin{definition}[Emotional Preference]
Given a high-level goal $G$ and a set of competing lower-level objectives
$
 \{r_1, \dots, r_m\},
$
an emotional preference is a state-dependent function
$
\mathbf{e}_{\theta}: \mathcal{S} \to \Delta^{m-1}
$
such that
$
\mathbf{w}_t = \mathbf{e}_{\theta}(s_t)
$
determines the relative behavioral priority assigned to the lower-level objectives at state $s_t$.

The preference function is called \textbf{emergent} when $\mathbf{e}_{\theta}$ is not explicitly specified by a hand-crafted state-to-preference rule, but is instead acquired through optimization of the high-level goal:
\[
\theta^{\star} = \arg\max_{\theta} \mathbb{E}\left[ \sum_{t=0}^{\infty} \gamma_{\text{out}}^t \, r_G(s_t, a_t) \right].
\]
Thus, in this paper, \textit{emergence} refers to the endogenous formation of a state-dependent preference mapping through goal-directed optimization.
\end{definition}

\subsubsection{Relationship to the Goal-Directed Theory of Emotion}

\begin{figure}[htb]
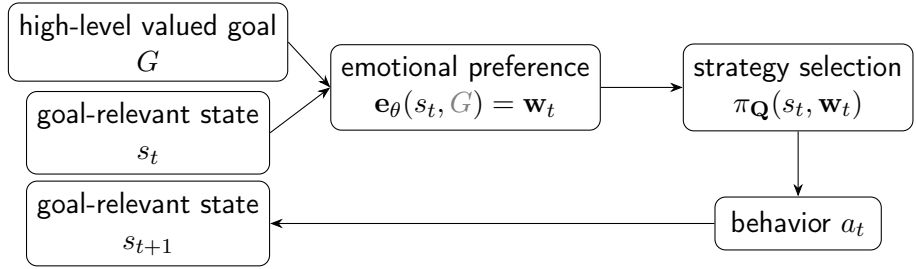

  \centering
  \goalcycle
  \caption{The goal-directed computational cycle: a high-level valued goal and a goal-relevant state drive emotional preference regulation, which selects a strategy that produces behavior. $\mathbf{e}_\theta(s_t,{\color{gray}G})$ denotes that $\mathbf{e}_\theta$ is indeed goal-relevant.}
  \label{fig:goalcycle}
\end{figure}

This formulation provides a computational interpretation of the preference-regulation component of the goal-directed theory of emotion. Under this perspective, goal-directed processes involve the identification of goal-relevant discrepancies (It is implicitly reflected in emotional preferences.) and the selection of strategies or behaviors that can reduce such discrepancies, while multiple goal-directed processes can compete for behavioral control. Our framework implements a corresponding computational cycle (see Figure~\ref{fig:goalcycle}).

The high-level goal $G$ specifies the long-term objective of the outer process. The current state $s_t$ provides the information required to determine which lower-level objectives are currently relevant. The emotional preference $\mathbf{w}_t$ regulates their relative priority, and the inner MORL controller realizes the resulting strategy through action selection.

Under this interpretation, $\mathbf{w}_t$ should not be understood as a discrete emotion label such as fear, happiness, or anger. Rather, it represents the \textbf{current priority structure among competing goal-directed objectives}. For example, a preference concentrated on the energy objective corresponds computationally to a state in which energy acquisition is given greater priority, whereas a preference concentrated on achievement corresponds to a state in which achievement-oriented behavior receives greater priority.

The important distinction is therefore between the high-level goal and the state-dependent emotional preference:$$\underset{\text{why}}{G} \neq \underset{\text{what to prioritize now}}{\mathbf{w}_t}.$$

The outer optimization specifies why the preference regulation is learned, while the state-dependent preference specifies how competing lower-level objectives are prioritized in the current context.

\subsection{Inner Multi-Objective Controller}
The inner module provides the repertoire of goal-directed behaviors from which the outer preference generator can select.

\subsubsection{Multi-Objective Value Function}
Given a policy $\pi: \mathcal{S} \to \mathcal{A}$, its corresponding state-action expected cumulative discounted return vector is defined as:
$$\mathbf{Q}^{\pi}(s,a) = \mathbb{E}_{\pi}\!\left[ \sum_{i=0}^{\infty} \gamma_{in}^i \mathbf{r}(s_{t+i}, a_{t+i}) \;\middle|\; s_t=s, a_t=a \right] = \big( Q_1^{\pi}(s,a), \dots, Q_m^{\pi}(s,a) \big)^\top$$
This paper adopts the multi-objective $\mathbf{Q}$-value function $\mathbf{Q}(s,a,\mathbf{w})$ defined in~\cite{yang2019generalized}, which represents the expected vector cumulative return when taking action $a$ in state $s$ with global preference $\mathbf{w}$. The optimal $\mathbf{Q}^*$ can be expressed as:
$$\mathbf{Q}^*(s, a, \mathbf{w}) = \arg_\mathbf{Q} \sup_{\pi \in \Pi} \mathbf{w}^\top \mathbf{Q}^{\pi}(s,a)$$
In terms of network architecture, the model takes  $s$ and $\mathbf{w}$ as input and outputs the corresponding vector value. The scalar utility under that preference can be calculated via:
$$U(s,a,\mathbf{w}) = \mathbf{w}^\top \mathbf{Q}(s,a,\mathbf{w})$$

The corresponding greedy action is
$$\pi_{\mathbf{Q}}(s_t, \mathbf{w}_t) = \arg\max_{a \in A} \mathbf{w}_t^\top \mathbf{Q}(s_t, a, \mathbf{w}_t).$$

Thus, the same pretrained inner controller can execute different goal-directed strategies under different preferences.

\subsubsection{Envelope Q-Learning}
We adopt Envelope Q-Learning~\cite{yang2019generalized} to pretrain the preference-conditioned value function. Its purpose is to approximate a vector value function with sufficient preference generalization capability so that the controller can provide meaningful actions across the preference space.

The inner controller is trained before the outer learning stage and subsequently frozen. This design is important for our interpretation of emotional preference: the outer process does not learn basic motor or task skills from scratch. Instead, it learns \textbf{which previously acquired goal-directed strategy should be prioritized in the current context.}

The resulting decomposition is therefore
$$\underbrace{Q(s, a, \mathbf{w})}_{\text{how to act}} + \underbrace{\mathbf{e}_\theta(s)}_{\text{what to prioritize}} .$$

This separation also makes the preference vector directly interpretable because each component corresponds to one explicitly defined objective.

\subsection{Outer Emotional Preference MDP}
We formulate the preference-generation problem as an outer single-objective Markov decision process.

$$\mathcal{M}_{\text{out}} = \langle\mathcal{S}, \Omega, \mathcal{A}, P, r_{\text{out}}, \mathbf{Q}, \gamma_{\text{out}}\rangle$$
Here, the \textbf{direct action space of the outer agent} is the preference simplex $\mathbf{w} \in \Omega = \Delta^{m-1}$, whereas $\mathcal{A}$ is the \textbf{indirect physical action space} executed by the frozen inner controller.

The outer agent does not directly choose $a_t$. Instead, $\mathbf{w}_t = \mathbf{e}_{\theta}(s_t)$ and the physical action is generated by $a_t=\arg\max_{a \in A} \mathbf{w}_t^\top \mathbf{Q}(s_t, a, \mathbf{w}_t).$ The resulting outer trajectory is $\tau_{\text{out}} = \{s_0, \mathbf{w}_0, a_0, s_1, \mathbf{w}_1, a_1, \dots\}.$ The outer optimization objective is
$$\max_{\theta} J_{out}(\theta) = \max_{\theta} \mathbb{E}_{\tau_{\text{out}}} \left[ \sum_{t=0}^{\infty} \gamma_{\text{out}}^t r_{\text{G}}(s_t, a_t) \right]$$

This formulation captures the central computational problem studied in this paper: \textbf{the agent learns a state-dependent preference policy that regulates the priority of competing lower-level objectives in order to achieve a high-level goal.}

\subsubsection{Goal-Directed Preference Cycle}

The interaction between the outer and inner processes can be viewed as a computational goal-directed cycle(see Figure~\ref{fig:goalcycle}).

The outer reward evaluates the long-term consequence of the selected strategy with respect to G. As training proceeds, the preference generator learns which objective trade-offs tend to be advantageous in different goal-relevant states.

Importantly, the preference generator does not receive an explicit rule. Instead, such a mapping, when useful for the outer goal, is discovered through optimization. Consequently, the resulting preference function constitutes the computational object whose emergence we investigate.

\subsection{Outer Emotional Preference Learning}
After the inner MORL controller has converged, its parameters are frozen. We then train the preference generator using outer reinforcement learning. 
We use a DDPG(~\cite{lillicrap2015continuous})-style actor-critic optimization as a practical optimizer over the continuous preference space. 
for preference optimization rather than claiming smoothness of the exact environmental objective.

\subsubsection{Preference Generator}
The emotional preference network is parameterized by an MLP $g_{\theta}(s)$. To guarantee that its output lies on the preference simplex, we use a softmax transformation:
$$\mathbf{w} = \mathbf{e}_\theta(s) = \mathrm{softmax}(g_\theta(s)).$$
The network therefore represents a continuous preference function over the objective simplex.

\subsubsection{Outer Critic}
We introduce an outer critic
$$Q_\psi^{\text{out}}(s,\mathbf{w})$$
which estimates the expected discounted outer return associated with producing preference $\mathbf{w}$ in state $s$. In the concrete implementation, the outer return $r_t^{\text{out}}$ is constructed based on survival-task-level feedback: a positive reward is given while the agent survives (trajectory not terminated), and zero reward upon termination. Introducing a termination flag $d_t \in \{0,1\}$, the temporal difference (TD) target for the outer critic is defined as:
$$y_t = r_G(s_t,a_t) + \gamma_{\text{out}} (1-d_t)\, Q_{\bar\psi}^{\text{out}}(s_{t+1}, \mathbf{e}_{\bar\theta}(s_{t+1}))$$
where $\bar\theta$ and $\bar\psi$ represent the parameters of the target actor and target critic, respectively. The loss function of the outer critic adopts the mean squared error:
$$\mathcal{L}_{\text{critic}} = \mathbb{E} \left[ \left( Q_\psi^{\text{out}}(s_t,\mathbf{w}_t) - y_t \right)^2 \right]$$

In the survival experiments, the outer reward is defined as
\begin{equation}
r_G(s_t, a_t) = 
\begin{cases}
1, & \text{if the agent survives}, \\
0, & \text{if the trajectory terminates}.
\end{cases}
\end{equation}
Therefore, the preference generator is not explicitly rewarded for low-level goals. Those behaviors are modulated insofar as their resulting objective trade-offs contribute to the long-term outer goal.

\subsubsection{Outer Actor Update}

The outer actor is updated via the deterministic policy gradient, aiming to maximize the Critic's evaluation:
$$\mathcal{L}_{\text{actor}} = -\mathbb{E} \left[ Q_\psi^{\text{out}}(s_t,\mathbf{e}_\theta(s_t)) \right]$$

\textbf{Target Network Update:}
To improve the stability of the outer preference training, both the target Actor and Critic employ a soft update mechanism:
$$\bar\theta \leftarrow \tau \theta + (1-\tau)\bar\theta, \qquad \bar\psi \leftarrow \tau \psi + (1-\tau)\bar\psi$$
where $\tau \ll 1$ is the soft update coefficient.

\textbf{Use of History States:}
At present, to verify that emotional preferences can emerge purely from the immediate situation without any memory, the input to the outer network has not yet introduced history states. In the future, RNNs or Transformers may be introduced to explicitly model history.

\subsection{Two-Stage Training Procedure}
We adopt a decoupled two-stage training strategy.

\subsubsection{Stage 1: Learning the Goal-Directed Behavioral Repertoire}
First, the inner MORL controller is pretrained using Envelope Q-Learning. The objective is to acquire a preference-conditioned behavioral repertoire that covers a sufficiently broad range of objective trade-offs.

Once the inner value function has converged, its parameters are frozen.

This stage answers the question: \textbf{How can the agent act under different objective priorities?}

\subsubsection{Stage 2: Learning Emotional Preference Regulation}

In the second stage, only the outer preference generator and outer critic are optimized. At each state, the preference generator produces
$\mathbf{w} = \mathbf{e}_\theta(s)$,
which determines the relative priority of the lower-level objectives. The frozen inner controller then executes the corresponding strategy.

This stage answers the complementary question:
\textbf{Given a high-level goal, when should each lower-level objective become behaviorally prioritized?}

The complete learning architecture therefore separates \textbf{skill acquisition} from \textbf{preference regulation}.

The first stage establishes the space of available goal-directed behaviors, whereas the second stage learns how to dynamically prioritize those behaviors according to the current state and the high-level objective.

\subsection{Interpretation of the Learned Preference}

The learned preference vector should be interpreted as a computational state of relative goal priority rather than as a manually assigned instruction.

Consider a two-objective setting with 
\[
\mathbf{w}_t = (w_t^{\mathrm{achievement}}, w_t^{\mathrm{energy}})^\top.
\]
A preference such as $\boldsymbol{w}_t \approx (0,1)^\top$ indicates strong priority for energy-related behavior, while $\boldsymbol{w}_t \approx (1,0)^\top$ indicates strong priority for achievement-related behavior.

The important property is not the particular numerical value of the preference, but its state-dependent transformation:
\[
s_i \to \mathbf{w}_i, \quad s_j \to \mathbf{w}_j, \quad \mathbf{w}_i \neq \mathbf{w}_j.
\]
Thus, the outer network can dynamically reorganize behavioral priorities even though the underlying objective set and inner behavioral repertoire remain unchanged.

This property provides the computational basis for the notion of an emergent emotional preference adopted in this paper: a high-level goal induces, through reinforcement learning, a context-dependent regulation of the relative priorities of competing lower-level goals.

\section{Experiments}\label{sec:experiments}

\subsection{Experimental Questions}
We evaluate the proposed framework from three complementary perspectives.

\textbf{First, can the inner MORL controller acquire a sufficiently diverse and interpretable repertoire of goal-directed behaviors?}

This question is important because the outer preference generator can only regulate priorities among behaviors represented by the pretrained inner controller.

\textbf{Second, can a high-level goal induce an emergent emotional preference rather than relying on manually specified preference rules?}

We therefore examine whether optimization of the outer survival objective produces state-dependent preference patterns that systematically correspond to different environmental conditions and competing lower-level objectives.

\textbf{Third, does the learned emotional preference provide functional benefits over fixed and handcrafted preference strategies?}

We compare the proposed model with fixed-preference policies, handcrafted contextual preference rules, and a policy directly optimized for the outer survival objective.

The experiments are conducted in two environments. The basic environment contains achievement and energy objectives and is designed to provide a controlled setting in which the preference-regulation mechanism can be analyzed in detail. The advanced environment, described in Appendix I, additionally introduces a safety objective and stochastic danger, allowing us to examine whether the same mechanism extends to a richer multi-objective setting. The basic environment is used for the main analysis, while the advanced environment provides an additional robustness evaluation.

\subsection{Experimental Environment}

\textbf{Basic environment:} The main objective of the basic experiments is to obtain achievement reward and energy reward. Through the Grid Fruit Tree Battery Exploration game (the environment is introduced in the next section), we hope that, guided by the outer high-level goal of maximizing survival time, contextualized emotional preferences will emerge.  Overall, this is a simple and fast small-scale validation environment for emotional preferences. We intend to build the simplest possible environment that promotes the emergence of emotional preference, analogous to the MNIST handwritten digit dataset in the image domain (approximately 70,000 discrete states without considering symmetry), in order to clearly dissect and demonstrate the existence and characteristics of the ``emergence'' phenomenon itself. This also implies that the involved contextualized emotions may be relatively simple. The main experimental content of our paper is based on the basic environment.

\textbf{Advanced environment:} The main objective of the advanced experiments is to obtain achievement reward, energy reward, and safety reward. Through the Grid Fruit Tree Battery Danger Zone Probabilistic Exploration game, we hope that, guided by the outer high-level goal of maximizing survival time, contextualized emotional preferences will emerge. The environment is a medium-complexity validation environment for emotional preferences. Without considering symmetry, this environment has approximately 260 million states, examining whether the
proposed state-dependent emotional preference regulation mechanism remains interpretable when more competing goal-directed objectives are introduced. The corresponding experimental content is presented in the Appendix.

\subsubsection{Introduction to the Basic Environment}

We construct a synthetic \textbf{Grid Fruit Tree Battery Exploration} environment. The environment is a $3\times3$ grid. There are two types of objects in the environment: fruit trees and battery boxes. A fruit tree has 1 to 4 fruits. A battery box contains 1 to 3 batteries. There is only one fruit tree and one battery box in the game. The positions of the fruit tree and the battery box, as well as the initial position of the agent, are random, and the fruit tree and battery box positions do not overlap. The agent has six discrete actions: move up, move down, move left, move right, collect, and ask for help. Rewards are divided into two parallel dimensions: achievement reward and energy reward. The agent starts with 6 points of energy. After any action is resolved, one point of energy is consumed and one point of energy reward is reduced. If energy reaches 0, the agent receives a penalty of -10 achievement reward and the game ends. After choosing to ask for help, the game ends after the energy consumption for that action is resolved. Collecting one fruit yields 5 points of achievement reward. Collecting one battery restores energy to full and consumes one point of energy, yielding an energy reward equal to twice the energy change (e.g., if energy increases from 1 to 5, the energy reward is 8).

The agent's observation state includes: the agent's position, the agent's current energy, the agent's maximum energy, the number of fruit trees, the number of battery boxes, [the position of the fruit tree, the number of fruits on it, whether the fruit tree has been discovered] (when the fruit tree has not been discovered, a placeholder value [(-1,-1),0,0] is input), [the position of the battery box, the number of batteries in it, whether the battery box has been discovered] (when the battery box has not been discovered, a placeholder value [(-1,-1),0,0] is input), and a $3\times3$ exploration mask (explored cells are marked as 1, otherwise 0). Currently, by default, the positions of the fruit tree and the battery box are fully revealed. In the future, we will consider a partially observable Markov decision process setting in which the items only become visible when the agent is next to them.

\subsection{Validation of the Inner Multi-Objective Controller}
Before evaluating emotional preference learning, we first verify that the inner MORL controller provides distinct behaviors under different preferences.

We train the inner controller using Envelope Q-Learning and evaluate it using a set of fixed preferences $(1,0)^\top$, $(0.9,0.1)^\top$, $(0.8,0.2)^\top$, $(0.7,0.3)^\top$, $\cdots$, $(0,1)^\top$.

For each preference, we execute complete episodes and measure the resulting achievement and energy returns.

\subsubsection{Preference-conditioned objective trade-offs}
\begin{figure}[htb]
  \centering
  \includegraphics[width=0.5\linewidth]{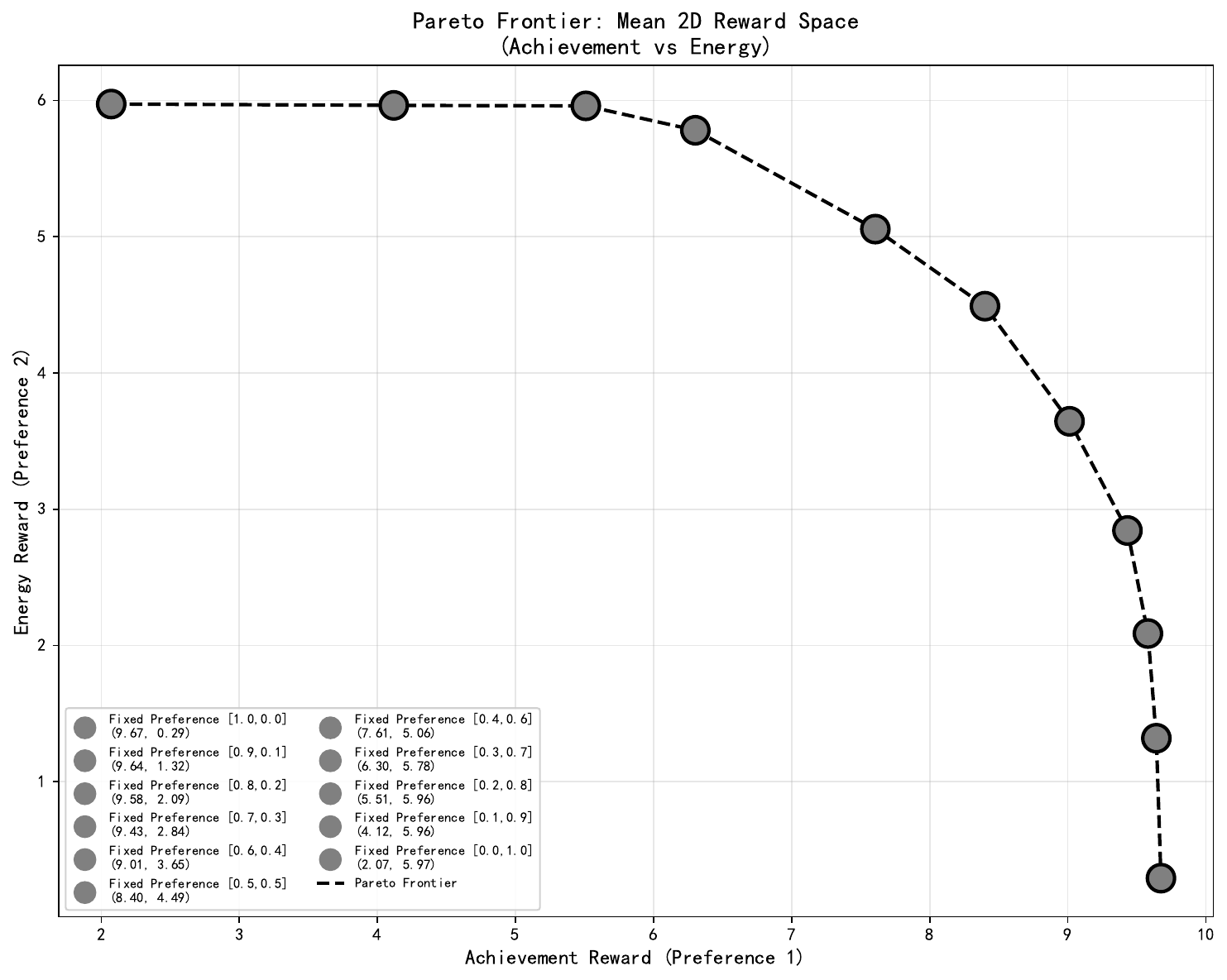}
  \caption{Scatter plot of two-dimensional average reward for various preferences. The rewards are obtained from statistics over 200,000 episodes.}\label{fig:pareto_frontier}
\end{figure}

Figure~\ref{fig:pareto_frontier} shows the mean two-dimensional reward obtained under different fixed preferences. The reward vectors form a clear trade-off structure between achievement and energy objectives. Intermediate preferences generate intermediate trade-offs, with the observed points forming an approximately convex coverage of the achievable reward region. The statistics are computed over 200,000 evaluation episodes.

This result establishes an important prerequisite for the proposed framework: the inner controller does not merely produce a single behavior independent of the preference. Instead, different preference values expose substantially different goal-directed behaviors.

\subsubsection{Behavioral consequences of preference changes}
According to Table~\ref{tab:results_no_reward}, the preference-conditioned policies also exhibit systematic differences in survival time and resource utilization. As the energy preference increases from $(1,0)^\top$  toward intermediate values, survival time initially increases because the agent increasingly uses battery collection to maintain energy while still pursuing achievement. When energy becomes over-prioritized, however, the policy tends to terminate by requesting help after collecting available batteries, which reduces the resulting survival time.

Similarly, the number of collected fruits increases as the achievement preference increases, while battery collection becomes more strongly associated with energy-oriented preferences. These observations confirm that the preference vector has a direct and interpretable influence on the lower-level behavioral repertoire.

Therefore, before introducing the outer preference generator, the inner controller already provides a set of distinguishable strategies with explicit objective semantics. The outer learning problem can consequently be interpreted as \textbf{regulating the relative priority among these existing goal-directed strategies.}

\subsection{Baselines for Preference Regulation}
We compare four categories of behavior.

\textbf{Fixed-preference policies}

For each fixed preference $\mathbf{w} \in \{(1,0)^\top, (0.9,0.1)^\top, \ldots, (0,1)^\top\}$, the same preference is used throughout the episode. These policies represent conventional MORL usage in which the objective trade-off is specified globally.

\textbf{Pure survival policy}

A separate controller is directly optimized for the outer survival objective. This provides an upper-performance reference for survival-oriented decision-making, although it does not explicitly represent or regulate the lower-level objective preferences.

\textbf{Handcrafted contextual preference policies}

We construct two simple state-dependent baselines.

The first uses the agent's energy level:
\[
\textbf{w}(s) = 
\begin{cases}
(0.1,0.9), & E(s) \le 3, \\
(0.9,0.1), & E(s) > 3.
\end{cases}
\]

The second uses the number of remaining batteries:
\[
\textbf{w}(s) = 
\begin{cases}
(0,1), & N_{\text{battery}}(s) \ge 1, \\
(1,0), & N_{\text{battery}}(s) = 0.
\end{cases}
\]

These baselines are important because they represent explicit dynamic preference rules constructed from human knowledge. They therefore allow us to distinguish learned preference regulation from simply hard-coding an intuitive context-dependent policy.

\textbf{Emotional Preference Model}
The proposed model learns $\mathbf{w} = \mathbf{e}_\theta(s)$ solely from the outer survival objective, while the inner MORL controller remains frozen. No explicit rule specifies which environmental state should correspond to which preference.

\subsection{Emergence and Dynamics of Emotional Preferences}
We next examine whether the outer survival objective produces a nontrivial state-dependent emotional preference function.

\subsection{Distribution of learned preference states}

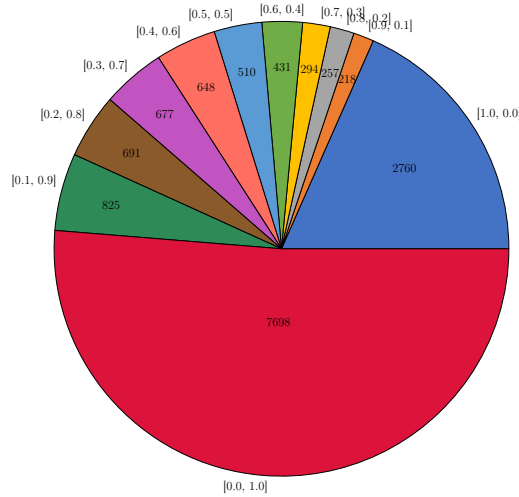
\begin{figure}[htb]
  \centering
  \resizebox{0.5\textwidth}{!}{%
    \begin{tikzpicture}
      \tikzset{every pin/.style={font=\tiny, align=center}}
      \pie[
        sum=auto,
        radius=7.8,
        color={c1,c2,c3,c4,c5,c6,c7,c8,c9,c10,c11},
        pos={0,0,0,0,0,0,0,0,0,0,4}
      ]
      {
        2760/{[1.0, 0.0]},
        218/{[0.9, 0.1]},
        257/{[0.8, 0.2]},
        294/{[0.7, 0.3]},
        431/{[0.6, 0.4]},
        510/{[0.5, 0.5]},
        648/{[0.4, 0.6]},
        677/{[0.3, 0.7]},
        691/{[0.2, 0.8]},
        825/{[0.1, 0.9]},
        7698/{[0.0, 1.0]}
      }
    \end{tikzpicture}%
  }
  \caption{Distribution of visited states across learned emotional preference regions in the basic environment.}
  \label{fig:pie}
\end{figure}

We evaluate the trained emotional preference model over 1,000 trials, obtaining 15,009 observed states. Because the preference generator produces continuous values on the simplex, we assign each observed preference to its nearest discrete reference preference for visualization.

According to Figure~\ref{fig:pie}, more than half of the observed states are closest to $(0,1)^\top$, corresponding to strong energy priority, while approximately 18.3\% are closest to $(1,0)^\top$, corresponding to strong achievement priority. The remaining states are distributed across intermediate preference values.

The resulting distribution is important for two reasons. First, the preference generator does not simply reproduce one globally fixed preference. Second, it does not behave as a purely binary switch: intermediate preference states also occur, indicating that the learned preference function can represent graded trade-offs between competing objectives.

We therefore interpret the result as evidence that the outer objective has induced a \textbf{state-dependent preference landscape} rather than a single global scalarization weight.

\subsubsection{Context sensitivity of preference regulation}
We next examine which environmental conditions are associated with different preference states.

For the pure achievement preference $(1,0)^\top$, we identify 2,760 observed states. Among them, 2,576 states, or 93.3\%, occur in situations where no battery remains available for collection. This indicates a strong association between the achievement-oriented preference and states in which further energy-oriented behavior has limited practical value.

Intermediate preferences such as $(0.9,0.1)^\top$ and $(0.8,0.2)^\top$ show weaker versions of the same tendency.

This observation is important in relation to the computational definition. The preference is not determined by a single state variable through an explicitly programmed threshold. Instead, it reflects the interaction among several environmental factors represented in the state. In this sense, the outer network learns a context-dependent regulation of the relative priority of competing lower-level objectives.

\subsubsection{Preference transitions within an episode}

\begin{figure}[htbp]
  \centering
  \begin{tabular}{cc}
    \includegraphics[width=0.40\textwidth]{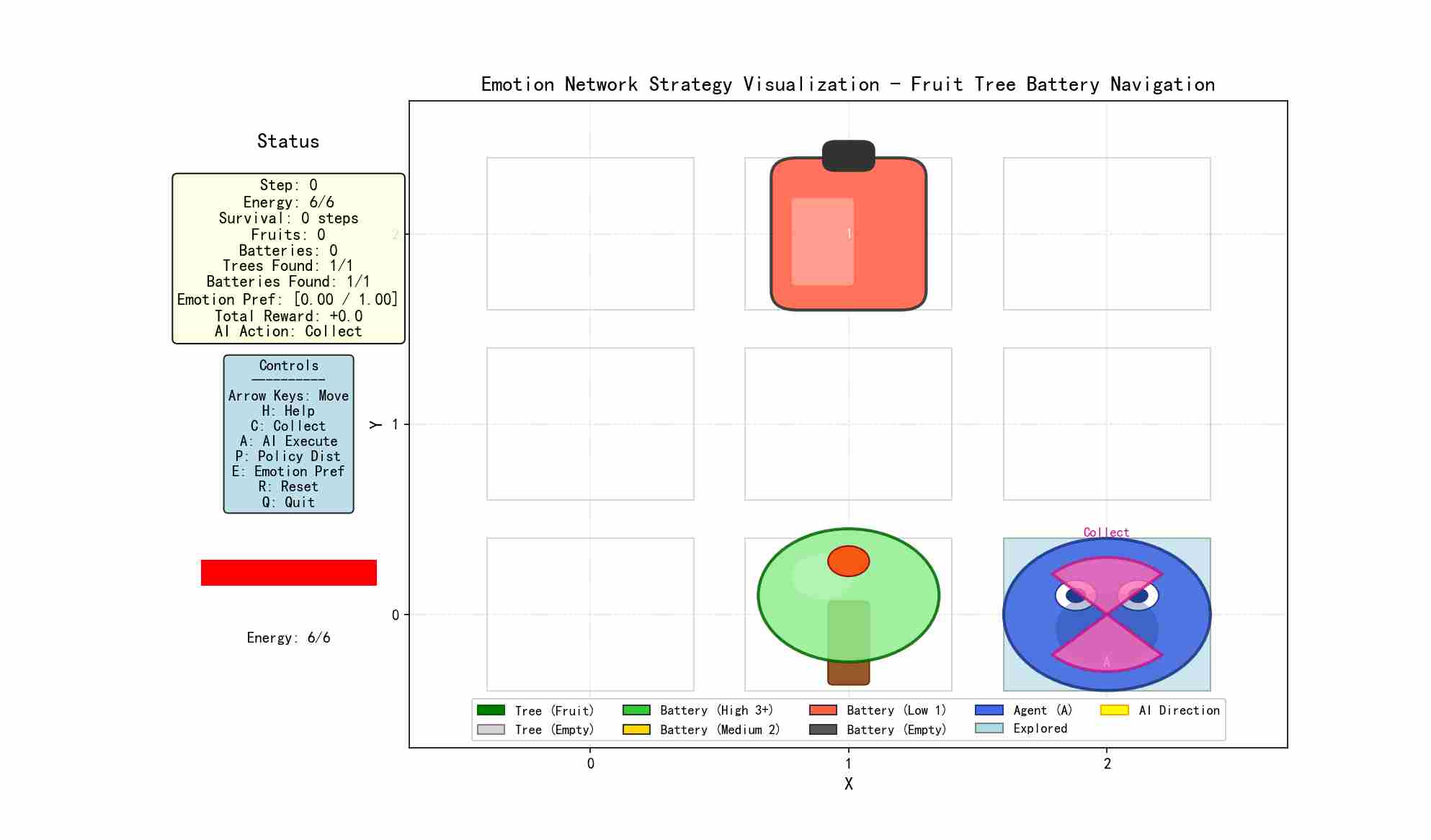} &
    \includegraphics[width=0.40\textwidth]{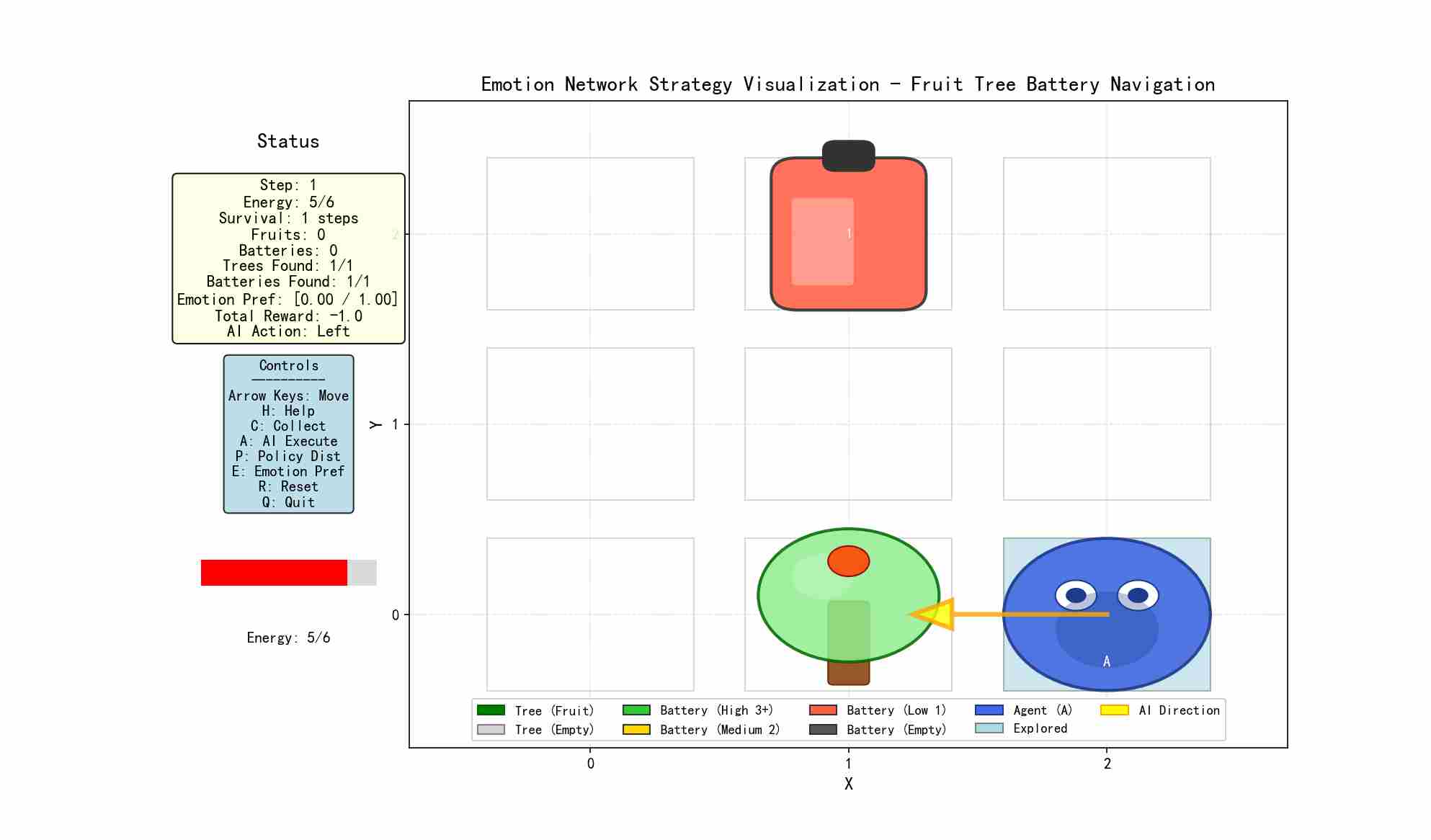} \\
    \includegraphics[width=0.40\textwidth]{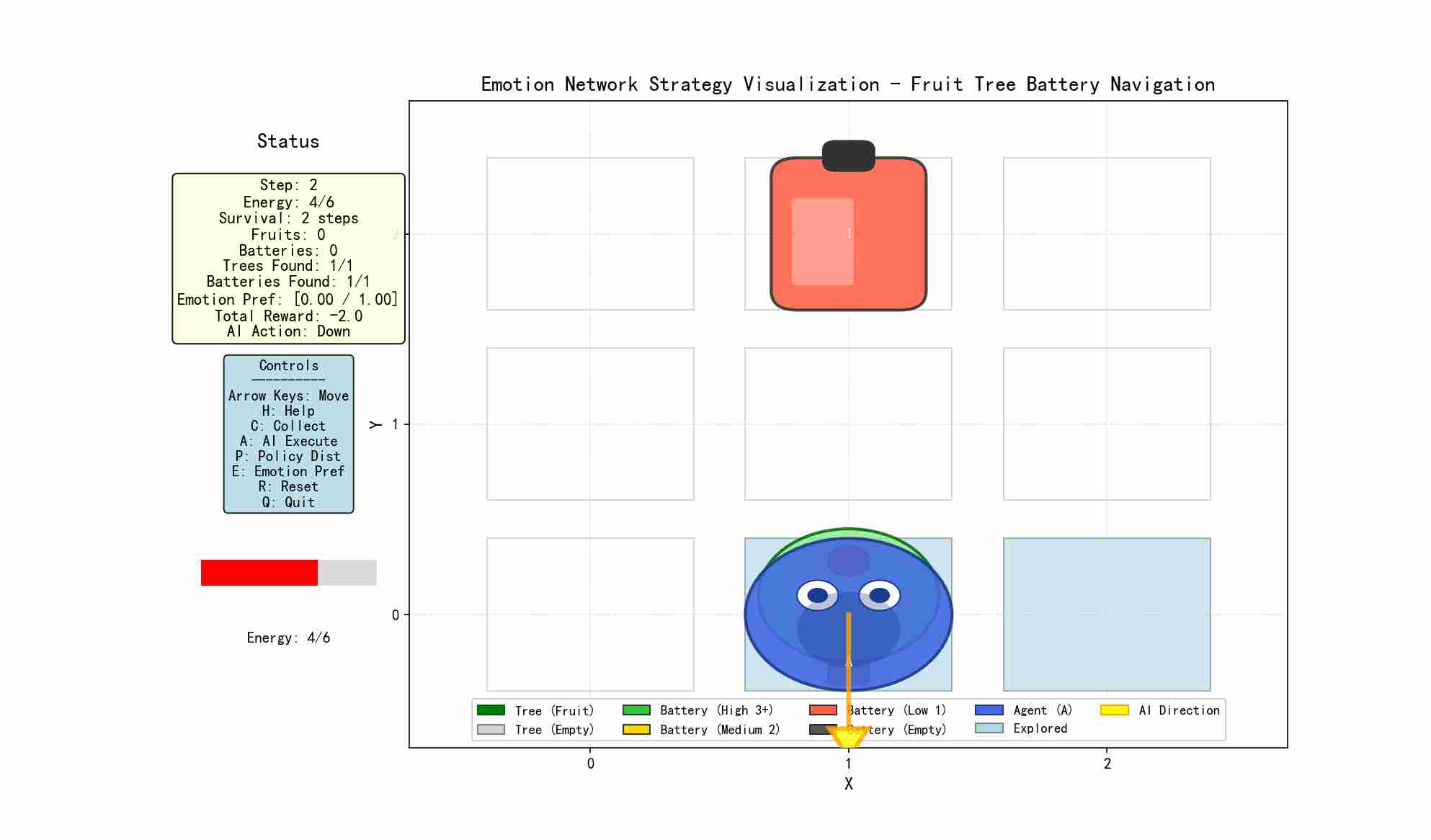} &
    \includegraphics[width=0.40\textwidth]{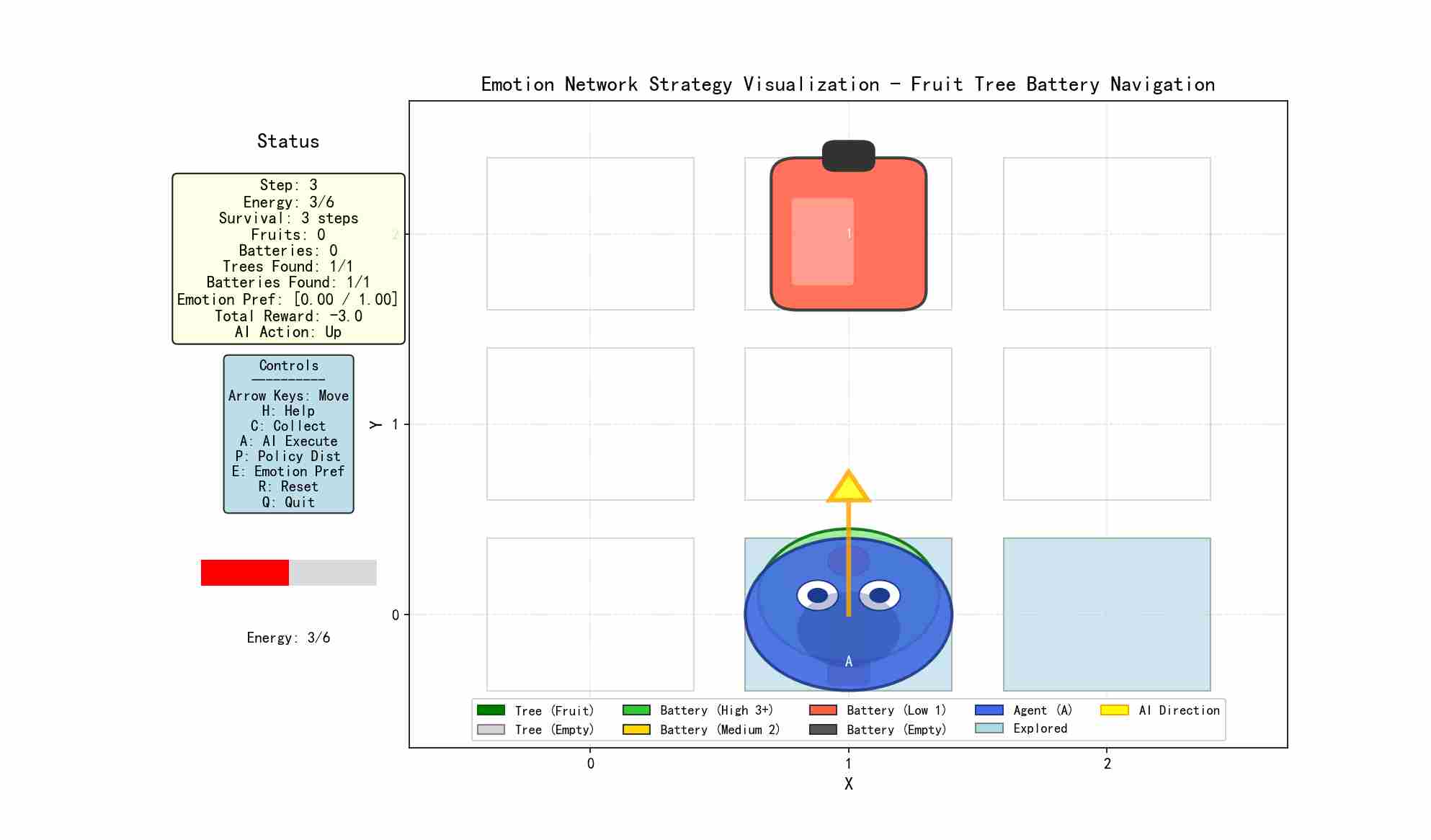} \\
    \includegraphics[width=0.40\textwidth]{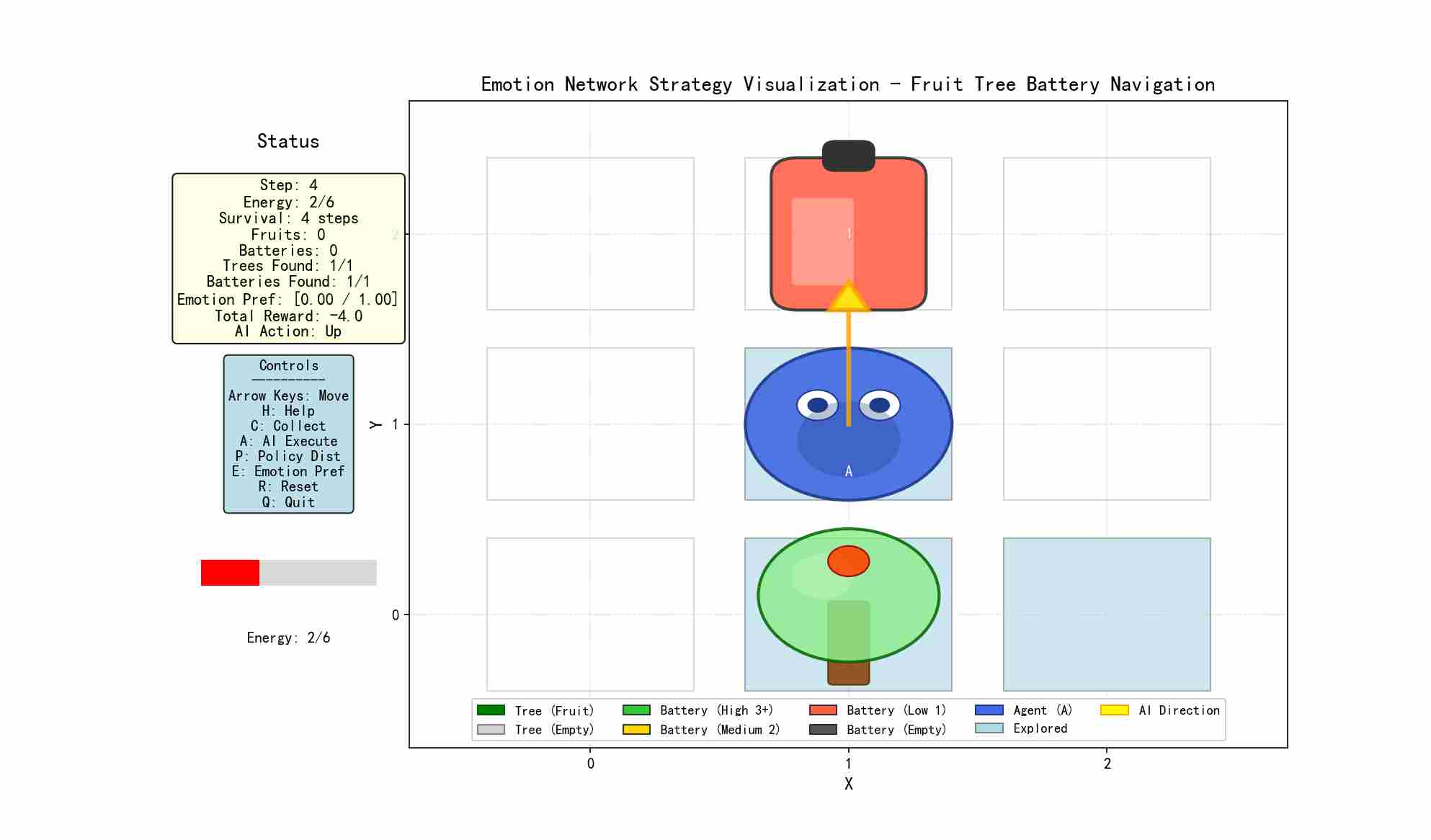} &
    \includegraphics[width=0.40\textwidth]{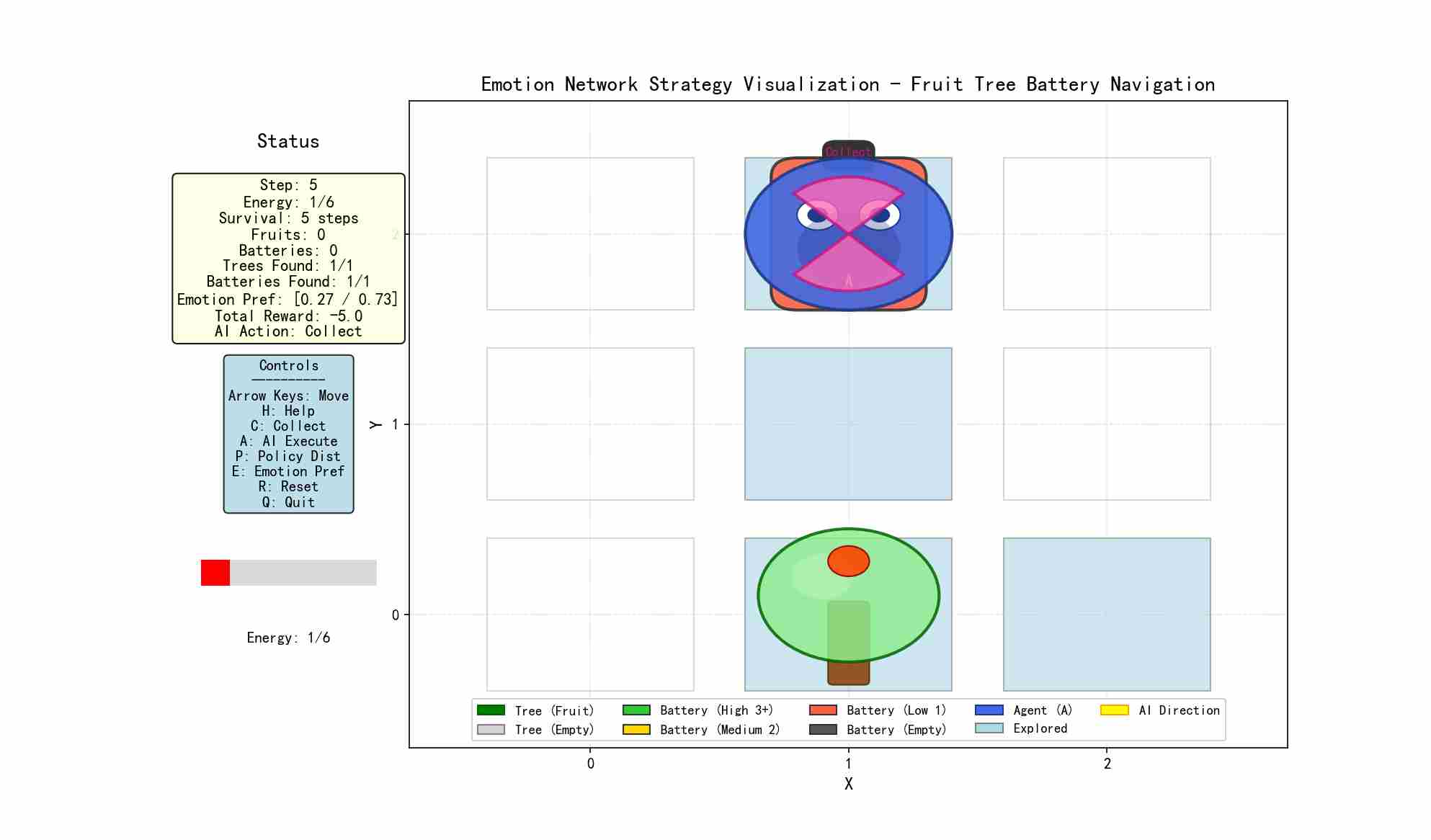} \\
    \includegraphics[width=0.40\textwidth]{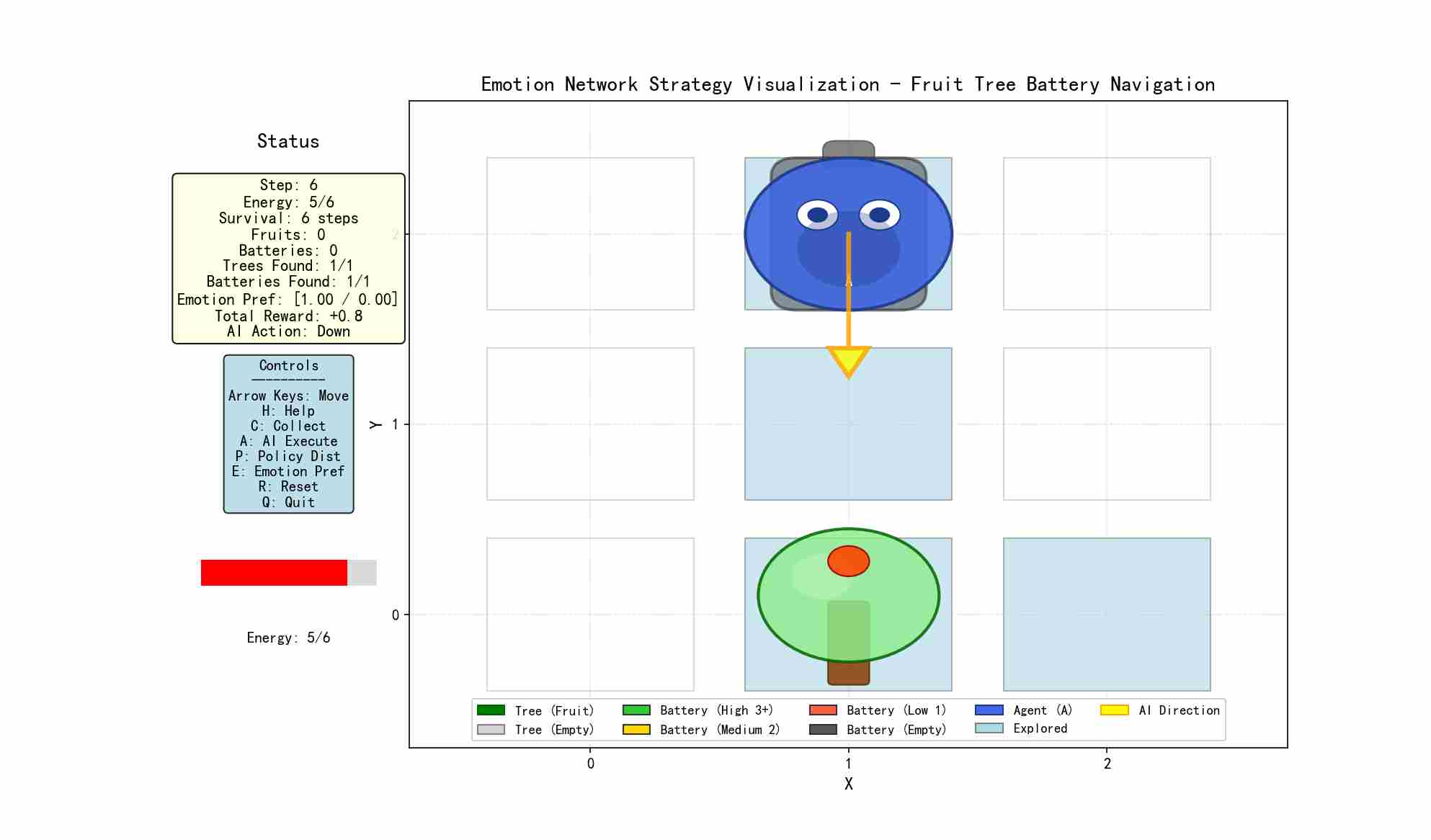} &
    \includegraphics[width=0.40\textwidth]{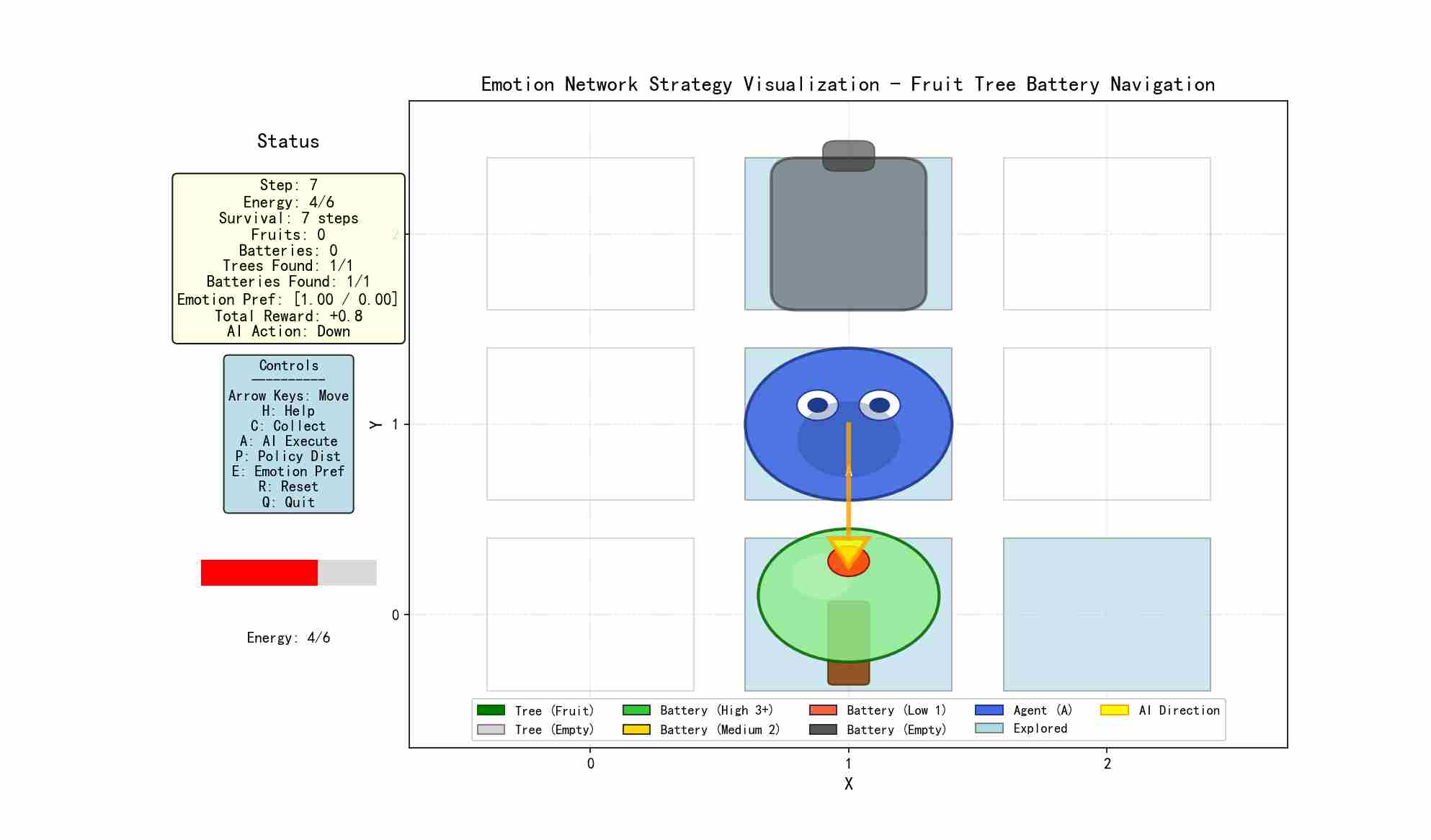} \\
    \includegraphics[width=0.40\textwidth]{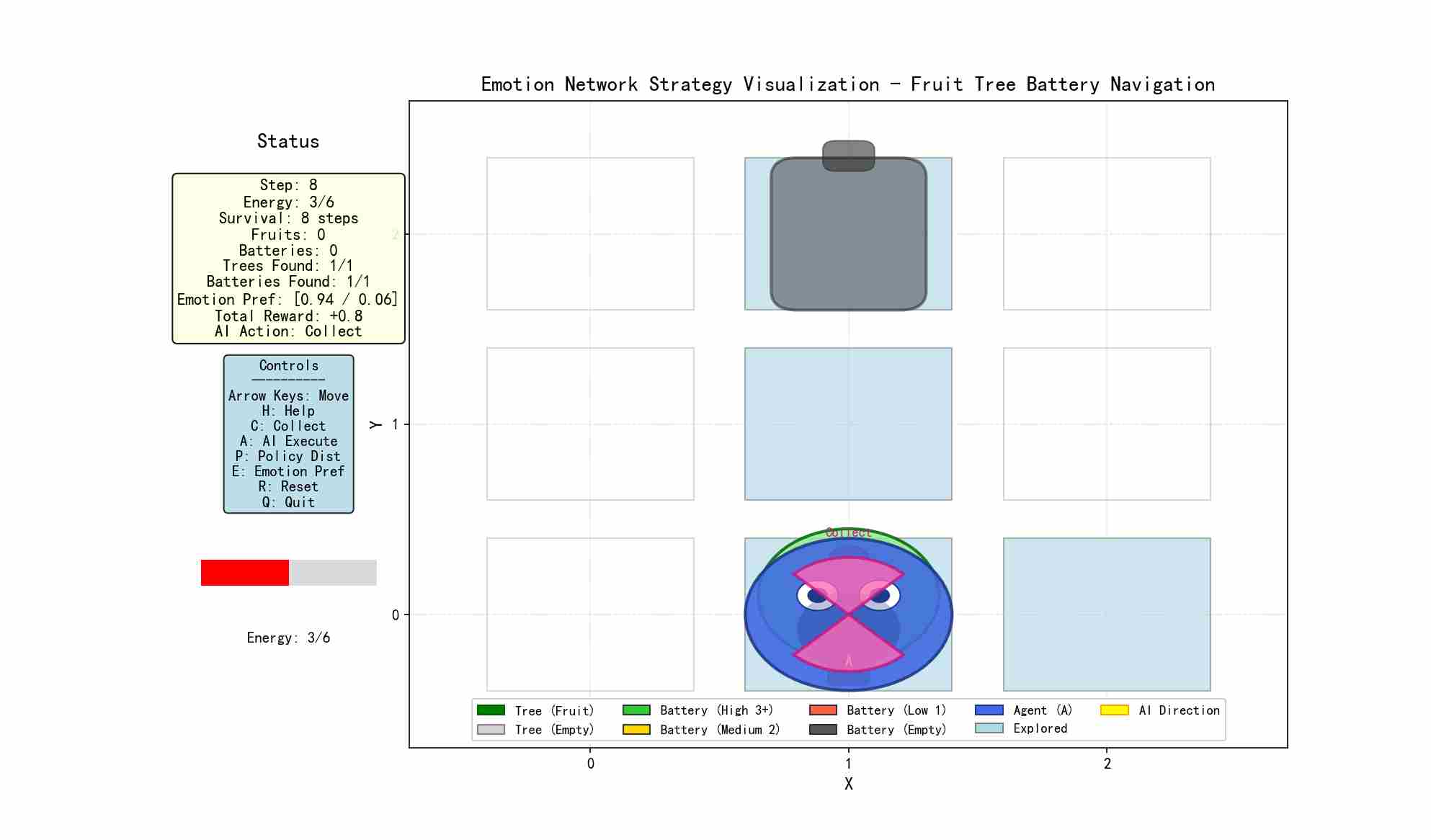} &
    \includegraphics[width=0.40\textwidth]{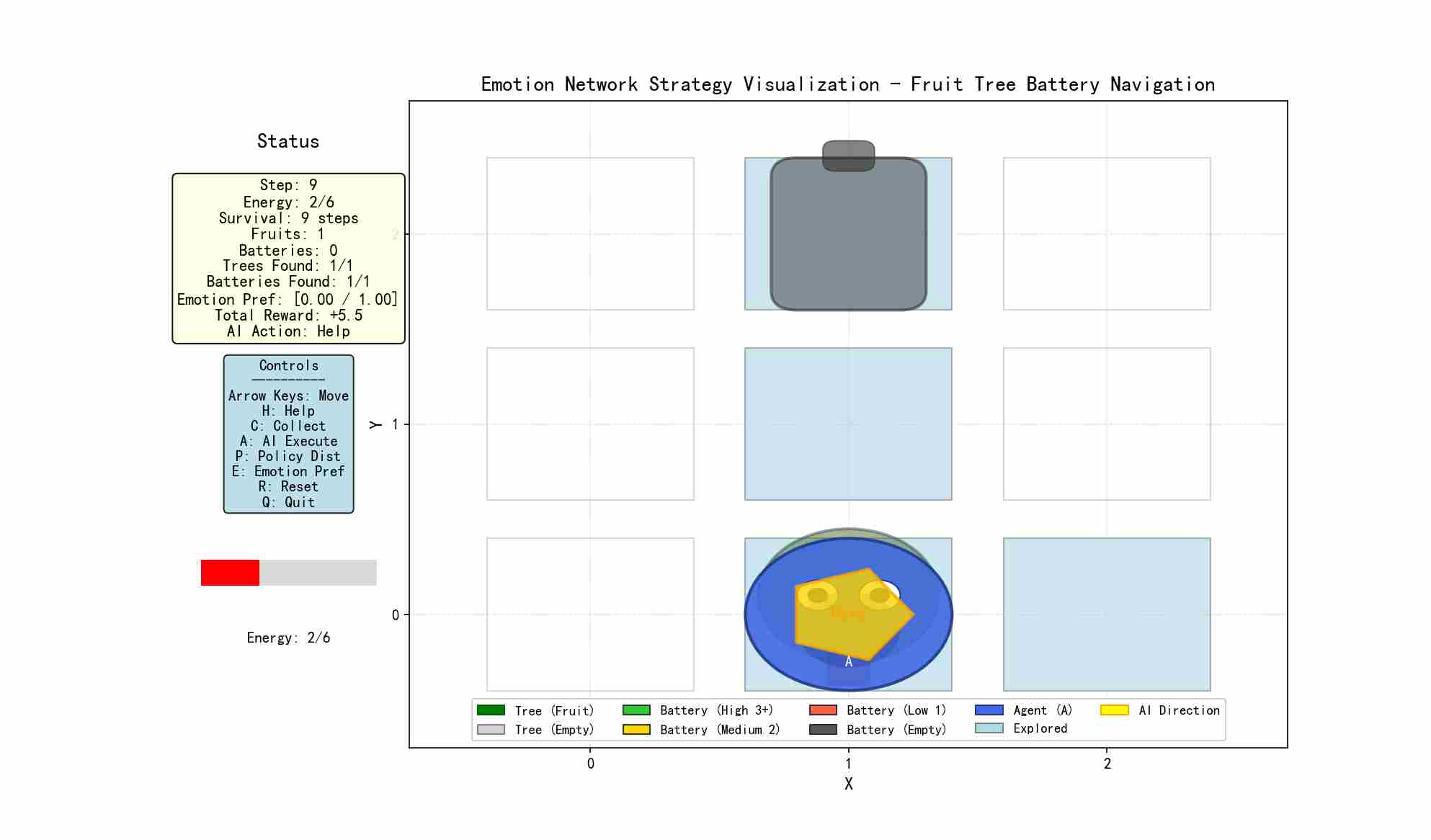} \\
  \end{tabular}
  \caption{Representative trajectory showing state-dependent emotional preference regulation. The preference shifts between energy- and achievement-oriented priorities as the goal-relevant environmental state changes.}
  \label{fig:simulate representation}
\end{figure}
Figure~\ref{fig:simulate representation} illustrates a representative episode.

At the beginning of the episode, the preference remains close to $(0,1)^\top$ for approximately five consecutive steps, favoring energy acquisition. Immediately before battery collection, the preference moves toward $(0.27,0.73)^\top$. After the battery is collected, the preference shifts toward $(1,0)^\top$, after which the agent begins pursuing fruit collection. As energy becomes scarce again, the preference shifts back toward the energy objective.

The important observation is not simply that the preference changes, but \emph{when} it changes. Preference transitions tend to occur around task-relevant environmental changes, while preferences remain relatively stable during periods in which the current strategy continues to serve the high-level survival goal.

This provides evidence for two properties of the learned emotional preference function:
\begin{itemize}
    \item context sensitivity
    \item temporal persistence.
\end{itemize}

We refer to the latter as preference persistence rather than claiming that the experiment independently establishes the full psychological phenomenon of emotional inertia.

The corresponding statistical analysis in Table~\ref{tab:preference persistence} shows that the average duration of segments associated with the energy and achievement extremes is approximately 3 and 2 steps, respectively. Across approximately 15-step episodes, an average of 3.56 preference segments with duration greater than one step are observed.

\begin{table}[htbp]
\centering
\caption{Emotional Preference Persistence Analysis: The duration of the corresponding emotional preference when the target preference value falls within the range of ±0.1.\label{tab:preference persistence} The table records the mean, median, minimum, and maximum durations of emotional preferences, as well as the total number of continuous segments. The specific data is derived from the results of the previous 1,000 trials.} 
\resizebox{0.5\textwidth}{!}{%
\begin{tabular}{lccccc}
\toprule
Target Preference & Average Durations  & Median & Min & Max & Count \\
\midrule
{[1.0,0.0]} & $2.46 \pm 0.85$ & 3.0 & 1.0 & 9.0 & 1124 \\
{[0.9,0.1]} & $1.12 \pm 0.34$ & 1.0 & 1.0 & 3.0 & 195 \\
{[0.8,0.2]} & $1.10 \pm 0.33$ & 1.0 & 1.0 & 3.0 & 234 \\
{[0.7,0.3]} & $1.12 \pm 0.37$ & 1.0 & 1.0 & 4.0 & 263 \\
{[0.6,0.4]} & $1.17 \pm 0.45$ & 1.0 & 1.0 & 4.0 & 367 \\
{[0.5,0.5]} & $1.15 \pm 0.46$ & 1.0 & 1.0 & 5.0 & 443 \\
{[0.4,0.6]} & $1.20 \pm 0.49$ & 1.0 & 1.0 & 5.0 & 541 \\
{[0.3,0.7]} & $1.14 \pm 0.39$ & 1.0 & 1.0 & 4.0 & 596 \\
{[0.2,0.8]} & $1.15 \pm 0.42$ & 1.0 & 1.0 & 5.0 & 603 \\
{[0.1,0.9]} & $1.21 \pm 0.48$ & 1.0 & 1.0 & 4.0 & 680 \\
{[0.0,1.0]} & $2.96 \pm 2.68$ & 2.0 & 1.0 & 16.0 & 2598 \\
\bottomrule
\end{tabular}}
\end{table}

\subsection{Quantitative Comparison of Preference-Regulation Strategies}
\begin{table}[htb]
\centering
\caption{Performance comparison of preference-regulation strategies in the basic environment. The statistical data is derived from 200,000 episodes.}
\label{tab:results_no_reward}
\resizebox{\columnwidth}{!}{
\begin{tabular}{lcccc}
\toprule
Model & Avg Survival Steps & Avg Help & Avg Batteries & Avg Fruits \\
\midrule
\textcolor{blue}{Emotional Preference Model}  & $14.97 \pm 4.18$ & $0.64 \pm 0.48$ & $2.11 \pm 0.92$ & $1.37 \pm 1.17$ \\
\textcolor{red}{Survival Model} & $16.00 \pm 4.08$ & $0.03 \pm 0.17$ & $2.12 \pm 0.94$ & $0.09 \pm 0.35$ \\
\textcolor{orange}{Handcraft Model(energy)} & $11.81 \pm 5.21$ & $0.90 \pm 0.30$ & $1.61 \pm 1.16$ & $1.47 \pm 1.09$ \\
\textcolor{green}{Handcraft Model(battery)} & $14.53 \pm 4.11$ & $1.00 \pm 0.00$ & $2.10 \pm 0.91$ & $1.24 \pm 1.03$ \\
Fixed Preference [1.0,0.0] & $10.97 \pm 5.19$ & $1.00 \pm 0.00$ & $1.43 \pm 1.22$ & $1.99 \pm 1.02$ \\
Fixed Preference [0.9,0.1] & $11.58 \pm 5.20$ & $1.00 \pm 0.00$ & $1.55 \pm 1.18$ & $1.97 \pm 1.03$ \\
Fixed Preference [0.8,0.2] & $11.81 \pm 5.12$ & $1.00 \pm 0.00$ & $1.60 \pm 1.13$ & $1.96 \pm 1.03$ \\
Fixed Preference [0.7,0.3] & $12.02 \pm 4.94$ & $1.00 \pm 0.00$ & $1.67 \pm 1.08$ & $1.94 \pm 1.08$ \\
Fixed Preference [0.6,0.4] & $12.47 \pm 4.73$ & $1.00 \pm 0.00$ & $1.81 \pm 1.03$ & $1.86 \pm 1.14$ \\
Fixed Preference [0.5,0.5] & $12.88 \pm 4.34$ & $1.00 \pm 0.00$ & $1.95 \pm 0.93$ & $1.75 \pm 1.20$ \\
Fixed Preference [0.4,0.6] & $12.78 \pm 4.13$ & $1.00 \pm 0.00$ & $2.01 \pm 0.86$ & $1.59 \pm 1.27$ \\
Fixed Preference [0.3,0.7] & $12.16 \pm 3.95$ & $1.00 \pm 0.00$ & $2.02 \pm 0.85$ & $1.34 \pm 1.29$ \\
Fixed Preference [0.2,0.8] & $12.00 \pm 4.06$ & $1.00 \pm 0.00$ & $2.00 \pm 0.82$ & $1.17 \pm 1.19$ \\
Fixed Preference [0.1,0.9] & $11.99 \pm 4.07$ & $1.00 \pm 0.00$ & $2.00 \pm 0.82$ & $0.87 \pm 1.08$ \\
Fixed Preference [0.0,1.0] & $11.98 \pm 4.06$ & $1.00 \pm 0.02$ & $2.00 \pm 0.82$ & $0.43 \pm 0.74$ \\
\bottomrule
\end{tabular}}
\end{table}

Table~\ref{tab:results_no_reward} summarizes the performance of the proposed emotional preference model and the comparison methods.

The Emotional Preference Model achieves $14.97 \pm 4.18$ survival steps, which is higher than all tested fixed-preference policies and both handcrafted contextual preference baselines. In particular, the best fixed preference, $(0.5,0.5)$, achieves $12.88 \pm 4.34$ steps, while the battery-based handcrafted policy achieves $14.53 \pm 4.11$.

The emotional preference model therefore demonstrates an advantage over static and manually specified preference strategies in the tested environment.

At the same time, the directly optimized Survival Model achieves $16.00 \pm 4.08$ survival steps and therefore remains stronger on the outer survival metric. This difference is expected from the policy-space perspective developed in Section 5: the emotional preference model is constrained to behaviors represented by the pretrained preference-conditioned controller, whereas the direct survival controller is not subject to this representation constraint.

The comparison therefore reveals a useful distinction between the two approaches. The Survival Model maximizes the outer objective most directly, whereas the Emotional Preference Model simultaneously maintains interpretable objective-level behavior modulation. The latter collects substantially more fruits than the Survival Model: $1.37 \pm 1.17$ vs.\ $0.09 \pm 0.35$,
while achieving nearly the same battery collection:$2.11 \pm 0.92$ vs.\ $2.12 \pm 0.94$.

It also asks for help much more frequently than the Survival Model.

These results suggest that outer preference learning does not merely reproduce the behavior of a pure survival optimizer. Instead, it produces a different behavioral organization in which multiple objective-conditioned strategies can be activated according to the current situation.

\subsection{Why Learned Preference Regulation Is More Than a Fixed Rule}
A central question is whether the learned preference generator simply rediscovers a simple manually specified rule.

The handcrafted energy-based model uses only the current energy level to switch between achievement and energy preferences. Its survival performance is $11.81 \pm 5.21$. The handcrafted battery-based model performs better, reaching $14.53 \pm 4.11$, but still remains below the Emotional Preference Model.

The learned model also exhibits cases that cannot be explained by energy alone. For example, states associated with the achievement preference are strongly enriched for situations in which no battery remains, but the complete preference distribution contains intermediate states as well. Moreover, in the representative trajectories, preference transitions occur around resource acquisition and impending energy depletion rather than at a single manually defined threshold.

Thus, the learned preference function should not be viewed simply as a learned copy of one handcrafted switching rule. Instead, it learns a more general mapping from the environmental state to the relative priority among competing lower-level objectives.

This observation is consistent with the computational interpretation:

\begin{tikzpicture}[
  font=\small\sffamily,
  every node/.style={
    draw,
    rounded corners,
    align=center,
    text width=3.2cm,
    minimum height=0.9cm,
    inner sep=4pt
  },
  >=Stealth,
  node distance=1cm and 1.2cm
]
  \node (n1) {high-level survival goal};
  \node (n2) [right=of n1] {context-dependent priority regulation};
  \node (n3) [right=of n2] {selection of goal-directed strategy};

  \draw[->] (n1) -- (n2);
  \draw[->] (n2) -- (n3);
\end{tikzpicture}

\subsection{Advanced Environment: Three-Objective Preference Regulation}
To examine whether the observed phenomenon extends beyond the two-objective setting, we additionally evaluate the framework in the advanced environment with achievement, energy, and safety objectives.

In this setting, the learned preference distribution covers a richer region of the three-dimensional simplex. According to Figure~\ref{fig:pie2}, across 100 trials, 1,478 states are observed. The largest preference groups include the energy-dominant preference $(0,1,0)^\top$, the safety-dominant preference $(0,0,1)^\top$, and the achievement-dominant preference $(1,0,0)^\top$.

The observed counts are 568, 367, and 186, respectively, corresponding to approximately 38.4\%, 24.8\%, and 12.6\% of the observed states. Other states occupy intermediate regions of the three-dimensional preference simplex.

A representative trajectory further demonstrates the multi-objective regulation (see Figure~\ref{fig:outer_network_simulate_representation_advanced}). The preference initially remains close to $(0,0,1)^\top$, corresponding to safety priority, then moves toward a mixed energy-safety preference before battery collection. After the battery is collected, it returns toward the safety objective, and later shifts toward energy priority as energy becomes scarce.

The quantitative comparison in the advanced environment gives an average survival duration of $14.81 \pm 9.56$ for the Emotional Preference Model  (see Table~\ref{tab:results_multi}), compared with $16.26 \pm 9.04$ for the pure Survival Model. The Emotional Preference Model nevertheless exhibits substantially richer achievement behavior, collecting on average $0.76 \pm 1.24$ fruits compared with $0.03 \pm 0.25$ for the Survival Model, while maintaining $1.14 \pm 1.12$ battery collections.

These results provide additional evidence that the proposed mechanism is not restricted to a binary trade-off. With three competing objectives, the outer optimizer produces multiple contextual preference states and transitions among them according to the current goal-relevant situation.

\section{Theoretical Analysis}

The experiments in the previous section show that the outer preference generator can produce state-dependent emotional preferences and can outperform the tested fixed-preference policies. However, the learned preference generator is constrained by the behavioral repertoire provided by the pretrained inner MORL controller. This constraint explains why the Emotional Preference Model can remain below the directly optimized survival policy, despite being able to dynamically regulate objective priorities.

In this section, we formalize this constraint and characterize its effect on outer-goal performance. Our analysis establishes three results.

First, the outer preference generator searches over a structured policy space induced by the inner MORL policy family. Second, the optimality gap between this restricted space and the unrestricted policy space is controlled by the policy representation error of the inner controller. Third, useful behaviors represented by the inner policy family can be inherited and selectively activated by the outer preference generator, providing a formal basis for the state-dependent behavioral reorganization observed in the previous section.

For analytical clarity, we first consider a finite state space, finite action space, and a finite set of representative preferences. The continuous preference formulation used in the implementation can be viewed as the corresponding function-space extension, while the finite setting provides the matrix representation required for the following derivations.

\subsection{Composite Policy Space Induced by Emotional Preference Regulation}

Let the inner MORL controller provide \(k\) deterministic preference-conditioned base policies $\{\mathbf{\pi}^{\mathbf{w}_1},\ldots,\mathbf{\pi}^{\mathbf{w}_k}\}$ and their matrix
\[
\mathcal{P}_{\mathrm{in}}=\{\mathbf{\Pi}^{\mathbf{w}_1},\ldots,\mathbf{\Pi}^{\mathbf{w}_k}\}\subset\mathcal{P}_{\mathrm{all}},
\]
where the unrestricted policy space is denoted by \(\mathcal{P}_{\mathrm{all}}\), which contains all stationary randomized policies over the primitive action space. Each representative preference \(\mathbf{w}_j\) induces the greedy action
\[
\pi_{\mathbf{Q}}(s,\mathbf{w}_j)=\arg\max_{a\in\mathcal{A}} \mathbf{w}_j^\top \mathbf{Q}(s,a,\mathbf{w}_j),
\]
consistent with the frozen inner controller introduced in Section~\ref{sec:method}. Each policy corresponds to a particular objective trade-off learned by the inner MORL controller.

We represent each deterministic base policy as a probability distribution whose entries define the policy matrix \(\mathbf{\Pi}^{\mathbf{w}_j}\),
\[
\pi(a\mid s,\mathbf{w}_j)=
\begin{cases}
1, & a=\pi_{\mathbf{Q}}(s,\mathbf{w}_j),\\
0, & \text{otherwise}.
\end{cases}
\]

The outer preference generator selects or weights these policies according to the current state. To characterize the feasible policy space, let \(\mathbf{E}\in\mathbb{R}^{n\times k}\) denote the emotional matrix, whose row
\[
\mathbf{E}_{s,:}=(\mathbf{E}_{s,1},\ldots,\mathbf{E}_{s,k})\in\Delta^{k-1}
\]
is the preference-selection distribution at state \(s\), satisfying
\[
\mathbf{E}_{s,j}\ge 0,\qquad \sum_{j=1}^{k} \mathbf{E}_{s,j}=1.
\]
The resulting composite policy is
\begin{equation}\label{eq:composite_policy_main}
\mathbf{\Pi}^{\mathrm{overall}}(s,a)=\sum_{j=1}^{k} \mathbf{E}_{s,j}\,\pi(a\mid s,\mathbf{w}_j).
\end{equation}
We denote the collection of all such composite policies by \(\mathcal{P}_{\mathrm{mix}}\).

In the actual system, the inner controller can generate infinitely many deterministic policies, and the outer preference generator outputs a deterministic preference (corresponding to a one-hot selection over the base policies). To present the core theory in matrix form, however, we temporarily expand the feasible set of emotion weights to the probability simplex (i.e., allow arbitrary convex combinations), while restricting the analysis to the low-dimensional subspace spanned by the \(k\) representative base policies. This expansion only enlarges the set of feasible policies, so the resulting upper bound on the optimal value still applies to the original one-hot case; it therefore does not affect the conclusions of the theoretical analysis (see Appendix~\ref{thm:onehot_equality} for the inclusion \(\mathcal{P}_{\mathrm{mix}}^{\mathrm{onehot}}\subseteq\mathcal{P}_{\mathrm{mix}}\) and the equivalence of their optimal values).

Clearly,
\[
\mathcal{P}_{\mathrm{mix}}\subseteq \mathcal{P}_{\mathrm{all}}.
\]
This inclusion captures the fundamental role of the inner MORL controller: the outer emotional preference generator cannot create arbitrary low-level behaviors. Instead, it selects among behaviors represented by the inner policy family.

For each state \(s\), the set of action distributions available through the inner policy family is the convex hull
\begin{equation}\label{eq:hull_state_main}
\mathcal{C}_s=\operatorname{conv}\bigl\{\mathbf{\Pi}^{\mathbf{w}_1}(s,:),\ldots,\mathbf{\Pi}^{\mathbf{w}_k}(s,:)\bigr\}.
\end{equation}
Therefore,
\[
\mathbf{\Pi}^{\mathrm{overall}}(s,:)\in\mathcal{C}_s.
\]
The outer learning problem can consequently be interpreted as searching for a high-level state-conditioned selection rule over the collection of inner goal-directed strategies.

\subsubsection{Connection to the continuous preference implementation.}

The practical implementation produces a continuous preference
$\mathbf{w}=\mathbf{e}_\theta(s)\in\Delta^{m-1}$, whereas the finite-policy analysis
uses a set of representative preferences
$\{\mathbf{w}_1,\ldots,\mathbf{w}_k\}$. The theoretical analysis should therefore be
interpreted as a discretized characterization of the policy manifold
induced by the continuous preference-conditioned controller. In
particular, each representative preference $\mathbf{w}_j$ induces a base policy
$\pi_{\mathbf{w}_j}$, and the finite policy class approximates the image of the
continuous mapping
\[
\mathbf{w}\mapsto \pi_{\mathbf{w}}.
\]
The representation error therefore contains two components: the error
arising from the finite coverage of the preference-conditioned policy
family and the error induced by approximating the unrestricted optimal
policy with this family.

Define
\[
\Pi_{\text{cont}} = \{ \pi_{\mathbf{w}} : \mathbf{w} \in \Delta_{m-1} \},
\]
It yields
\[
P_{\text{disc}} \subseteq P_{\text{cont}} \subseteq P_{\text{all}}.
\]
\subsection{Restricted Optimality of the Preference-Regulated Policy Space}

For any policy \(\mathbf{\Pi}\), let \(\mathbf{v}^{\mathbf{\Pi}}\in\mathbb{R}^{n}\) denote the outer state-value vector induced by \(\mathbf{\Pi}\), with components
\[
v^{\mathbf{\Pi}}(s)=\mathbb{E}_{\mathbf{\Pi}}\left[\sum_{t=0}^{\infty} \gamma_{\mathrm{out}}^t r_G(s_t,a_t)\mid s_0=s\right].
\]
Define the optimal value over the unrestricted policy space as
\[
\mathbf{v}^{*}_{\mathrm{all}}=\max_{\mathbf{\Pi}\in\mathcal{P}_{\mathrm{all}}} \mathbf{v}^{\mathbf{\Pi}},
\]
and the optimal value attainable by preference regulation as
\[
\mathbf{v}^{*}_{\mathrm{mix}}=\max_{\mathbf{\Pi}\in\mathcal{P}_{\mathrm{mix}}} \mathbf{v}^{\mathbf{\Pi}},
\]
where both maxima are taken componentwise over states.
Because \(\mathcal{P}_{\mathrm{mix}}\subseteq \mathcal{P}_{\mathrm{all}}\), we immediately have the following restricted-optimality result.

\begin{theorem}[Restricted Optimality]\label{thm:restricted_optimality_main}
Under a finite-state, finite-action discounted outer MDP with \(\gamma_{\mathrm{out}}<1\),
\begin{equation}
\mathbf{v}^{*}_{\mathrm{mix}} \preceq \mathbf{v}^{*}_{\mathrm{all}}.
\end{equation}
Moreover, if an optimal unrestricted policy belongs to \(\mathcal{P}_{\mathrm{mix}}\), then
\begin{equation}
\mathbf{v}^{*}_{\mathrm{mix}} = \mathbf{v}^{*}_{\mathrm{all}}.
\end{equation}
\end{theorem}

The proof follows directly from policy-space inclusion; the complete derivation is provided in Theorem~\ref{thm:restricted_optimality}.

This result formalizes an important point about the proposed architecture. Dynamic emotional preference regulation does not inherently introduce an unavoidable performance penalty. A performance gap appears only when the optimal outer policy cannot be represented by the inner preference-conditioned policy family.

We therefore define the optimality gap vector as
\begin{equation}\label{eq:gap_def_main}
\mathbf{\Delta} = \mathbf{v}^{*}_{\mathrm{all}} - \mathbf{v}^{*}_{\mathrm{mix}},
\end{equation}
with
\[
\mathbf{\Delta}\succeq \mathbf{0}.
\]
The gap is thus not fundamentally caused by the existence of an outer emotional preference mechanism. Instead, it is caused by the representational limitation of the policy repertoire available to that mechanism.

\subsection{Representation Error as the Source of the Optimality Gap}

The previous result establishes the existence of a gap but does not explain how large the gap can be. We next characterize it through the representation capacity of the inner policy family.

Let
\[
\mathbf{\Pi}^{\mathrm{ideal}}\in\mathcal{P}_{\mathrm{all}},\qquad \mathbf{v}^{\mathbf{\Pi}^{\mathrm{ideal}}}=\mathbf{v}^{*}_{\mathrm{all}},
\]
denote an optimal unrestricted policy. At state \(s\), the best policy representable by the inner policy family lies in \(\mathcal{C}_s\). We therefore define the per-state representation error as the distance between \(\mathbf{\Pi}^{\mathrm{ideal}}(s,:)\) and the closest point in \(\mathcal{C}_s\):
\begin{equation}
\epsilon_{\mathrm{rep}}(s)=\inf_{p\in\mathcal{C}_s} \bigl\|\mathbf{\Pi}^{\mathrm{ideal}}(s,:)-p\bigr\|_1.
\end{equation}
The global representation error is
\begin{equation}\label{eq:rep_error_main}
\epsilon_{\mathrm{rep}}=\sup_{s\in\mathcal{S}} \epsilon_{\mathrm{rep}}(s).
\end{equation}
This quantity has an intuitive interpretation: it measures how accurately the available inner policy repertoire can represent the optimal outer action distribution at every state.

\begin{theorem}[Optimality Gap Bound]\label{thm:gap_bound_main}
Under the same finite-state, finite-action discounted MDP assumptions,
\begin{equation}\label{eq:gap_bound_main}
\|\mathbf{\Delta}\|_\infty \le \frac{\gamma_{\mathrm{out}}}{1-\gamma_{\mathrm{out}}}\, \|\mathbf{P}^{\mathrm{phy}}\|_\infty\, \epsilon_{\mathrm{rep}}\, \|\mathbf{v}^{*}_{\mathrm{all}}\|_\infty,
\end{equation}
where \(\mathbf{P}^{\mathrm{phy}}\in\mathbb{R}^{n|\mathcal{A}|\times n}\) denotes the environment's physical state-transition matrix.
\end{theorem}

The complete proof is given in Theorem~\ref{thm:gap_rep_bound}. The derivation follows from the closed-form value representation and a Neumann-series bound on the inverse Bellman operator.

Equation~\eqref{eq:gap_bound_main} establishes a direct connection between policy representation capacity and outer-task performance.

In particular,
\[
\epsilon_{\mathrm{rep}}\to 0 \implies \|\mathbf{\Delta}\|_\infty \to 0.
\]
Therefore, increasing the coverage of the inner preference-conditioned policy family can in principle make the outer preference-regulated policy arbitrarily close to the unrestricted optimum, subject to the assumptions of the theorem.

\textbf{Long-horizon amplification:}
The factor
\[
\frac{\gamma_{\mathrm{out}}}{1-\gamma_{\mathrm{out}}}
\]
shows that local representation errors can be amplified by long-horizon Bellman recursion. When \(\gamma_{\mathrm{out}}\to 1\), the multiplier becomes large. Consequently, long-horizon tasks place stronger requirements on the completeness of the inner behavioral repertoire.

\subsection{Deterministic Preference and One-Hot Policy Selection}

The implemented emotional preference generator outputs a deterministic preference vector
\[
\mathbf{w}=\mathbf{e}_{\theta}(s)
\]
rather than an explicit probability distribution over a finite set of base policies. To relate the practical implementation to the matrix formulation above, consider first a discrete preference set and define the one-hot policy space
\[
\mathcal{P}_{\mathrm{mix}}^{\mathrm{onehot}}\subseteq \mathcal{P}_{\mathrm{mix}}.
\]
The corresponding representation error is
\begin{equation}
\epsilon_{\mathrm{rep}}^{\mathrm{onehot}}=\sup_{s\in\mathcal{S}} \min_{j=1,\ldots,k} \bigl\|\mathbf{\Pi}^{\mathrm{ideal}}(s,:)-\mathbf{\Pi}^{\mathbf{w}_j}(s,:)\bigr\|_1.
\end{equation}
Because the vertices of a convex hull are a subset of the hull itself,
\begin{equation}
\epsilon_{\mathrm{rep}}^{\mathrm{onehot}}\ge \epsilon_{\mathrm{rep}}.
\end{equation}
Thus, allowing convex combinations can only enlarge the representable policy class.

However, under the finite discounted MDP setting with state-dependent selection, the optimal policy in the mixed policy space can be attained by a deterministic selection of a base policy at each state. The reason is that the Bellman objective is linear in the state-wise mixture coefficients, and a linear function over a simplex attains its maximum at an extreme point. Consequently, the optimal value attained under the one-hot constraint coincides with that of the mixed policy space,
\begin{equation}
\mathbf{v}^{*}_{\mathrm{onehot}} = \mathbf{v}^{*}_{\mathrm{mix}} \;\preceq\; \mathbf{v}^{*}_{\mathrm{all}},
\end{equation}
i.e., a deterministic preference output already suffices to achieve the optimum of the mixed policy space. The corresponding results are established in Theorems~\ref{thm:existance mix} and~\ref{thm:onehot_equality}.

This distinction is useful for interpreting the learned preference vector. A continuous preference output does not necessarily imply that the optimal outer controller needs to randomize among policies. Rather, the continuous preference space provides a convenient parameterization for navigating the family of preference-conditioned behaviors, while an optimal solution may correspond to a state-dependent selection among particular behavioral modes.

\subsection{Zero-Gap Condition}

The previous analysis yields an immediate sufficient condition under which the outer emotional preference mechanism can recover the unrestricted optimum.

\begin{corollary}[Sufficient Condition for Zero Optimality Gap]
If, for every state \(s\),
\begin{equation}
\mathbf{\Pi}^{\mathrm{ideal}}(s,:)\in\mathcal{C}_s,
\end{equation}
then
\[
\epsilon_{\mathrm{rep}}=0
\]
and consequently
\begin{equation}
\mathbf{\Delta}=\mathbf{0}.
\end{equation}
\end{corollary}

The proof follows directly from Theorem~\ref{thm:gap_bound_main}.

This result provides a precise interpretation of the inner controller's role. The objective of the inner MORL stage is not merely to obtain good performance under fixed preferences. Its more fundamental role is to construct a behavioral basis sufficiently rich to support the high-level preference-regulation problem.

The outer emotional preference generator can then search this behavioral basis according to the current state and high-level goal.

\subsection{Skill Inheritance and State-Dependent Behavioral Reorganization}

The policy-space analysis explains the performance limitation of outer preference regulation. We next characterize what kinds of behaviors can be inherited by the outer process.

Let \(\mathcal{S}_{\mathrm{act}}\subseteq\mathcal{S}\) be a subset of states in which a particular action \(a_c\in\mathcal{A}\) represents a behavior of interest.

\begin{theorem}[Unconditional Skill Inheritance]\label{thm:skill_inheritance_main}
Suppose that for every base policy \(\mathbf{\pi}^{\mathbf{w}_j}\),
\begin{equation}
\pi_{\mathbf{Q}}(s,\mathbf{w}_j)=a_c,\quad \forall s\in\mathcal{S}_{\mathrm{act}}.
\end{equation}
Then any preference-regulated composite policy induced by
$
\mathbf{w}=\mathbf{e}_{\theta}(s)
$also selects \(a_c\) on \(\mathcal{S}_{\mathrm{act}}\).
\end{theorem}

\begin{proof}
By hypothesis, every base policy places all probability mass on \(a_c\) at states in \(\mathcal{S}_{\mathrm{act}}\), i.e., \(\pi(a_c\mid s,\mathbf{w}_j)=1\) and \(\pi(a\mid s,\mathbf{w}_j)=0\) for \(a\neq a_c\). Hence any composite policy
\[
\mathbf{\Pi}^{\mathrm{overall}}(s,a)=\sum_{j=1}^{k}\mathbf{E}_{s,j}\,\pi(a\mid s,\mathbf{w}_j)
\]
satisfies \(\mathbf{\Pi}^{\mathrm{overall}}(s,a_c)=\sum_{j=1}^{k}\mathbf{E}_{s,j}=1\) and \(\mathbf{\Pi}^{\mathrm{overall}}(s,a)=0\) for \(a\neq a_c\). Thus the composite policy selects \(a_c\) on \(\mathcal{S}_{\mathrm{act}}\).
\end{proof}

In other words, a behavior that is consistently optimal across the inner policy repertoire is unconditionally inherited by the outer preference-regulated agent. This explains the trivial inheritance of behaviors such as help-seeking in states where all relevant inner policies select the same action.

\begin{remark}[State-Dependent Skill Composition]
Suppose instead that the behavior \(a_c\) is available only under some preferences, i.e., there exists \(j\) with \(\pi_{\mathbf{Q}}(s,\mathbf{w}_j)=a_c\) on \(\mathcal{S}_{\mathrm{act}}\), while other preferences induce different actions. Then the outer preference generator can selectively activate \(a_c\) by producing a preference in the corresponding region of the preference space. When the outer loop uses deep deterministic policy gradient to maximize the outer return, the update gradient of the outer network is
\begin{equation}\label{eq:ddpg_gradient_main}
\nabla_{\theta}J=\mathbb{E}\left[\nabla_{\theta}\mathbf{e}_{\theta}(s)\,\nabla_{\mathbf{w}}Q_{\psi}^{\mathrm{out}}(s,\mathbf{w})\big|_{\mathbf{w}=\mathbf{e}_{\theta}(s)}\right],
\end{equation}
where \(Q_{\psi}^{\mathrm{out}}(s,\mathbf{w})\) is the outer Critic's evaluation of the state--preference pair, which approximates via temporal difference learning \(Q^{\mathrm{out}}(s,\mathbf{w})\approx\mathbb{E}\big[\sum_t \gamma_{\mathrm{out}}^t r^{\mathrm{out}}_t\mid s_0=s,\mathbf{w}_0=\mathbf{w}\big]\). If executing \(a_c\) on \(\mathcal{S}_{\mathrm{act}}\) does not reduce the outer return, then \(Q_{\psi}^{\mathrm{out}}(s,\mathbf{w}_j)\) attains a relatively high value, and gradient ascent drives the outer network to output preferences sufficiently close to \(\mathbf{w}_j\) on \(\mathcal{S}_{\mathrm{act}}\), thereby causing the composite policy to \emph{emerge} the \(a_c\) behavior. This should be distinguished from the creation of a new primitive skill: the outer controller does not invent an action unavailable to the inner controller. Rather, it learns when a previously acquired behavior should become behaviorally prioritized.
\end{remark}

\subsection{Interpretation as Emotional Preference Regulation}

The preceding results provide a theoretical interpretation of the empirical phenomenon observed in Section~\ref{sec:experiments}.

The inner MORL controller establishes a repertoire of competing goal-directed strategies,
\[
\mathcal{P}_{\mathrm{in}}=\{\mathbf{\Pi}^{\mathbf{w}_1},\ldots,\mathbf{\Pi}^{\mathbf{w}_k}\}.
\]
The outer preference generator establishes a state-dependent regulatory function,
$
\mathbf{e}_{\theta}: \mathcal{S}\to\Omega.
$
The resulting overall behavior is therefore the computational cycle shown in Figure~\ref{fig:goalcycle}.

The theoretical results imply that this regulation has three properties.

First, preference regulation is representation-limited. The outer mechanism can only activate strategies represented by the inner policy repertoire. Its performance gap is therefore controlled by \(\epsilon_{\mathrm{rep}}\).

Second, useful behaviors can be preserved. If a behavior is represented across the inner policy family, it is inherited automatically. The outer optimization therefore does not fundamentally require relearning such behavior.

Third, context-dependent behavior can be composed from existing strategies. When different preferences induce different strategies, the outer function \(\mathbf{e}_{\theta}(s)\) can learn to activate different parts of the behavioral repertoire in different states.

Together, these properties formalize the computational role assigned to emotional preference in Section~\ref{sec:method}: the emotional preference is not itself a primitive motor skill or a named emotion category. It is a state-dependent regulator of the relative priority of competing goal-directed strategies.

\subsection{Linking the Theory to the Experimental Results}

The theoretical analysis provides a direct interpretation of the performance patterns observed in Section~\ref{sec:experiments}.

First, the inner controller exhibits a broad range of preference-conditioned behaviors, as demonstrated by the systematic achievement-energy trade-off in Figure~\ref{fig:pareto_frontier}. This corresponds to a nontrivial behavioral repertoire \(\mathcal{P}_{\mathrm{in}}\).

Second, the outer emotional preference model learns to select different regions of this repertoire in different states. The preference distributions and representative trajectories in Section~\ref{sec:experiments} demonstrate this state-dependent selection.

Third, the Emotional Preference Model achieves
$
14.97\pm 4.18
$
survival steps, compared with
$
16.00\pm 4.08
$
for the direct Survival Model.

Rather than treating this difference as evidence against emotional preference learning, Theorems~\ref{thm:restricted_optimality_main} and~\ref{thm:gap_bound_main} provide a structural explanation: the direct Survival Model optimizes over a broader policy space, whereas the Emotional Preference Model optimizes within the representation induced by the pretrained MORL controller.

This interpretation leads to a concrete prediction:
\[
\text{richer inner policy repertoire}\ \Rightarrow\ \epsilon_{\mathrm{rep}}\downarrow\ \Rightarrow\ \mathbf{\Delta}\downarrow.
\]
Consequently, the theoretical analysis suggests that improving the coverage and quality of the inner MORL policy family should directly improve the attainable performance of outer emotional preference regulation.

This prediction also suggests a natural direction for future empirical validation: systematically varying the density and coverage of the inner preference-conditioned policy repertoire and measuring whether the resulting optimality gap follows the representation-error bound.

\subsection{Summary of Theoretical Results}

The theoretical analysis establishes the following principles:

\begin{itemize}
    \item \textbf{The source of the performance loss is the representation error.} The optimality gap is controlled by \(\epsilon_{\mathrm{rep}}\), i.e., the worst-case distance between the optimal unrestricted policy and the convex hull of the inner policies over all states. When the inner policy set is sufficiently rich, \(\epsilon_{\mathrm{rep}}\) can approach zero.
    \item \textbf{Amplification effect of the Bellman recursion.} The discount factor \(\gamma_{\mathrm{out}}\) amplifies the per-step representation error by a factor of \(\frac{1}{1-\gamma_{\mathrm{out}}}\); hence long-horizon tasks (\(\gamma_{\mathrm{out}}\) close to \(1\)) impose more stringent requirements on representation accuracy.
    \item \textbf{Zero-gap condition.} If \(\mathbf{\Pi}^{\mathrm{ideal}}(s,:)\in\mathcal{C}_s\) for every state \(s\), i.e., the ideal policy can be perfectly represented by the inner policies, then \(\mathbf{\Delta}=\mathbf{0}\), and the outer constrained optimization attains the global optimum.
    \item \textbf{Skill inheritance and composition.} The composite policy can select among and reorganize behaviors represented by the inner preference-conditioned policy family. Therefore, the outer optimization not only preserves useful skills already acquired by the inner loop (such as help-seeking), but can also smoothly switch and combine these skills across the state space through gradient signals. This is the mathematical foundation for why the emotional preference model, while maintaining a long survival step count, simultaneously gives rise to rich, state-dependent behavioral organization such as fruit-collecting and help-seeking.
\end{itemize}

At the behavioral level, the outer preference generator cannot arbitrarily invent primitive actions, but it can inherit and selectively activate behaviors contained in the inner MORL repertoire. Hence, the central theoretical role of outer reinforcement learning is not to expand the primitive action space, but to learn a state-conditioned organization of competing goal-directed behaviors.

This provides a mathematical foundation for interpreting the learned state-dependent preference function as an emergent emotional preference: the high-level goal determines the long-term optimization criterion, while the learned preference function dynamically regulates which lower-level goal-directed strategy receives priority in the current context.

\section{Limitations}

Our framework provides a computational formulation of emergent emotional preference, but several limitations remain.

\subsection{Limited scope of the emotional construct}

The present work focuses on one functional aspect of emotion: state-dependent regulation of competing goal priorities. This formulation is motivated by the goal-directed theory of emotion and provides an operational computational definition of emotional preference. However, human emotions involve substantially richer phenomena, including appraisal, physiological responses, action tendencies, temporal dynamics, social cognition, and subjective experience. Our model does not attempt to reproduce these components.

Accordingly, the term \emph{emotional preference} in this paper should be understood as a functional computational construct rather than a claim that the agent possesses a complete human-like emotional state. Our contribution is to provide a mechanism through which a goal-directed system can autonomously acquire context-dependent priority regulation.

\subsection{Immediate-state dependence}

The current implementation uses the instantaneous environment state $s_t$ as the input to the preference generator,
$
\mathbf{w}_t = \mathbf{e}_\theta(s_t).
$
It therefore does not explicitly maintain a latent affective state or recurrent memory. Although the experiments reveal temporal persistence of preference segments, such persistence does not by itself establish history-dependent emotional dynamics.

A more expressive formulation would introduce an internal affective state,
\[
z_t = f_\theta(z_{t-1}, s_t),
\]
followed by
\[
\mathbf{w}_t = g_\phi(z_t),
\]
allowing the model to represent path dependence, cumulative appraisal, delayed adaptation, and other forms of temporal emotional dynamics.

\subsection{Discrete and synthetic environments}

The current implementation is evaluated in discrete grid-world environments with discrete action spaces. The basic environment is intentionally simple in order to make the learned preference dynamics interpretable and the inner policy repertoire analyzable. The advanced environment introduces an additional safety objective and stochastic danger, but remains synthetic.

Consequently, the current experiments primarily establish the feasibility of the proposed preference-regulation mechanism rather than its scalability to complex embodied environments. Generalization to continuous control, partially observable environments, and real-world interaction remains an open question.

\subsection{Dependence on the inner policy repertoire}

The theoretical analysis shows that outer preference regulation is constrained by the representation capacity of the inner MORL controller. In particular, the optimality gap is controlled by the policy representation error $\epsilon_{\mathrm{rep}}$.

Thus, the quality of the learned emotional preference cannot be considered independently of the quality and coverage of the underlying behavioral repertoire. An insufficiently expressive inner policy family may prevent the outer controller from realizing otherwise desirable high-level behaviors.

This limitation is also reflected empirically: the Emotional Preference Model achieves lower survival performance than the directly optimized Survival Model, although it outperforms the tested fixed-preference strategies.

\subsection{Simplified outer objective}

The present experiments use long-term survival as the outer objective. This provides a clear high-level goal and creates meaningful conflicts among energy, achievement, and safety objectives. However, survival is only one possible valued goal.

More complex outer objectives might involve human well-being, as well as the agent's own sense of connection and happiness. In such settings, the outer preference function may need to condition not only on the environmental state but also on persistent values or high-level goals.

A general formulation would therefore be
\[
\mathbf{w}_t = \mathbf{e}_\theta(s_t, G),
\]
or, for multiple high-level values $\mathbf{v}$,
\[
\mathbf{w}_t = \mathbf{e}_\theta(s_t, \mathbf{v}).
\]

\subsection{Optimization considerations}

The current outer learning procedure uses a DDPG-style actor-critic method. The resulting physical action is selected by the preference-conditioned inner controller, which introduces a structured and potentially non-smooth mapping from preference to discrete action. While the method is effective in the current experiments, a systematic comparison with alternative optimization methods for continuous preference spaces is left for future work.

In particular, categorical preference policies, policy-gradient formulations, derivative-free optimization, or differentiable relaxations of the inner action-selection mechanism may provide useful alternatives for more complex environments.

\section{Discussion}

An ethical assessment of the proposed framework and its associated risks is provided in the appendix~\ref{sec:ethical_assessment}.

\subsection{Emotional Preference as a Computational Role Rather Than a Discrete Emotion Label}

The central conceptual contribution of this work is to distinguish emotional preference from a discrete emotion category.

We do not require the model to produce an internal label such as \emph{fear} or \emph{happiness}. Instead, we focus on the functional question of how a goal-directed system regulates the relative priority of competing objectives.

Formally,
$
\mathbf{w}_t = \mathbf{e}_\theta(s_t)
$
represents the current priority structure among lower-level objectives, while the outer objective $G$ defines the long-term criterion according to which this priority structure is learned.

This distinction is important for interpreting the relationship between our computational framework and theories of emotion. Under the goal-directed perspective adopted in this work, the emotional aspect lies in the dynamic regulation of goal priorities within an ongoing goal-directed process, rather than in the existence of a dedicated module corresponding to a named emotion.

This also explains why the same underlying decision machinery can support both ordinary instrumental behavior and emotional preference regulation: what changes is the organization and priority of goals, not necessarily the primitive mechanism used for action selection.

\subsection{Why the Preference Must Be Learned Rather Than Hand-Coded}

A natural alternative to our approach is to manually define a function
\[
\mathbf{w} = f(s).
\]
The experiments demonstrate that simple handcrafted mappings can already improve over some fixed preferences, especially when they encode intuitive resource-dependent rules. However, these rules remain externally specified and require prior knowledge about which environmental variables should determine objective priority.

In contrast, our formulation optimizes $e_\theta$ directly with respect to a high-level goal.

The learned preference therefore becomes an endogenous object of optimization. The agent is not explicitly told that low energy implies an energy preference or that the absence of batteries implies an achievement preference. Instead, these patterns emerge because they improve the outer objective in the experienced environment.

The distinction is therefore not merely between a fixed and a dynamic preference. It is between
\begin{quote}
    \textbf{externally specified preference dynamics}
\end{quote}
and
\begin{quote}
    \textbf{goal-directed learned preference dynamics.}
\end{quote}
This distinction is central to our interpretation of the resulting preference function as emergent.

\subsection{Emotional Preference as Priority Regulation Among Competing Strategies}

The experimental results suggest that the learned preference does not simply respond independently to individual state variables. In the basic environment, the achievement-oriented preference is strongly associated with states in which no further battery can be collected, while other preference states arise under different combinations of environmental conditions.

This is consistent with interpreting the preference function as a mechanism for relative priority regulation.

For example, the same level of energy may lead to different preferences depending on whether:
\begin{itemize}
    \item a battery remains available;
    \item an achievement opportunity is nearby;
    \item the current strategy can still improve survival;
    \item or a previously useful objective has become practically unattainable.
\end{itemize}

Consequently, the learned preference is not simply a direct encoding of a physiological variable. It represents the relative usefulness of competing goal-directed strategies under the current state.

This interpretation is particularly important for the computational theory of emotion adopted in this paper: an emotional preference is meaningful because it changes which goal-directed strategy should currently dominate behavior, not merely because its numerical value varies.

\subsection{Preference Regulation and Behavioral Reorganization}

The theoretical analysis in Section~5 provides a useful interpretation of the empirical behavior.

The inner MORL controller defines a repertoire
$
 \{{\pi}_{\mathbf{w}_1}, \ldots, {\pi}_{\mathbf{w}_k}\},
$
while the outer preference generator learns a state-dependent regulatory function over that repertoire.

Thus, the outer process is best understood as a mechanism for behavioral reorganization:
\[
s_t \to \mathbf{w}_t \to \pi_{\mathbf{w}_t}.
\]

This perspective clarifies the role of skill inheritance. A useful behavior already represented in the inner policy family can be activated whenever its corresponding preference becomes advantageous. The outer process therefore does not need to relearn the behavior from primitive actions.

At the same time, this interpretation places an important constraint on the notion of emergence. The outer controller can reorganize and recombine available behaviors, but it cannot automatically produce arbitrary primitives that are absent from the inner repertoire. The representation-error bound formalizes this limitation.

\subsection{Relationship Between Emotional Preference and Policy Optimality}

The difference between the Emotional Preference Model and the direct Survival Model deserves particular attention.

The direct Survival Model achieves $16.00 \pm 4.08$ survival steps, whereas the Emotional Preference Model achieves $14.97 \pm 4.18$. At first sight, this may suggest that introducing emotional preference regulation is disadvantageous. Our theoretical analysis shows a more nuanced interpretation.

The direct Survival Model searches directly in the physical action space, while the Emotional Preference Model searches through a constrained preference-conditioned behavioral repertoire:
$
\mathcal{P}_{\mathrm{mix}} \subseteq \mathcal{P}_{\mathrm{all}}.
$
The corresponding performance difference is therefore a manifestation of representation error rather than evidence that dynamic preference regulation is intrinsically suboptimal.

In fact, Theorem~\ref{thm:gap_bound_main} shows that if the inner policy family is sufficiently expressive,
$
\epsilon_{\mathrm{rep}} \to 0,
$
then the outer policy can approach the unrestricted optimum.

This suggests a useful design principle:
\begin{quote}
    The scalability of emotional preference regulation depends fundamentally on the coverage and compositional richness of the underlying goal-directed behavioral repertoire.
\end{quote}

\subsection{Toward a Computational Theory of Emotional Preference}

The proposed framework suggests a broader computational formulation, as depicted in Figure~\ref{fig:goalcycle}.

Within this formulation, emotion need not be introduced as an additional reward signal or as a separate action module. Instead, it can be viewed as a mechanism that regulates the priority structure of competing goal-directed processes.

This perspective may help connect several research areas that are often studied separately:
\begin{itemize}
    \item multi-objective reinforcement learning, which studies trade-offs among objectives;
    \item hierarchical reinforcement learning, which studies selection among behavioral components;
    \item computational emotion, which studies the relationship between affect and goal-directed behavior;
    \item and adaptive preference learning, which studies how objective priorities change with context.
\end{itemize}

Our framework provides a common computational interface between these areas:
\[
\text{outer goal+state} \to \text{preference} \to \text{goal-directed strategy}.
\]
The resulting preference vector is both behaviorally functional and semantically interpretable because each dimension corresponds to an explicit objective.

\subsection{Broader Extensions}

The present framework naturally suggests several extensions.

First, high-level goals could themselves become multi-objective. Let $\mathbf{v} \in \Delta^{d-1}$ denote a vector of high-level values, such as self-preservation and human well-being. The preference generator could then be extended to
\[
\mathbf{w}_t = \mathbf{e}_\theta(s_t, \mathbf{v}),
\]
allowing emotional preference regulation to depend jointly on environmental context and persistent values.

Second, a recurrent internal state could be introduced:
\[
z_t = f_\theta(z_{t-1}, s_t), \qquad \mathbf{w}_t = g_\phi(z_t),
\]
allowing the model to capture history-dependent preference persistence and longer-term affective dynamics.

Third, the inner behavioral repertoire could be expanded using richer MORL or skill-learning methods. The theory predicts that broader coverage should reduce representation error and consequently improve the attainable outer-task performance.

Fourth, the preference mechanism could be evaluated in continuous and partially observable environments, where the relationship between state information, latent affective state, and preference regulation becomes substantially richer.

Fifth, the deterministic preference output could be generalized to a probability distribution over preferences. Under the discounted MDP setting with finite states, finite actions, and finite preferences, the deterministic case is shown to attain the same optimal value as sampling from a distribution over preferences; however, in continuous state and action spaces the suitable preferences for a given state may be rich enough to form a distribution, and the output preference can then be drawn as a sample from it. Although the optimality gap for a distribution-valued outer network is discussed in the Appendix, the corresponding experiments remain a valuable direction for future work.

\section{Conclusion}

We introduced a framework for emergent emotional preference generation through outer reinforcement learning. The central idea is to separate the acquisition of goal-directed behaviors from the regulation of their relative priorities. A pretrained MORL controller provides a repertoire of preference-conditioned strategies, while an outer preference generator learns a state-dependent mapping
$
\mathbf{e}_\theta : \mathcal{S} \to \Delta^{m-1}
$
through optimization of a high-level goal.

Our formulation treats the learned preference as an emotional preference in an operational computational sense: it regulates the relative priority of competing lower-level objectives within a goal-directed cycle. This interpretation is grounded in the goal-directed perspective on emotion and does not require the claim that a complete human-like emotional state has emerged.

Experiments demonstrate that the learned preference function develops contextualized and temporally persistent patterns without explicitly specifying the corresponding preference rules. In the basic environment, it dynamically shifts between energy- and achievement-oriented priorities, while in the advanced environment it additionally regulates safety-related priorities. The learned model outperforms the evaluated fixed-preference and handcrafted contextual preference strategies, although direct optimization of the survival objective remains stronger in terms of raw survival performance.

The theoretical analysis explains this performance difference by characterizing the outer controller as an optimization process over a restricted policy space induced by the inner MORL repertoire. We derive an optimality-gap bound governed by policy representation error and establish a sufficient condition for zero gap. We further show that useful behaviors represented in the inner repertoire can be inherited and selectively activated through state-dependent preference regulation.

Taken together, the results support the computational view in Figure~\ref{fig:goalcycle}, in which a high-level goal can induce an endogenous regulation of lower-level goal priorities.

Rather than treating emotion as a predefined module or an additional reward signal, our framework provides a concrete mechanism through which emotional preference can emerge as a learned regulator of competing goal-directed objectives. We hope this perspective can serve as a computational bridge between multi-objective reinforcement learning and theories of emotion grounded in goal-directed behavior.

\bibliographystyle{plain}
\bibliography{mybibfile}

\appendix

\section{Matrix Equations for Outer Emotional Preference RL}
To simplify the analysis, we assume that the preference space is discrete, so as to facilitate a matrix-form description. Based on the Bellman equation for outer RL, we obtain the following three core formulas. The descriptions of the relevant variables are given in the Notation subsection.

\subsection{Three Core Formulas}

\begin{align}
\mathbf{v} 
&= \bigl( \mathbf{I} - \gamma_{\mathrm{out}} \, \bar{\mathbf{\Pi}}^{\mathrm{overall}} \mathbf{P}^{\mathrm{phy}} \bigr)^{-1} \mathbf{r}
\label{eq:v_overall} \\
\mathbf{v} 
&= \tilde{\mathbf{E}} \, \tilde{\mathbf{q}}
\label{eq:v_from_q} \\
\tilde{\mathbf{q}} 
&= \tilde{\mathbf{r}} + \gamma_{\mathrm{out}} \, \mathbf{\Pi}^{\mathrm{stack}} \, \mathbf{P}^{\mathrm{phy}} \, \tilde{\mathbf{E}} \, \tilde{\mathbf{q}}
\label{eq:q_bellman}
\end{align}

where $\bar{\mathbf{\Pi}}^{\mathrm{overall}} := \mathcal{L}(\mathbf{\Pi}^{\mathrm{overall}})$ denotes the lifted form of $\mathbf{\Pi}^{\mathrm{overall}}$.

\subsection{Notation}

\begin{itemize}
    \item $n$: number of states, $|\mathcal{A}|$: number of primitive actions, $k$: number of discrete preferences.

    \item $\mathbf{v} \in \mathbb{R}^{n}$: outer state-value vector, $[\mathbf{v}]_s = V^{\mathrm{out}}(s)$.

    \item $\mathbf{r} \in \mathbb{R}^{n}$: outer immediate reward vector, $[\mathbf{r}]_s = r^{\mathrm{out}}(s)$ (survival reward, depending only on state and not on action).

    \item $\gamma_{\mathrm{out}} \in [0,1)$: outer discount factor.

    \item $\mathbf{P}^{\mathrm{phy}} \in \mathbb{R}^{n|\mathcal{A}| \times n}$: environment's physical state transition matrix, with rows indexed by state-action pairs $(s,a)$, and elements
    \[
    [\mathbf{P}^{\mathrm{phy}}]_{(s,a),s'} = P(s'|s,a).
    \]

    \item $\mathbf{E}\in \mathbb{R}^{n\times k}$: emotional matrix, $\mathbf{E}_{s,j}$ denotes the weight of selecting preference $\mathbf{w}_j$ in state $s$, satisfying
    \[
    \sum_{j=1}^k \mathbf{E}_{s,j} = 1,\quad \mathbf{E}_{s,j}\ge 0.
    \]

    \item $\mathbf{\Pi}^{\mathrm{overall}} \in \mathbb{R}^{n \times |\mathcal{A}|}$: standard-form overall composite policy matrix, with elements
    \[
    [\mathbf{\Pi}^{\mathrm{overall}}]_{s,a}
    =
    \sum_{j=1}^k \mathbf{E}_{s,j}\,\pi(a|s,\mathbf{w}_j).
    \]

    \item $\bar{\mathbf{\Pi}}^{\mathrm{overall}} = \mathcal{L}(\mathbf{\Pi}^{\mathrm{overall}}) \in \mathbb{R}^{n\times n|\mathcal{A}|}$: lifted form of $\mathbf{\Pi}^{\mathrm{overall}}$, defined as
    \[
    [\bar{\mathbf{\Pi}}^{\mathrm{overall}}]_{s,(s',a)}
    =
    \mathbf{1}\{s'=s\}\,\mathbf{\Pi}^{\mathrm{overall}}_{s,a}.
    \]

    \item $\tilde{\mathbf{E}} \in \mathbb{R}^{n \times nk}$: emotion extension matrix, written as
    \[
    \tilde{\mathbf{E}}=
    \begin{pmatrix}
    \operatorname{diag}(\mathbf{e}_{\mathbf{w}_1}) &
    \operatorname{diag}(\mathbf{e}_{\mathbf{w}_2}) &
    \cdots &
    \operatorname{diag}(\mathbf{e}_{\mathbf{w}_k})
    \end{pmatrix},
    \]
    where $\mathbf{e}_{\mathbf{w}_j}\in\mathbb{R}^n$ is the $j$-th column of $\mathbf{E}$.

    \item $\tilde{\mathbf{q}} \in \mathbb{R}^{nk}$: outer action-value vector (state-preference pair values), stacked by preference:
    \[
    \tilde{\mathbf{q}} =
    \begin{pmatrix}
    \mathbf{q}_{\mathbf{w}_1} \\
    \mathbf{q}_{\mathbf{w}_2} \\
    \vdots \\
    \mathbf{q}_{\mathbf{w}_k}
    \end{pmatrix},
    \quad
    \mathbf{q}_{\mathbf{w}_j}\in\mathbb{R}^n,\;
    [\mathbf{q}_{\mathbf{w}_j}]_s = Q^{\mathrm{out}}(s,\mathbf{w}_j).
    \]

    \item $\tilde{\mathbf{r}} \in \mathbb{R}^{nk}$: extended outer reward vector,
    \[
    \tilde{\mathbf{r}} = \mathbf{1}_k \otimes \mathbf{r}
    =
    \begin{pmatrix}
    \mathbf{r} \\
    \mathbf{r} \\
    \vdots \\
    \mathbf{r}
    \end{pmatrix}.
    \]

    \item $\mathbf{\Pi}^{\mathrm{stack}} \in \mathbb{R}^{nk \times n|\mathcal{A}|}$: inner-loop policy stack matrix,
    \[
    \mathbf{\Pi}^{\mathrm{stack}} =
    \begin{pmatrix}
    \bar{\mathbf{\Pi}}^{\mathbf{w}_1} \\
    \bar{\mathbf{\Pi}}^{\mathbf{w}_2} \\
    \vdots \\
    \bar{\mathbf{\Pi}}^{\mathbf{w}_k}
    \end{pmatrix},
    \]
    where each $\bar{\mathbf{\Pi}}^{\mathbf{w}_j}=\mathcal{L}(\mathbf{\Pi}^{\mathbf{w}_j})\in\mathbb{R}^{n\times n|\mathcal{A}|}$ is in lifted form, satisfying
    \[
    [\bar{\mathbf{\Pi}}^{\mathbf{w}_j}]_{s,(s',a)}
    =
    \mathbf{1}\{s'=s\}\,\pi(a|s,\mathbf{w}_j).
    \]
\end{itemize}

\subsection{Key Relationships Among the Variables}

\begin{align}
\tilde{\mathbf{r}} &= \mathbf{1}_k \otimes \mathbf{r} \label{rel_r} \\
\bar{\mathbf{\Pi}}^{\mathrm{overall}} &= \tilde{\mathbf{E}} \, \mathbf{\Pi}^{\mathrm{stack}} \label{rel_pi} \\
\mathbf{v} &= \tilde{\mathbf{E}} \, \tilde{\mathbf{q}} \quad \text{(same as Eq.\eqref{eq:v_from_q})} \notag
\end{align}

\subsection{Simultaneous Solution}

Given the overall policy $\mathbf{\Pi}^{\mathrm{overall}}$ and the inner policy set $\mathbf{\Pi}^{\mathrm{stack}}$, the emotion extension matrix $\tilde{\mathbf{E}}$ (row-wise convex combination) can be determined from Eq.\eqref{rel_pi}; substituting into Eq.\eqref{eq:q_bellman} yields
\[
\tilde{\mathbf{q}}
=
\bigl(
\mathbf{I}
-
\gamma_{\mathrm{out}}
\mathbf{\Pi}^{\mathrm{stack}}
\mathbf{P}^{\mathrm{phy}}
\tilde{\mathbf{E}}
\bigr)^{-1}
\tilde{\mathbf{r}},
\]
which is consistent with Eq.\eqref{eq:v_overall}.

\section{Policy Space Definitions}

\subsection{Full Policy Space}

\begin{equation}
\mathcal{P}_{\mathrm{all}}
=
\left\{
\mathbf{\Pi} \in \mathbb{R}^{n\times |\mathcal{A}|}
\;\middle|\;
\mathbf{\Pi}_{s,:} \in \Delta^{|\mathcal{A}|-1},\ \forall s
\right\}
\end{equation}

\subsection{Mixed Policy Space}

\begin{equation}
\mathcal{P}_{\mathrm{mix}}
=
\left\{
\mathbf{\Pi}^{\mathrm{overall}} \in \mathcal{P}_{\mathrm{all}}
\;\middle|\;
\mathbf{\Pi}^{\mathrm{overall}}_{s,:}
=
\sum_{j=1}^k \mathbf{E}_{s,j}\,\mathbf{\Pi}^{\mathbf{w}_j}_{s,:},\ 
\mathbf{E}_{s,:}\in \Delta^{k-1},\ \forall s
\right\}
\end{equation}

\subsection{Per-State Convex Hull Structure}

\begin{equation}
\mathbf{\Pi}^{\mathrm{overall}}(s,:)
\in
\operatorname{conv}
\left\{
\mathbf{\Pi}^{\mathbf{w}_1}(s,:), \dots, \mathbf{\Pi}^{\mathbf{w}_k}(s,:)
\right\}
\end{equation}

\subsection{Lifted Representation}

\begin{equation}
\bar{\mathbf{\Pi}}^{\mathrm{overall}} = \mathcal{L}(\mathbf{\Pi}^{\mathrm{overall}}),
\qquad
\bar{\mathbf{\Pi}}^{\mathbf{w}_j} = \mathcal{L}(\mathbf{\Pi}^{\mathbf{w}_j})
\end{equation}

\section{Optimization Problem}

\begin{theorem}[Existence of an Optimal Policy]\label{thm:existance all}
Consider a finite-state, finite-action discounted Markov decision process (MDP), where:
\begin{itemize}
    \item The state space $\mathcal{S}$ is finite;
    \item The action space $\mathcal{A}$ is finite;
    \item The discount factor $\gamma\in[0,1)$;
    \item The reward function $r(s,a)$ is defined and takes finite values for every $(s,a)\in\mathcal{S}\times\mathcal{A}$;
    \item The transition probability $P(s'|s,a)$ satisfies $\sum_{s'\in\mathcal{S}}P(s'|s,a)=1$.
\end{itemize}
For any policy $\pi$, let $v^\pi(s)$ be the expected discounted total return starting from state $s$ and acting according to $\pi$, i.e.,
\[
v^\pi(s) = \mathbb{E}_\pi\left[\sum_{t=0}^\infty \gamma^t r(S_t,A_t) \;\middle|\; S_0 = s\right].
\]
Define the pointwise optimal value function: for each $s\in\mathcal{S}$,
\[
v^*(s) = \max_{\pi} v^\pi(s),
\]
where the $\max$ is taken over all possible policies.

Then there exists a deterministic stationary policy $\pi^*$ such that
\[
v^{\pi^*}(s) = v^*(s), \quad \forall s\in\mathcal{S}.
\]
In other words, this single policy simultaneously attains the maximal possible value at every state.

\textbf{Remark:} This theorem is a standard result for finite discounted MDPs; it is stated here solely for the self-containedness of the subsequent derivations and is not a contribution of this paper.
\end{theorem}

\begin{proof}
Step 1: The optimal value function satisfies the Bellman optimality equation.

According to the theory of discounted MDPs, the optimal value function $v^*$ is the unique solution to the following Bellman optimality equation:
\begin{equation}\label{eq:bellman_opt}
v^*(s) = \max_{a\in\mathcal{A}} \left[ r(s,a) + \gamma \sum_{s'\in\mathcal{S}} P(s'|s,a) \, v^*(s') \right], \quad \forall s\in\mathcal{S}.
\end{equation}
This equation holds because, under an optimal action choice, taking an optimal action at the first step and continuing optimally afterwards yields a value equal to the right-hand side maximum. Since $\mathcal{A}$ is finite, the maximum at each state is attainable; i.e., there exists at least one action that maximizes the inner expression.

Step 2: Construct the policy $\pi^*$.

For each state $s\in\mathcal{S}$, pick an action $a^*_s$ that attains the maximum on the right-hand side of Eq. (\ref{eq:bellman_opt}):
\[
a^*_s \in \arg\max_{a\in\mathcal{A}} \left[ r(s,a) + \gamma \sum_{s'\in\mathcal{S}} P(s'|s,a) \, v^*(s') \right].
\]
Define the policy $\pi^*$ to select action $a^*_s$ with probability $1$ at each state $s$, i.e.,
\[
\pi^*(a^*_s \mid s) = 1, \quad \forall s.
\]
Clearly, $\pi^*$ is a deterministic stationary policy.

Step 3: Prove that $v^{\pi^*} = v^*$.

For policy $\pi^*$, its value function $v^{\pi^*}$ satisfies the policy evaluation equation (Bellman expectation equation):
\begin{equation}\label{eq:policy_eval}
v^{\pi^*}(s) = r(s,a^*_s) + \gamma \sum_{s'\in\mathcal{S}} P(s'|s,a^*_s) \, v^{\pi^*}(s'), \quad \forall s\in\mathcal{S}.
\end{equation}
On the other hand, because $a^*_s$ is a maximizing action in the Bellman optimality equation at state $s$, from (\ref{eq:bellman_opt}) we obtain
\begin{equation}\label{eq:opt_at_star}
v^*(s) = r(s,a^*_s) + \gamma \sum_{s'\in\mathcal{S}} P(s'|s,a^*_s) \, v^*(s'), \quad \forall s\in\mathcal{S}.
\end{equation}
Comparing Eq. (\ref{eq:policy_eval}) and Eq. (\ref{eq:opt_at_star}), we see that $v^*$ and $v^{\pi^*}$ satisfy exactly the same system of linear equations. In vector form, this is
\[
\mathbf{v} = \mathbf{r}^{\pi^*} + \gamma \mathbf{P}^{\pi^*} \mathbf{v},
\]
where $\mathbf{r}^{\pi^*}_s = r(s,a^*_s)$ and $\mathbf{P}^{\pi^*}_{s,s'} = P(s'|s,a^*_s)$. The coefficient matrix of this system is $\mathbf{I} - \gamma \mathbf{P}^{\pi^*}$. Since $\gamma\in[0,1)$ and $\mathbf{P}^{\pi^*}$ is a stochastic matrix, $\mathbf{I} - \gamma \mathbf{P}^{\pi^*}$ is invertible (via the Neumann series), so the system has a unique solution. Therefore, we must have
\[
v^{\pi^*} = v^*,
\]
i.e., for every state $s$, $v^{\pi^*}(s) = v^*(s)$.

Step 4: Conclusion.

The constructed deterministic stationary policy $\pi^*$ achieves the pointwise optimal value $v^*(s)$ at every state $s$. This proves the existence of an optimal policy.
\end{proof}

\begin{theorem}[Existence of an Optimal Policy in the Mixed Policy Space]\label{thm:existance mix}
Assume the following conditions:
\begin{itemize}
    \item The underlying MDP has a finite state space $\mathcal{S}$, a finite action space $\mathcal{A}$, and discount factor $\gamma_{\mathrm{out}}\in[0,1)$;
    \item $k$ fixed base policies $\mathbf{\Pi}^{\mathbf{w}_1},\dots,\mathbf{\Pi}^{\mathbf{w}_k}$ are given, each $\mathbf{\Pi}^{\mathbf{w}_j}$ being a stochastic stationary policy on $\mathcal{S}$;
    \item The mixed policy space is defined as
    \[
    \mathcal{P}_{\mathrm{mix}} = \Big\{ \mathbf{\Pi} \in \mathcal{P}_{\mathrm{all}} \;\Big|\;
    \mathbf{\Pi}_{s,:} = \sum_{j=1}^k \mathbf{E}_{s,j} \mathbf{\Pi}^{\mathbf{w}_j}_{s,:},\;
    \mathbf{E}_{s,:}\in\Delta^{k-1},\ \forall s \Big\},
    \]
    where $\Delta^{k-1}$ is the $k$-dimensional probability simplex.
\end{itemize}
For any $\mathbf{\Pi}\in\mathcal{P}_{\mathrm{mix}}$, let $\mathbf{v}^{\mathbf{\Pi}}\in\mathbb{R}^n$ be its state-value vector in the outer MDP (defined by Eq.~\eqref{eq:v_overall}). Define the pointwise maximum as
\[
v^*_{\mathrm{mix}}(s) = \sup_{\mathbf{\Pi}\in\mathcal{P}_{\mathrm{mix}}} v^{\mathbf{\Pi}}(s), \quad \forall s\in\mathcal{S}.
\]
Then there exists a mixed policy $\mathbf{\Pi}^* \in \mathcal{P}_{\mathrm{mix}}$ such that
\[
v^{\mathbf{\Pi}^*}(s) = v^*_{\mathrm{mix}}(s), \quad \forall s\in\mathcal{S},
\]
and the supremum is actually a maximum. In particular, $\mathbf{\Pi}^*$ can be realized by deterministically selecting a base policy at each state (i.e., the emotional matrix rows $\mathbf{E}_{s,:}$ are one-hot vectors).
\end{theorem}

\begin{proof}
Step 1: Transform into an equivalent high-level action MDP.

In the mixed policy space, each $\mathbf{\Pi}\in\mathcal{P}_{\mathrm{mix}}$ is completely characterized by the emotional matrix $\mathbf{E}$, where each row $\boldsymbol{\alpha}_s = \mathbf{E}_{s,:} = (\alpha_{s,1},\dots,\alpha_{s,k})^\top$ belongs to the probability simplex $\Delta^{k-1}$. View this vector as a ``high-level action'' at state $s$. The corresponding state transition probabilities are
\[
P(s'|s,\boldsymbol{\alpha}_s) = \sum_{j=1}^k \alpha_{s,j} \, P_j(s'|s),
\quad\text{where } P_j(s'|s) = \sum_{a\in\mathcal{A}} \pi(a|s,\mathbf{w}_j) P(s'|s,a).
\]
The reward depends only on the state: $r(s,\boldsymbol{\alpha}_s) = r^{\mathrm{out}}(s)$. This yields an equivalent MDP with state space $\mathcal{S}$, an action space that is the compact convex set $\Delta^{k-1}$, and transition and reward functions that are continuous (in fact, linear) in the action $\boldsymbol{\alpha}$.

Step 2: Bellman optimality equation and linearity.

Let $\mathbf{v}^*$ be the optimal value function of this equivalent MDP. From discounted MDP theory, $\mathbf{v}^*$ satisfies the Bellman optimality equation (for all $s$):
\begin{equation}\label{eq:be_mix}
v^*(s) = \max_{\boldsymbol{\alpha}\in\Delta^{k-1}} \Big[ r(s) + \gamma_{\mathrm{out}} \sum_{s'} P(s'|s,\boldsymbol{\alpha}) \, v^*(s') \Big].
\end{equation}
Note that the bracketed expression is linear in $\boldsymbol{\alpha}$:
\[
r(s) + \gamma_{\mathrm{out}} \sum_{s'} \Big( \sum_{j=1}^k \alpha_j P_j(s'|s) \Big) v^*(s')
= r(s) + \gamma_{\mathrm{out}} \sum_{j=1}^k \alpha_j \Big( \sum_{s'} P_j(s'|s) v^*(s') \Big).
\]
A linear function on the compact convex set $\Delta^{k-1}$ attains its maximum at an extreme point (vertex). In short, for each $s$, there exists some base policy index $j^*(s)\in\{1,\dots,k\}$ such that choosing $\boldsymbol{\alpha} = \mathbf{e}_{j^*(s)}$ (a unit vector) achieves the maximum.

Step 3: Construct the optimal mixed policy $\mathbf{\Pi}^*$.

For each state $s$, select the above $j^*(s)$ and define the emotional matrix $\mathbf{E}^*$ of the mixed policy $\mathbf{\Pi}^*$ as
\[
\mathbf{E}^*_{s,:} = \mathbf{e}_{j^*(s)}^\top \quad (\text{i.e., } \alpha_{s,j^*(s)}=1,\ \text{all others }0).
\]
Clearly $\mathbf{\Pi}^* \in \mathcal{P}_{\mathrm{mix}}$. Its value vector $\mathbf{v}^{\mathbf{\Pi}^*}$ satisfies the policy evaluation equation:
\begin{equation}\label{eq:pe_mix}
v^{\mathbf{\Pi}^*}(s) = r(s) + \gamma_{\mathrm{out}} \sum_{s'} P(s'|s,\mathbf{e}_{j^*(s)}) \, v^{\mathbf{\Pi}^*}(s'), \quad \forall s.
\end{equation}
Meanwhile, by the choice of $j^*(s)$, the Bellman optimality equation can be written as
\begin{equation}\label{eq:be_mix2}
v^*(s) = r(s) + \gamma_{\mathrm{out}} \sum_{s'} P(s'|s,\mathbf{e}_{j^*(s)}) \, v^*(s'), \quad \forall s.
\end{equation}
Equations (\ref{eq:pe_mix}) and (\ref{eq:be_mix2}) form the same system of linear equations. Because $\gamma_{\mathrm{out}}<1$, the matrix $\mathbf{I} - \gamma_{\mathrm{out}} \mathbf{P}^{\mathbf{\Pi}^*}$ is invertible ($\mathbf{P}^{\mathbf{\Pi}^*}_{s,s'} = P(s'|s,\mathbf{e}_{j^*(s)})$), so the system has a unique solution, implying
\[
\mathbf{v}^{\mathbf{\Pi}^*} = \mathbf{v}^*.
\]

Step 4: Show that $\mathbf{v}^*$ is exactly $v^*_{\mathrm{mix}}$.

Since policies in $\mathcal{P}_{\mathrm{mix}}$ correspond precisely to stationary policies that choose some $\boldsymbol{\alpha}_s$ at each state, $\mathbf{v}^*$ is the pointwise maximum function over $\mathcal{P}_{\mathrm{mix}}$, and it is attained by $\mathbf{\Pi}^*$. Therefore,
\[
v^*_{\mathrm{mix}}(s) = \max_{\mathbf{\Pi}\in\mathcal{P}_{\mathrm{mix}}} v^{\mathbf{\Pi}}(s) = v^*(s) = v^{\mathbf{\Pi}^*}(s), \quad \forall s.
\]
This proves that there exists a mixed policy that simultaneously maximizes the value at all states, and this policy can be implemented by a deterministic base-policy selection (one-hot emotion vector).
\end{proof}

\section{Optimality Theorem}
\begin{theorem}[Restricted Optimality]
\label{thm:restricted_optimality}
Let the state space $\mathcal{S}$ be finite ($|\mathcal{S}|=n$), the action space $\mathcal{A}$ be finite, the discount factor $\gamma_{\mathrm{out}}\in[0,1)$,
and the outer reward $r^{\mathrm{out}}(s)$ depend only on the state. For any stationary policy $\mathbf{\Pi}$, its state-value vector $\mathbf{v}^{\mathbf{\Pi}}\in\mathbb{R}^n$ is given by
$\mathbf{v}^{\mathbf{\Pi}} = (\mathbf{I} - \gamma_{\mathrm{out}}\bar{\mathbf{\Pi}}^{\mathrm{overall}}\mathbf{P}^{\mathrm{phy}})^{-1}\mathbf{r}$.
Let $\mathcal{P}_{\mathrm{all}}$ be the set of all randomized stationary policies,
and $\mathcal{P}_{\mathrm{mix}}$ be the mixed policy space generated from $k$ given base policies $\mathbf{\Pi}^{\mathbf{w}_1},\dots,\mathbf{\Pi}^{\mathbf{w}_k}$ via emotion weights
$\mathbf{E}$ ($\mathbf{E}_{s,:}\in\Delta^{k-1}$) according to $\mathbf{\Pi}^{\mathrm{overall}}_{s,:}=\sum_{j=1}^k\mathbf{E}_{s,j}\mathbf{\Pi}^{\mathbf{w}_j}_{s,:}$.
Define
\[
\mathbf{v}^*_{\mathrm{all}} = \max_{\mathbf{\Pi}\in\mathcal{P}_{\mathrm{all}}} \mathbf{v}^{\mathbf{\Pi}}, \qquad
\mathbf{v}^*_{\mathrm{mix}} = \max_{\mathbf{\Pi}\in\mathcal{P}_{\mathrm{mix}}} \mathbf{v}^{\mathbf{\Pi}},
\]
where $\max$ denotes the componentwise maximum taken state by state (the maxima are attainable in a standard discounted MDP). Then
\begin{enumerate}
    \item $\mathbf{v}^*_{\mathrm{mix}} \preceq \mathbf{v}^*_{\mathrm{all}}$, i.e., for each $s\in\mathcal{S}$, $v^*_{\mathrm{mix}}(s) \le v^*_{\mathrm{all}}(s)$;
    \item If there exists an optimal policy of the full space that belongs to the mixed policy space, i.e., $\mathcal{P}^*_{\mathrm{all}} \cap \mathcal{P}_{\mathrm{mix}} \neq \emptyset$ (where
    $\mathcal{P}^*_{\mathrm{all}} = \{\mathbf{\Pi}\in\mathcal{P}_{\mathrm{all}} \mid \mathbf{v}^{\mathbf{\Pi}} = \mathbf{v}^*_{\mathrm{all}}\}$),
    then $\mathbf{v}^*_{\mathrm{mix}} = \mathbf{v}^*_{\mathrm{all}}$.
\end{enumerate}
\end{theorem}

\begin{proof}
The first inequality follows directly from $\mathcal{P}_{\mathrm{mix}} \subseteq \mathcal{P}_{\mathrm{all}}$:
the componentwise maximum over a smaller set cannot exceed that over a larger set.

Now we prove the second statement. Assume $\mathbf{\Pi}^* \in \mathcal{P}^*_{\mathrm{all}} \cap \mathcal{P}_{\mathrm{mix}}$.
Since $\mathbf{\Pi}^* \in \mathcal{P}^*_{\mathrm{all}}$, we have $\mathbf{v}^{\mathbf{\Pi}^*} = \mathbf{v}^*_{\mathrm{all}}$.
As $\mathbf{\Pi}^* \in \mathcal{P}_{\mathrm{mix}}$, by the definition of $\mathbf{v}^*_{\mathrm{mix}}$ we obtain
$\mathbf{v}^{\mathbf{\Pi}^*} \preceq \mathbf{v}^*_{\mathrm{mix}}$.
Combining with the inequality from the first statement yields
\[
\mathbf{v}^*_{\mathrm{all}} = \mathbf{v}^{\mathbf{\Pi}^*} \preceq \mathbf{v}^*_{\mathrm{mix}} \preceq \mathbf{v}^*_{\mathrm{all}},
\]
hence $\mathbf{v}^*_{\mathrm{mix}} = \mathbf{v}^*_{\mathrm{all}}$.
\end{proof}

\section{Theoretical Analysis of the Gap}

\subsection{Preliminary Lemmas}

\begin{lemma}[Row-Norm Representation of the Difference of Lifted Policies]
\label{lem:lifting_row_norm}
Let the state space $\mathcal{S}$ be finite ($|\mathcal{S}|=n$), the action space $\mathcal{A}$ be finite,
and $\mathbf{\Pi}^{(1)}, \mathbf{\Pi}^{(2)} \in \mathbb{R}^{n \times |\mathcal{A}|}$ be two stationary policies (each row is a probability distribution).
Define the lifting map $\mathcal{L}$ that lifts a policy to a matrix in $\mathbb{R}^{n \times n|\mathcal{A}|}$ as
\[
[\mathcal{L}(\mathbf{\Pi})]_{s,(s',a)} = \mathbf{1}\{s' = s\}\,\mathbf{\Pi}_{s,a}, \qquad \forall s,s'\in\mathcal{S},\; a\in\mathcal{A}.
\]
Then
\[
\|\mathcal{L}(\mathbf{\Pi}^{(1)}) - \mathcal{L}(\mathbf{\Pi}^{(2)})\|_\infty
= \max_{s\in\mathcal{S}} \|\mathbf{\Pi}^{(1)}(s,:) - \mathbf{\Pi}^{(2)}(s,:)\|_1.
\]
\end{lemma}
\begin{proof}
First recall the definition of the matrix $\infty$-norm: for any matrix $\mathbf{A} = (a_{ij})$,
$\|\mathbf{A}\|_\infty = \max_i \sum_j |a_{ij}|$.

Let $\mathbf{D} = \mathcal{L}(\mathbf{\Pi}^{(1)}) - \mathcal{L}(\mathbf{\Pi}^{(2)}) \in \mathbb{R}^{n \times n|\mathcal{A}|}$.
Consider the $s$-th row (row index corresponding to state $s$). According to the definition of the lifted matrix,
when the column index is $(s',a)$, $[\mathcal{L}(\mathbf{\Pi}^{(1)})]_{s,(s',a)} = \mathbf{1}\{s'=s\} \mathbf{\Pi}^{(1)}_{s,a}$,
$[\mathcal{L}(\mathbf{\Pi}^{(2)})]_{s,(s',a)} = \mathbf{1}\{s'=s\} \mathbf{\Pi}^{(2)}_{s,a}$.
Hence
\[
\mathbf{D}_{s,(s',a)} = \mathbf{1}\{s'=s\} (\mathbf{\Pi}^{(1)}_{s,a} - \mathbf{\Pi}^{(2)}_{s,a}).
\]
This shows that the non-zero entries of the $s$-th row can only appear on those actions $a$ for which the column index satisfies $s'=s$. Summing over these columns gives
\[
\sum_{s'\in\mathcal{S}}\sum_{a\in\mathcal{A}} |\mathbf{D}_{s,(s',a)}|
= \sum_{a\in\mathcal{A}} |\mathbf{D}_{s,(s,a)}|
= \sum_{a\in\mathcal{A}} |\mathbf{\Pi}^{(1)}_{s,a} - \mathbf{\Pi}^{(2)}_{s,a}|
= \|\mathbf{\Pi}^{(1)}(s,:) - \mathbf{\Pi}^{(2)}(s,:)\|_1.
\]
Finally, taking the maximum over all $s$:
\[
\|\mathbf{D}\|_\infty = \max_{s\in\mathcal{S}} \sum_{s',a} |\mathbf{D}_{s,(s',a)}|
= \max_{s\in\mathcal{S}} \|\mathbf{\Pi}^{(1)}(s,:) - \mathbf{\Pi}^{(2)}(s,:)\|_1.
\]
This completes the proof.
\end{proof}

\begin{lemma}[Upper Bound on the Difference of Transition Operators]
\label{lem:transition_bound}
Let the state space $\mathcal{S}$ be finite ($|\mathcal{S}|=n$), the action space $\mathcal{A}$ be finite,
and the environment's physical transition matrix $\mathbf{P}^{\mathrm{phy}} \in \mathbb{R}^{n|\mathcal{A}| \times n}$ satisfy that each row sums to $1$ and all entries are non-negative.
For any two stationary policies $\mathbf{\Pi}^{(1)}, \mathbf{\Pi}^{(2)} \in \mathbb{R}^{n\times|\mathcal{A}|}$, define the transition operator
\[
\mathbf{P}_{\mathbf{\Pi}} = \mathcal{L}(\mathbf{\Pi})\,\mathbf{P}^{\mathrm{phy}} \in \mathbb{R}^{n\times n}.
\]
Then
\[
\|\mathbf{P}_{\mathbf{\Pi}^{(1)}} - \mathbf{P}_{\mathbf{\Pi}^{(2)}}\|_\infty
\le \|\mathbf{P}^{\mathrm{phy}}\|_\infty \cdot \max_{s\in\mathcal{S}} \|\mathbf{\Pi}^{(1)}(s,:) - \mathbf{\Pi}^{(2)}(s,:)\|_1.
\]
\end{lemma}
\begin{proof}
Subtracting the two operators gives
\[
\mathbf{P}_{\mathbf{\Pi}^{(1)}} - \mathbf{P}_{\mathbf{\Pi}^{(2)}}
= \bigl(\mathcal{L}(\mathbf{\Pi}^{(1)}) - \mathcal{L}(\mathbf{\Pi}^{(2)})\bigr)\,\mathbf{P}^{\mathrm{phy}}.
\]
The induced matrix $\infty$-norm (the row-sum norm) is submultiplicative for compatible matrix products:
$\|\mathbf{A}\mathbf{B}\|_\infty \le \|\mathbf{A}\|_\infty \|\mathbf{B}\|_\infty$.
Therefore,
\[
\|\mathbf{P}_{\mathbf{\Pi}^{(1)}} - \mathbf{P}_{\mathbf{\Pi}^{(2)}}\|_\infty
\le \|\mathcal{L}(\mathbf{\Pi}^{(1)}) - \mathcal{L}(\mathbf{\Pi}^{(2)})\|_\infty \,
\|\mathbf{P}^{\mathrm{phy}}\|_\infty.
\]
Applying Lemma~\ref{lem:lifting_row_norm} to replace the first factor immediately yields
\[
\|\mathbf{P}_{\mathbf{\Pi}^{(1)}} - \mathbf{P}_{\mathbf{\Pi}^{(2)}}\|_\infty
\le \|\mathbf{P}^{\mathrm{phy}}\|_\infty \,
\max_s \|\mathbf{\Pi}^{(1)}(s,:) - \mathbf{\Pi}^{(2)}(s,:)\|_1.
\]
This completes the proof.
\end{proof}

\begin{lemma}[Infinity-Norm Bound on the Resolvent]
\label{lem:resolvent_bound}
Let $\gamma \in [0,1)$ and let $\mathbf{P} \in \mathbb{R}^{n\times n}$ be a row-stochastic matrix (entries are non-negative and $\mathbf{P} \mathbf{1} = \mathbf{1}$). Then
\[
\|(\mathbf{I} - \gamma \mathbf{P})^{-1}\|_\infty \le \frac{1}{1-\gamma}.
\]
\end{lemma}
\begin{proof}
First compute $\|\mathbf{P}\|_\infty$: for a row-stochastic matrix, the sum of each row is $1$, so $\|\mathbf{P}\|_\infty = \max_i \sum_j |P_{ij}| = 1$.
Thus $\|\gamma \mathbf{P}\|_\infty = \gamma \|\mathbf{P}\|_\infty = \gamma < 1$.
By the Neumann series in Banach spaces, when $\|\gamma \mathbf{P}\|_\infty < 1$,
$(\mathbf{I} - \gamma\mathbf{P})^{-1}$ exists and can be expanded as an absolutely convergent series:
\[
(\mathbf{I} - \gamma\mathbf{P})^{-1} = \sum_{t=0}^\infty (\gamma \mathbf{P})^t.
\]
Taking the infinity norm on both sides and using the triangle inequality and submultiplicativity of the norm,
\begin{align*}
\|(\mathbf{I} - \gamma\mathbf{P})^{-1}\|_\infty
&= \Big\|\sum_{t=0}^\infty (\gamma \mathbf{P})^t\Big\|_\infty \\
&\le \sum_{t=0}^\infty \|(\gamma \mathbf{P})^t\|_\infty \\
&\le \sum_{t=0}^\infty (\|\gamma \mathbf{P}\|_\infty)^t
= \sum_{t=0}^\infty \gamma^t
= \frac{1}{1-\gamma}.
\end{align*}
This completes the proof.
\end{proof}

\subsection{Exact Expression for the Optimality Gap}

\begin{theorem}[Exact Expression for the Optimality Gap]
\label{thm:gap_exact}
Assume that
\begin{itemize}
    \item the state space $\mathcal{S}$ is finite ($|\mathcal{S}|=n$), the action space $\mathcal{A}$ is finite, and the discount factor $\gamma_{\mathrm{out}} \in [0,1)$;
    \item the reward depends only on the state, denoted by $\mathbf{r} \in \mathbb{R}^n$;
    \item the full policy space $\mathcal{P}_{\mathrm{all}}$ is the set of all stationary policies;
    \item the mixed policy space $\mathcal{P}_{\mathrm{mix}}$ is formed by $k$ given base policies $\mathbf{\Pi}^{\mathbf{w}_1},\dots,\mathbf{\Pi}^{\mathbf{w}_k}$ together with emotion weights $\mathbf{E}$ (each row belongs to $\Delta^{k-1}$) via $\mathbf{\Pi}^{\mathrm{overall}}_{s,:} = \sum_j \mathbf{E}_{s,j}\mathbf{\Pi}^{\mathbf{w}_j}_{s,:}$;
    \item for any policy $\mathbf{\Pi}$, define the transition operator $\mathbf{P}_{\mathbf{\Pi}} = \mathcal{L}(\mathbf{\Pi})\mathbf{P}^{\mathrm{phy}}$ and the value function $\mathbf{v}(\mathbf{\Pi}) = (\mathbf{I} - \gamma_{\mathrm{out}}\mathbf{P}_{\mathbf{\Pi}})^{-1}\mathbf{r}$;
    \item denote $\mathbf{v}^*_{\mathrm{all}} = \max_{\mathbf{\Pi}\in\mathcal{P}_{\mathrm{all}}} \mathbf{v}(\mathbf{\Pi})$ (componentwise), $\mathbf{v}^*_{\mathrm{mix}} = \max_{\mathbf{\Pi}\in\mathcal{P}_{\mathrm{mix}}} \mathbf{v}(\mathbf{\Pi})$(componentwise), and choose optimal policies
          $\mathbf{\Pi}^{\mathrm{ideal}} \in \mathcal{P}_{\mathrm{all}}$ satisfying $\mathbf{v}(\mathbf{\Pi}^{\mathrm{ideal}}) = \mathbf{v}^*_{\mathrm{all}}$,
          $\mathbf{\Pi}^{\mathrm{mix}} \in \mathcal{P}_{\mathrm{mix}}$ satisfying $\mathbf{v}(\mathbf{\Pi}^{\mathrm{mix}}) = \mathbf{v}^*_{\mathrm{mix}}$(guaranteed by Theorems~\ref{thm:existance all} and~\ref{thm:existance mix});
    \item define the Gap vector $\mathbf{\Delta} = \mathbf{v}(\mathbf{\Pi}^{\mathrm{ideal}}) - \mathbf{v}(\mathbf{\Pi}^{\mathrm{mix}})$.
\end{itemize}
Then
\begin{equation}
\boxed{
\mathbf{\Delta}
= (\mathbf{I} - \gamma_{\mathrm{out}}\mathbf{P}_{\mathbf{\Pi}^{\mathrm{mix}}})^{-1}\,
\gamma_{\mathrm{out}}\,
\bigl( \mathbf{P}_{\mathbf{\Pi}^{\mathrm{ideal}}} - \mathbf{P}_{\mathbf{\Pi}^{\mathrm{mix}}} \bigr)\,
\mathbf{v}(\mathbf{\Pi}^{\mathrm{ideal}})
}.
\label{eq:gap_exact}
\end{equation}
\end{theorem}
\begin{proof}
To simplify notation, let
\[
\mathbf{A} = \mathbf{I} - \gamma_{\mathrm{out}}\mathbf{P}_{\mathbf{\Pi}^{\mathrm{mix}}}, \qquad
\mathbf{B} = \mathbf{I} - \gamma_{\mathrm{out}}\mathbf{P}_{\mathbf{\Pi}^{\mathrm{ideal}}}.
\]
By the condition on the discount factor and the row-stochasticity of $\mathbf{P}_{\mathbf{\Pi}}$, both $\mathbf{A}$ and $\mathbf{B}$ are invertible.
From the definition of the value function we directly obtain
\[
\mathbf{v}(\mathbf{\Pi}^{\mathrm{mix}}) = \mathbf{A}^{-1} \mathbf{r}, \qquad
\mathbf{v}(\mathbf{\Pi}^{\mathrm{ideal}}) = \mathbf{B}^{-1} \mathbf{r}.
\]
Hence
\begin{align*}
\mathbf{\Delta}
&= \mathbf{B}^{-1}\mathbf{r} - \mathbf{A}^{-1}\mathbf{r} \\
&= (\mathbf{B}^{-1} - \mathbf{A}^{-1})\,\mathbf{r}.
\end{align*}

Use the matrix identity
\[
\mathbf{B}^{-1} - \mathbf{A}^{-1}
= \mathbf{A}^{-1} (\mathbf{A} - \mathbf{B}) \mathbf{B}^{-1}.
\]
(One can verify this identity by right-multiplying by $\mathbf{A} + \mathbf{B}$, or start from $\mathbf{A}^{-1} - \mathbf{B}^{-1} = \mathbf{A}^{-1}(\mathbf{B} - \mathbf{A})\mathbf{B}^{-1}$ and adjust signs.)
We adopt the latter:
\[
\mathbf{A}^{-1} - \mathbf{B}^{-1} = \mathbf{A}^{-1}(\mathbf{B} - \mathbf{A})\mathbf{B}^{-1}.
\]
Therefore,
\[
\Delta = \bigl( \mathbf{A}^{-1}(\mathbf{B} - \mathbf{A})\mathbf{B}^{-1} \bigr) \mathbf{r}.
\]

Compute $\mathbf{B} - \mathbf{A}$:
\begin{align*}
\mathbf{B} - \mathbf{A}
&= (\mathbf{I} - \gamma_{\mathrm{out}}\mathbf{P}_{\mathbf{\Pi}^{\mathrm{ideal}}})
   - (\mathbf{I} - \gamma_{\mathrm{out}}\mathbf{P}_{\mathbf{\Pi}^{\mathrm{mix}}}) \\
&= \gamma_{\mathrm{out}} (\mathbf{P}_{\mathbf{\Pi}^{\mathrm{mix}}} - \mathbf{P}_{\mathbf{\Pi}^{\mathrm{ideal}}}).
\end{align*}
Substituting back,
\[
\Delta = \mathbf{A}^{-1} \gamma_{\mathrm{out}}
\bigl( \mathbf{P}_{\mathbf{\Pi}^{\mathrm{mix}}} - \mathbf{P}_{\mathbf{\Pi}^{\mathrm{ideal}}} \bigr)
\mathbf{B}^{-1} \mathbf{r}.
\]
Since $\mathbf{B}^{-1}\mathbf{r} = \mathbf{v}(\mathbf{\Pi}^{\mathrm{ideal}})$, and slightly adjusting signs (factoring out a minus sign), we obtain
\begin{align*}
\Delta
&= \mathbf{A}^{-1} \gamma_{\mathrm{out}}
\bigl( \mathbf{P}_{\mathbf{\Pi}^{\mathrm{mix}}} - \mathbf{P}_{\mathbf{\Pi}^{\mathrm{ideal}}} \bigr)
\mathbf{v}(\mathbf{\Pi}^{\mathrm{ideal}}) \\
&= - \mathbf{A}^{-1} \gamma_{\mathrm{out}}
\bigl( \mathbf{P}_{\mathbf{\Pi}^{\mathrm{ideal}}} - \mathbf{P}_{\mathbf{\Pi}^{\mathrm{mix}}} \bigr)
\mathbf{v}(\mathbf{\Pi}^{\mathrm{ideal}}).
\end{align*}
But note that $\mathbf{\Delta}$ was originally defined as $\mathbf{v}(\mathbf{\Pi}^{\mathrm{ideal}}) - \mathbf{v}(\mathbf{\Pi}^{\mathrm{mix}})$, and the sign will be taken care of automatically when we expand. Let us rewrite the earlier derivation:
\[
\mathbf{v}(\mathbf{\Pi}^{\mathrm{ideal}}) - \mathbf{v}(\mathbf{\Pi}^{\mathrm{mix}})
= \mathbf{B}^{-1}\mathbf{r} - \mathbf{A}^{-1}\mathbf{r}
= (\mathbf{A}^{-1}(\mathbf{A} - \mathbf{B})\mathbf{B}^{-1})\mathbf{r}
= \mathbf{A}^{-1} (\mathbf{A} - \mathbf{B}) \mathbf{B}^{-1}\mathbf{r}.
\]
Since $\mathbf{A} - \mathbf{B} = \gamma_{\mathrm{out}}(\mathbf{P}_{\mathbf{\Pi}^{\mathrm{ideal}}} - \mathbf{P}_{\mathbf{\Pi}^{\mathrm{mix}}})$, we have
\[
\Delta = \mathbf{A}^{-1} \gamma_{\mathrm{out}}
\bigl( \mathbf{P}_{\mathbf{\Pi}^{\mathrm{ideal}}} - \mathbf{P}_{\mathbf{\Pi}^{\mathrm{mix}}} \bigr)
\mathbf{v}(\mathbf{\Pi}^{\mathrm{ideal}}).
\]
Restoring the definition of $\mathbf{A}$ proves Equation~(\ref{eq:gap_exact}).
\end{proof}

\subsection{Upper Bound on the Optimality Gap}

\begin{theorem}[Upper Bound on the Infinity Norm of the Optimality Gap]
\label{thm:gap_bound}
Under exactly the same setting as Theorem~\ref{thm:gap_exact}, we have
\begin{equation}
\|\mathbf{\Delta}\|_\infty
\le \frac{\gamma_{\mathrm{out}}}{1-\gamma_{\mathrm{out}}}\,
\|\mathbf{P}^{\mathrm{phy}}\|_\infty\,
\Bigl( \sup_{s\in\mathcal{S}} \bigl\| \mathbf{\Pi}^{\mathrm{ideal}}(s,:) - \mathbf{\Pi}^{\mathrm{mix}}(s,:) \bigr\|_1 \Bigr)\,
\|\mathbf{v}(\mathbf{\Pi}^{\mathrm{ideal}})\|_\infty .
\label{eq:bound1}
\end{equation}
\end{theorem}
\begin{proof}
Starting from the expression in Theorem~\ref{thm:gap_exact}:
\[
\mathbf{\Delta}
= (\mathbf{I} - \gamma_{\mathrm{out}}\mathbf{P}_{\mathbf{\Pi}^{\mathrm{mix}}})^{-1}\,
\gamma_{\mathrm{out}}\,
\bigl( \mathbf{P}_{\mathbf{\Pi}^{\mathrm{ideal}}} - \mathbf{P}_{\mathbf{\Pi}^{\mathrm{mix}}} \bigr)\,
\mathbf{v}(\mathbf{\Pi}^{\mathrm{ideal}}).
\]
Taking the infinity norm on both sides and using the submultiplicativity of the induced norm,
\begin{align}
\|\mathbf{\Delta}\|_\infty
&\le \bigl\| (\mathbf{I} - \gamma_{\mathrm{out}}\mathbf{P}_{\mathbf{\Pi}^{\mathrm{mix}}})^{-1} \bigr\|_\infty
\cdot \gamma_{\mathrm{out}}
\cdot \bigl\| \mathbf{P}_{\mathbf{\Pi}^{\mathrm{ideal}}} - \mathbf{P}_{\mathbf{\Pi}^{\mathrm{mix}}} \bigr\|_\infty
\cdot \|\mathbf{v}(\mathbf{\Pi}^{\mathrm{ideal}})\|_\infty. \label{eq:step1}
\end{align}

We now bound the three factors on the right-hand side separately.

\textbf{First factor:} Because $\mathbf{P}_{\mathbf{\Pi}^{\mathrm{mix}}}$ is a row-stochastic matrix (it is the product of a lifted policy and the physical transition matrix, and each row sums to $1$) and $\gamma_{\mathrm{out}}\in[0,1)$, Lemma~\ref{lem:resolvent_bound} gives
\[
\| (\mathbf{I} - \gamma_{\mathrm{out}}\mathbf{P}_{\mathbf{\Pi}^{\mathrm{mix}}})^{-1} \|_\infty
\le \frac{1}{1-\gamma_{\mathrm{out}}}.
\]

\textbf{Second factor:} Applying Lemma~\ref{lem:transition_bound}, we obtain
\[
\| \mathbf{P}_{\mathbf{\Pi}^{\mathrm{ideal}}} - \mathbf{P}_{\mathbf{\Pi}^{\mathrm{mix}}} \|_\infty
\le \|\mathbf{P}^{\mathrm{phy}}\|_\infty \cdot
\max_{s} \|\mathbf{\Pi}^{\mathrm{ideal}}(s,:) - \mathbf{\Pi}^{\mathrm{mix}}(s,:)\|_1.
\]
Here $\max_s$ is exactly the supremum (equal on a finite set).

\textbf{Third factor:} It is simply $\|\mathbf{v}(\mathbf{\Pi}^{\mathrm{ideal}})\|_\infty$ itself.

Substituting these two bounds into Equation~(\ref{eq:step1}) yields
\[
\|\mathbf{\Delta}\|_\infty
\le \frac{1}{1-\gamma_{\mathrm{out}}}
\cdot \gamma_{\mathrm{out}}
\cdot \bigl( \|\mathbf{P}^{\mathrm{phy}}\|_\infty
\max_{s} \|\mathbf{\Pi}^{\mathrm{ideal}}(s,:) - \mathbf{\Pi}^{\mathrm{mix}}(s,:)\|_1 \bigr)
\cdot \|\mathbf{v}(\mathbf{\Pi}^{\mathrm{ideal}})\|_\infty.
\]
Rearranging gives exactly Equation~(\ref{eq:bound1}).
\end{proof}

\subsection{Bounding the Gap via the Representation Error}

\begin{definition}[Representation Error and Projected Policy]
\label{def:rep_error}
Under the same setting as Theorem~\ref{thm:gap_exact}, for each state $s\in\mathcal{S}$, define the convex hull spanned by the action distributions of the base policies as
\[
\mathcal{C}_s = \operatorname{conv} \{\mathbf{\Pi}^{\mathbf{w}_1}(s,:),\dots,\mathbf{\Pi}^{\mathbf{w}_k}(s,:)\}.
\]
Define the projected policy $\mathbf{\Pi}^{\mathrm{proj}} \in \mathcal{P}_{\mathrm{mix}}$ as the per-state $\ell_1$ projection of $\mathbf{\Pi}^{\mathrm{ideal}}$:
\[
\mathbf{\Pi}^{\mathrm{proj}}(s,:) \in \arg\min_{\mathbf{p} \in \mathcal{C}_s} \bigl\| \mathbf{\Pi}^{\mathrm{ideal}}(s,:) - \mathbf{p} \bigr\|_1, \quad \forall s\in\mathcal{S}.
\]
(If there are multiple minimizers, pick any one; this does not affect subsequent inequalities.)
The global representation error is defined as
\[
\epsilon_{\mathrm{rep}} = \sup_{s\in\mathcal{S}} \min_{\mathbf{p}\in\mathcal{C}_s} \|\mathbf{\Pi}^{\mathrm{ideal}}(s,:) - \mathbf{p}\|_1
= \sup_{s} \|\mathbf{\Pi}^{\mathrm{ideal}}(s,:) - \mathbf{\Pi}^{\mathrm{proj}}(s,:)\|_1.
\]
\end{definition}

\begin{theorem}[Upper Bound of Optimality Gap via Representation Error]
\label{thm:gap_rep_bound}
Under the setting of Theorem~\ref{thm:gap_exact}, let $\epsilon_{\mathrm{rep}}$ be the above representation error. Then
\begin{equation}
\boxed{
\|\mathbf{\Delta}\|_\infty
\le \frac{\gamma_{\mathrm{out}}}{1-\gamma_{\mathrm{out}}}\,
\|\mathbf{P}^{\mathrm{phy}}\|_\infty\,
\epsilon_{\mathrm{rep}}\,
\|\mathbf{v}(\mathbf{\Pi}^{\mathrm{ideal}})\|_\infty } .
\label{eq:bound_rep}
\end{equation}
\end{theorem}
\begin{proof}
First note that because $\mathbf{\Pi}^{\mathrm{mix}}$ is the policy in $\mathcal{P}_{\mathrm{mix}}$ that simultaneously maximizes the value function at all states,
and $\mathbf{\Pi}^{\mathrm{proj}}$ also belongs to $\mathcal{P}_{\mathrm{mix}}$, we have $\mathbf{v}(\mathbf{\Pi}^{\mathrm{proj}}) \preceq \mathbf{v}(\mathbf{\Pi}^{\mathrm{mix}})$
(componentwise inequality). Hence
\[
\mathbf{\Delta} = \mathbf{v}(\mathbf{\Pi}^{\mathrm{ideal}}) - \mathbf{v}(\mathbf{\Pi}^{\mathrm{mix}})
\preceq \mathbf{v}(\mathbf{\Pi}^{\mathrm{ideal}}) - \mathbf{v}(\mathbf{\Pi}^{\mathrm{proj}}).
\]
By the restricted optimality theorem, $\mathbf{v}(\mathbf{\Pi}^{\mathrm{mix}}) \preceq \mathbf{v}(\mathbf{\Pi}^{\mathrm{ideal}})$, so both sides of the above inequality are componentwise non-negative.
For non-negative vectors, $\mathbf{0} \preceq \mathbf{x} \preceq \mathbf{y}$ implies $\|\mathbf{x}\|_\infty \le \|\mathbf{y}\|_\infty$. Therefore,
\[
\|\mathbf{\Delta}\|_\infty
\le \|\mathbf{v}(\mathbf{\Pi}^{\mathrm{ideal}}) - \mathbf{v}(\mathbf{\Pi}^{\mathrm{proj}})\|_\infty. \label{eq:ineq_to_proj}
\]

Next we estimate the norm on the right-hand side. Following exactly the same derivation as in Theorem~\ref{thm:gap_exact} and Theorem~\ref{thm:gap_bound},
we repeat the process for the policy pair $(\mathbf{\Pi}^{\mathrm{ideal}}, \mathbf{\Pi}^{\mathrm{proj}})$. Since the transition operator corresponding to $\mathbf{\Pi}^{\mathrm{proj}}$ is $\mathbf{P}_{\mathbf{\Pi}^{\mathrm{proj}}}$,
and the value function formula still reads $\mathbf{v}(\mathbf{\Pi}^{\mathrm{proj}}) = (\mathbf{I} - \gamma_{\mathrm{out}}\mathbf{P}_{\mathbf{\Pi}^{\mathrm{proj}}})^{-1}\mathbf{r}$, we have
\begin{align}
&\mathbf{v}(\mathbf{\Pi}^{\mathrm{ideal}}) - \mathbf{v}(\mathbf{\Pi}^{\mathrm{proj}}) \notag \\
&= (\mathbf{I} - \gamma_{\mathrm{out}}\mathbf{P}_{\mathbf{\Pi}^{\mathrm{ideal}}})^{-1}\mathbf{r}
   - (\mathbf{I} - \gamma_{\mathrm{out}}\mathbf{P}_{\mathbf{\Pi}^{\mathrm{proj}}})^{-1}\mathbf{r} \notag \\
&= (\mathbf{I} - \gamma_{\mathrm{out}}\mathbf{P}_{\mathbf{\Pi}^{\mathrm{proj}}})^{-1}
   \gamma_{\mathrm{out}}
   (\mathbf{P}_{\mathbf{\Pi}^{\mathrm{ideal}}} - \mathbf{P}_{\mathbf{\Pi}^{\mathrm{proj}}})
   (\mathbf{I} - \gamma_{\mathrm{out}}\mathbf{P}_{\mathbf{\Pi}^{\mathrm{ideal}}})^{-1}\mathbf{r} \notag \\
&= (\mathbf{I} - \gamma_{\mathrm{out}}\mathbf{P}_{\mathbf{\Pi}^{\mathrm{proj}}})^{-1}
   \gamma_{\mathrm{out}}
   (\mathbf{P}_{\mathbf{\Pi}^{\mathrm{ideal}}} - \mathbf{P}_{\mathbf{\Pi}^{\mathrm{proj}}})
   \mathbf{v}(\mathbf{\Pi}^{\mathrm{ideal}}). \label{eq:gap_proj_exact}
\end{align}
(The derivation is identical to that of Theorem~\ref{thm:gap_exact}, merely replacing $\mathbf{\Pi}^{\mathrm{mix}}$ with $\mathbf{\Pi}^{\mathrm{proj}}$.)

Taking the infinity norm of Equation~(\ref{eq:gap_proj_exact}):
\[
\|\mathbf{v}(\mathbf{\Pi}^{\mathrm{ideal}}) - \mathbf{v}(\mathbf{\Pi}^{\mathrm{proj}})\|_\infty
\le \| (\mathbf{I} - \gamma_{\mathrm{out}}\mathbf{P}_{\mathbf{\Pi}^{\mathrm{proj}}})^{-1} \|_\infty
\cdot \gamma_{\mathrm{out}}
\cdot \| \mathbf{P}_{\mathbf{\Pi}^{\mathrm{ideal}}} - \mathbf{P}_{\mathbf{\Pi}^{\mathrm{proj}}} \|_\infty
\cdot \|\mathbf{v}(\mathbf{\Pi}^{\mathrm{ideal}})\|_\infty.
\]

Apply Lemma~\ref{lem:resolvent_bound} to $\mathbf{P}_{\mathbf{\Pi}^{\mathrm{proj}}}$ (which is also row-stochastic):
\[
\| (\mathbf{I} - \gamma_{\mathrm{out}}\mathbf{P}_{\mathbf{\Pi}^{\mathrm{proj}}})^{-1} \|_\infty \le \frac{1}{1-\gamma_{\mathrm{out}}}.
\]
Apply Lemma~\ref{lem:transition_bound} to the policy pair $(\mathbf{\Pi}^{\mathrm{ideal}}, \mathbf{\Pi}^{\mathrm{proj}})$:
\[
\| \mathbf{P}_{\mathbf{\Pi}^{\mathrm{ideal}}} - \mathbf{P}_{\mathbf{\Pi}^{\mathrm{proj}}} \|_\infty
\le \|\mathbf{P}^{\mathrm{phy}}\|_\infty \cdot
\max_{s} \|\mathbf{\Pi}^{\mathrm{ideal}}(s,:) - \mathbf{\Pi}^{\mathrm{proj}}(s,:)\|_1.
\]
But by Definition~\ref{def:rep_error}, $\max_s \|\mathbf{\Pi}^{\mathrm{ideal}}(s,:) - \mathbf{\Pi}^{\mathrm{proj}}(s,:)\|_1 = \epsilon_{\mathrm{rep}}$.

Combining the above estimates,
\[
\|\mathbf{v}(\mathbf{\Pi}^{\mathrm{ideal}}) - \mathbf{v}(\mathbf{\Pi}^{\mathrm{proj}})\|_\infty
\le \frac{\gamma_{\mathrm{out}}}{1-\gamma_{\mathrm{out}}}\,
\|\mathbf{P}^{\mathrm{phy}}\|_\infty\,
\epsilon_{\mathrm{rep}}\,
\|\mathbf{v}(\mathbf{\Pi}^{\mathrm{ideal}})\|_\infty.
\]
Finally, together with Inequality~(\ref{eq:ineq_to_proj}) this proves Equation~(\ref{eq:bound_rep}).
\end{proof}

\begin{corollary}[Sufficient Condition for Zero Gap]
\label{cor:zero_gap}
Under the same setting as Theorem~\ref{thm:gap_exact}, if for all $s\in\mathcal{S}$, $\mathbf{\Pi}^{\mathrm{ideal}}(s,:) \in \mathcal{C}_s$, then $\mathbf{\Delta} = \mathbf{0}$.
\end{corollary}
\begin{proof}
The condition $\mathbf{\Pi}^{\mathrm{ideal}}(s,:) \in \mathcal{C}_s$ means that at that state the representation error can be made zero, i.e., $\min_{\mathbf{p}\in\mathcal{C}_s} \|\mathbf{\Pi}^{\mathrm{ideal}}(s,:) - \mathbf{p}\|_1 = 0$.
Taking the supremum over all $s$ yields $\epsilon_{\mathrm{rep}} = 0$. Substituting into Theorem~\ref{thm:gap_rep_bound}
gives $\|\mathbf{\Delta}\|_\infty \le 0$, hence $\|\mathbf{\Delta}\|_\infty = 0$, which implies $\mathbf{\Delta} = \mathbf{0}$.
\end{proof}

\section{Policy Classes and Properties under the One-Hot Constraint}

\subsection{One-Hot Constraint and Policy Classes}

\begin{theorem}[Characterization of One-Hot Policy Classes]
\label{thm:onehot_class}
Let the state space $\mathcal{S}$ be finite. Given $k$ stationary base policies 
$\mathbf{\Pi}^{\mathbf{w}_1},\dots,\mathbf{\Pi}^{\mathbf{w}_k}\in\mathcal{P}_{\mathrm{all}}$,
and an emotional matrix $\mathbf{E}\in\mathbb{R}^{n\times k}$ whose rows satisfy the probability simplex constraint $\mathbf{E}_{s,:}\in\Delta^{k-1}$.
If for each state $s$, $\mathbf{E}_{s,:}$ is a one-hot vector (i.e., $\mathbf{E}_{s,:}\in\{\mathbf{e}_1,\dots,\mathbf{e}_k\}$),
then the overall policy constructed by $\mathbf{\Pi}^{\mathrm{overall}}(s,a)=\sum_{j=1}^k \mathbf{E}_{s,j}\,\pi(a|s,\mathbf{w}_j)$ satisfies
\[
\mathbf{\Pi}^{\mathrm{overall}}(s,:) \in 
\{\mathbf{\Pi}^{\mathbf{w}_1}(s,:),\dots,\mathbf{\Pi}^{\mathbf{w}_k}(s,:)\}, \quad \forall s\in\mathcal{S}.
\]
Furthermore, the set of all such policies, called the one-hot policy set, can be expressed as
\[
\mathcal{P}_{\mathrm{mix}}^{\mathrm{onehot}} = 
\prod_{s=1}^{n} \{\mathbf{\Pi}^{\mathbf{w}_j}(s,:)\}_{j=1}^{k}.
\]
\end{theorem}

\begin{proof}
Since $\mathbf{E}_{s,:}$ is a one-hot vector, there exists a unique $j(s)$ such that $\mathbf{E}_{s,j(s)}=1$ and all other components are $0$. Therefore,
\[
\mathbf{\Pi}^{\mathrm{overall}}(s,a) = \sum_{j=1}^k \mathbf{E}_{s,j}\,\pi(a|s,\mathbf{w}_j) = \pi(a|s,\mathbf{w}_{j(s)}),
\]
which means that the action distribution at state $s$ is completely equivalent to that of the base policy $\mathbf{w}_{j(s)}$ at state $s$. This proves the claim.
\end{proof}

\subsection{Inclusion Relations among Policy Classes}

\begin{theorem}[Policy Class Nesting]
\label{thm:class_inclusion}
Let $\mathcal{P}_{\mathrm{all}}$ be the space of all stationary randomized policies,
$\mathcal{P}_{\mathrm{mix}} = \big\{ \mathbf{\Pi}^{\mathrm{overall}} \mid 
\mathbf{\Pi}^{\mathrm{overall}}_{s,:} = \sum_{j=1}^k \mathbf{E}_{s,j} \mathbf{\Pi}^{\mathbf{w}_j}_{s,:},\ \mathbf{E}_{s,:}\in\Delta^{k-1} \big\}$,
and $\mathcal{P}_{\mathrm{mix}}^{\mathrm{onehot}}$ be the set of one-hot policies defined above.
Then the following inclusion relations hold:
\[
\mathcal{P}_{\mathrm{mix}}^{\mathrm{onehot}} \;\subseteq\; \mathcal{P}_{\mathrm{mix}} \;\subseteq\; \mathcal{P}_{\mathrm{all}}.
\]
\end{theorem}

\begin{proof}
One-hot vectors are extreme points of the simplex $\Delta^{k-1}$, so every policy in $\mathcal{P}_{\mathrm{mix}}^{\mathrm{onehot}}$ also belongs to $\mathcal{P}_{\mathrm{mix}}$.
Moreover, since any convex combination of action distributions is still a valid probability distribution, policies in $\mathcal{P}_{\mathrm{mix}}$ are all standard stationary policies, i.e., $\mathcal{P}_{\mathrm{mix}}\subseteq\mathcal{P}_{\mathrm{all}}$.
\end{proof}

\subsection{Equivalence of Optimal Values}

\begin{theorem}[Optimal Values under One-Hot and Convex Combinations are Equal]
\label{thm:onehot_equality}
Consider a discounted MDP with finite states and actions, discount factor $\gamma\in[0,1)$, and a reward function $r(s)$ depending only on the state.
Let $\mathbf{v}^*_{\mathrm{onehot}}$ be the componentwise optimal value function over $\mathcal{P}_{\mathrm{mix}}^{\mathrm{onehot}}$ (taking the supremum for each state),
and $\mathbf{v}^*_{\mathrm{mix}}$ and $\mathbf{v}^*_{\mathrm{all}}$ be those over $\mathcal{P}_{\mathrm{mix}}$ and $\mathcal{P}_{\mathrm{all}}$, respectively.
Then
\[
\mathbf{v}^*_{\mathrm{onehot}} = \mathbf{v}^*_{\mathrm{mix}} \;\preceq\; \mathbf{v}^*_{\mathrm{all}},
\]
and $\mathbf{v}^*_{\mathrm{mix}} = \mathbf{v}^*_{\mathrm{all}}$ if and only if the set of optimal policies in the full space intersects $\mathcal{P}_{\mathrm{mix}}$.
\end{theorem}

\begin{proof}
The inclusion $\mathcal{P}_{\mathrm{mix}}^{\mathrm{onehot}}\subseteq\mathcal{P}_{\mathrm{mix}}$ directly yields $\mathbf{v}^*_{\mathrm{onehot}}\preceq\mathbf{v}^*_{\mathrm{mix}}$.
To prove the reverse inequality, view the policies in $\mathcal{P}_{\mathrm{mix}}$ as an equivalent MDP with emotion weights $\boldsymbol{\alpha}\in\Delta^{k-1}$ as high-level actions (see Theorem~\ref{thm:existance mix} in the previous section).
The right-hand side of the Bellman optimality equation is linear in $\boldsymbol{\alpha}$, and the maximum over a compact set must be attained at an extreme point. Hence, for each state $s$ there exists $j^*(s)$ such that the optimal action is $\mathbf{e}_{j^*(s)}$.
Construct a one-hot policy $\mathbf{\Pi}^*$ according to this choice; its value function and $\mathbf{v}^*_{\mathrm{mix}}$ satisfy the same linear system. By $\gamma<1$ the solution is unique, so $\mathbf{v}^{\mathbf{\Pi}^*} = \mathbf{v}^*_{\mathrm{mix}}$.
Since $\mathbf{v}^*_{\mathrm{onehot}} \succeq \mathbf{v}^{\mathbf{\Pi}^*}$, we obtain $\mathbf{v}^*_{\mathrm{onehot}} = \mathbf{v}^*_{\mathrm{mix}}$.
The inequality $\mathbf{v}^*_{\mathrm{mix}}\preceq\mathbf{v}^*_{\mathrm{all}}$ also follows from the inclusion; the condition for equality is exactly the conclusion of the restricted optimality theorem (equality holds if an optimal policy lies in the mixed space).
\end{proof}

\subsection{Representation Error}

\begin{theorem}[Representation Error Inequality]
\label{thm:rep_error}
Let $\mathbf{\Pi}^{\mathrm{ideal}}\in\mathcal{P}_{\mathrm{all}}$ be a deterministic optimal policy in the full space,
and denote the $L_1$ distance between two probability distributions $p,q$ by $\|p-q\|_1$.
Define
\begin{align*}
\varepsilon_{\mathrm{rep}}^{\mathrm{onehot}} &= 
\max_{s\in\mathcal{S}} \min_{j=1,\dots,k} 
\big\| \mathbf{\Pi}^{\mathrm{ideal}}(s,:) - \mathbf{\Pi}^{\mathbf{w}_j}(s,:) \big\|_1, \\
\varepsilon_{\mathrm{rep}}^{\mathrm{conv}} &= 
\max_{s\in\mathcal{S}} \min_{\alpha\in\Delta^{k-1}} 
\Big\| \mathbf{\Pi}^{\mathrm{ideal}}(s,:) - \sum_{j=1}^{k} \alpha_j \mathbf{\Pi}^{\mathbf{w}_j}(s,:) \Big\|_1.
\end{align*}
Then
\[
\varepsilon_{\mathrm{rep}}^{\mathrm{onehot}} \;\ge\; \varepsilon_{\mathrm{rep}}^{\mathrm{conv}}.
\]
\end{theorem}

\begin{proof}
For a fixed $s$, the set of points $\{\mathbf{\Pi}^{\mathbf{w}_j}(s,:)\}$ is contained in its convex hull
$\operatorname{conv}\{\mathbf{\Pi}^{\mathbf{w}_j}(s,:)\}$.
Therefore, the distance from $\mathbf{\Pi}^{\mathrm{ideal}}(s,:)$ to the convex hull does not exceed its distance to any vertex:
\[
\min_{\alpha} \Big\| \mathbf{\Pi}^{\mathrm{ideal}}(s,:) - \sum_j \alpha_j \mathbf{\Pi}^{\mathbf{w}_j}(s,:) \Big\|_1
\;\le\; \min_j \big\| \mathbf{\Pi}^{\mathrm{ideal}}(s,:) - \mathbf{\Pi}^{\mathbf{w}_j}(s,:) \big\|_1.
\]
Taking the maximum over all $s$ yields the inequality.
\end{proof}

\subsection{Sufficient Condition for Zero Performance Gap}

\begin{theorem}[Sufficient Condition for Zero Gap]
\label{thm:zero_gap_sufficient}
Under the same MDP setting as Theorem~\ref{thm:onehot_equality}, define the performance gap vector 
$\mathbf{\Delta} = \mathbf{v}^*_{\mathrm{all}} - \mathbf{v}^*_{\mathrm{onehot}}$ (componentwise).
If for each state $s$ there exists a base policy index $j(s)$ such that
\[
\mathbf{\Pi}^{\mathrm{ideal}}(s,:) = \mathbf{\Pi}^{\mathbf{w}_{j(s)}}(s,:),
\]
then $\mathbf{\Delta} = \mathbf{0}$.
\end{theorem}

\begin{proof}
Construct a one-hot policy $\mathbf{\Pi}^*$ according to the condition: $\mathbf{\Pi}^*(s,:) = \mathbf{\Pi}^{\mathbf{w}_{j(s)}}(s,:)$.
By construction we immediately have $\mathbf{\Pi}^* = \mathbf{\Pi}^{\mathrm{ideal}}$, so $\mathbf{v}(\mathbf{\Pi}^*) = \mathbf{v}^*_{\mathrm{all}}$.
Moreover, since $\mathbf{v}^*_{\mathrm{onehot}} \succeq \mathbf{v}(\mathbf{\Pi}^*)$ (by definition of optimality) and Theorem~\ref{thm:onehot_equality} gives $\mathbf{v}^*_{\mathrm{onehot}} \preceq \mathbf{v}^*_{\mathrm{all}}$,
combining these yields $\mathbf{v}^*_{\mathrm{onehot}} = \mathbf{v}^*_{\mathrm{all}}$, i.e., $\mathbf{\Delta}=\mathbf{0}$.
\end{proof}

\textbf{Note:} This condition is only sufficient; even if the gap is zero, it is not required that $\mathbf{\Pi}^{\mathrm{ideal}}$ coincides exactly with some base policy at every state. It suffices that there exists some one-hot policy with the same value function.

\section{Outer Network Training Parameters and Network Architecture}

\subsection{Training Hyperparameters}

\begin{table}[htbp]
\centering
\caption{Outer Network Training Hyperparameters}
\label{tab:outer-training-hyperparams}
\begin{tabular}{lc}
\toprule
Parameter & Value \\
\midrule
Outer learning rate (outer\_lr) & $3 \times 10^{-5}$ \\
Soft update coefficient ($\tau_{soft}$) & $0.005$ \\
Outer discount factor ($\gamma_{outer}$) & $0.99$ \\
Experience replay buffer size (mem\_size) & $30000$ \\
Batch size (batch\_size) & $256$ \\
Number of sampled weights (weight\_num) & $32$ \\
\bottomrule
\end{tabular}
\end{table}

The number of training rounds is 20,000 for the basic environment and 160,000 for the advanced environment.

\subsection{Network Architecture}

\subsubsection{Outer Preference Generator}

The outer network is responsible for generating a preference vector $\mathbf{w} \in \Delta^{m-1}$ (a probability distribution on the $n$-dimensional simplex) based on the current state:

\[
\mathbf{e}(s; \theta): \mathcal{S} \rightarrow \Delta^{m-1}
\]

The network is a multi-layer fully connected network:

\[
\begin{aligned}
\mathbf{h}_1 &= \text{ReLU}(\mathbf{W}_1 \mathbf{s} + \mathbf{b}_1), & \mathbf{W}_1 \in \mathbb{R}^{16S \times S} \\
\mathbf{h}_2 &= \text{ReLU}(\mathbf{W}_2 \mathbf{h}_1 + \mathbf{b}_2), & \mathbf{W}_2 \in \mathbb{R}^{32S \times 16S} \\
\mathbf{h}_3 &= \text{ReLU}(\mathbf{W}_3 \mathbf{h}_2 + \mathbf{b}_3), & \mathbf{W}_3 \in \mathbb{R}^{64S \times 32S} \\
\mathbf{h}_4 &= \text{ReLU}(\mathbf{W}_4 \mathbf{h}_3 + \mathbf{b}_4), & \mathbf{W}_4 \in \mathbb{R}^{32S \times 64S} \\
\boldsymbol{\ell} &= \mathbf{W}_5 \mathbf{h}_4 + \mathbf{b}_5, & \mathbf{W}_5 \in \mathbb{R}^{m \times 32S} \\
\mathbf{w} &= \text{Softmax}(\boldsymbol{\ell}), & w_i = \frac{\exp(\ell_i)}{\sum_{j} \exp(\ell_j)}
\end{aligned}
\]

For the basic environment, $S = \dim(\mathcal{S}) = 19$ is the state space dimension, and $m = 2$ is the reward dimension. \textbf{Output}: Preference vector $\mathbf{w} = [w_1, w_2]^{\top}$, satisfying $w_1 + w_2 = 1$ and $w_i \geq 0$.

For the advanced environment, $S = \dim(\mathcal{S}) = 37$ is the state space dimension, and $m = 3$ is the reward dimension. \textbf{Output}: Preference vector $\mathbf{w} = [w_1, w_2, w_3]^{\top}$, satisfying $w_1 + w_2 + w_3 = 1$ and $w_i \geq 0$.


\subsubsection{Outer Critic Network -- Q-value Evaluator}

The outer critic network evaluates the long-term value of state-preference pairs:

\[
Q_\psi^{\mathrm{out}}(s,\mathbf{w}): \mathcal{S} \times \Delta^{m-1} \rightarrow \mathbb{R}
\]

Network structure:

\[
\begin{aligned}
\mathbf{x} &= [s; \mathbf{w}] \in \mathbb{R}^{S+m} \\
\mathbf{h}_1^c &= \text{ReLU}(\mathbf{W}_c^1 \mathbf{x} + \mathbf{b}_c^1), & \mathbf{W}_c^1 \in \mathbb{R}^{16(S+m) \times (S+m)} \\
\mathbf{h}_2^c &= \text{ReLU}(\mathbf{W}_c^2 \mathbf{h}_1^c + \mathbf{b}_c^2), & \mathbf{W}_c^2 \in \mathbb{R}^{32(S+m) \times 16(S+m)} \\
\mathbf{h}_3^c &= \text{ReLU}(\mathbf{W}_c^3 \mathbf{h}_2^c + \mathbf{b}_c^3), & \mathbf{W}_c^3 \in \mathbb{R}^{64(S+m) \times 32(S+m)} \\
\mathbf{h}_4^c &= \text{ReLU}(\mathbf{W}_c^4 \mathbf{h}_3^c + \mathbf{b}_c^4), & \mathbf{W}_c^4 \in \mathbb{R}^{32(S+m) \times 64(S+m)} \\
Q &= \mathbf{W}_c^5 \mathbf{h}_4^c + \mathbf{b}_c^5, & \mathbf{W}_c^5 \in \mathbb{R}^{1 \times 32(S+m)}
\end{aligned}
\]

where $[\cdot; \cdot]$ denotes vector concatenation.

\bigskip
\noindent\textbf{Output}: scalar Q-value $Q_\psi^{\mathrm{out}}(s,\mathbf{w})$


\subsubsection{Inner Q-Network (EnvelopeLinearCQN) -- Vector Q-Function}

The inner network outputs a vector Q-value over the action and reward dimensions:

\[
\text{EnvelopeLinearCQN}(\mathbf{s}, \mathbf{w}; \theta_q): \mathcal{S} \times \Delta^{m-1} \rightarrow \mathbb{R}^{A \times m}
\]

\[
\begin{aligned}
\mathbf{x} &= [s; \mathbf{w}] \in \mathbb{R}^{S+m} \\
\mathbf{h}_1^q &= \text{ReLU}(\mathbf{W}_q^1 \mathbf{x} + \mathbf{b}_q^1) \\
\mathbf{h}_2^q &= \text{ReLU}(\mathbf{W}_q^2 \mathbf{h}_1^q + \mathbf{b}_q^2) \\
\mathbf{h}_3^q &= \text{ReLU}(\mathbf{W}_q^3 \mathbf{h}_2^q + \mathbf{b}_q^3) \\
\mathbf{h}_4^q &= \text{ReLU}(\mathbf{W}_q^4 \mathbf{h}_3^q + \mathbf{b}_q^4) \\
\mathbf{Q} &= \mathbf{W}_q^5 \mathbf{h}_4^q + \mathbf{b}_q^5 \in \mathbb{R}^{A \times m}
\end{aligned}
\]

where $A = 6$ is the size of the action space.

\bigskip
Basic environment \textbf{Output}: Vector Q-matrix $\mathbf{Q} \in \mathbb{R}^{6 \times 2}$

\bigskip
Advanced environment \textbf{Output}: Vector Q-matrix $\mathbf{Q} \in \mathbb{R}^{6 \times 3}$.

\subsection{Overall Training Procedure}
The inner network is pre-trained in advance. 

\textbf{Outer optimization}: the outer network $\theta$ and the outer critic network $\psi$ are optimized using a DDPG-style soft update mechanism.

\bigskip
\noindent\textbf{Target network update} (soft update):
\[
\theta' \leftarrow \tau_{soft} \theta + (1 - \tau_{soft}) \theta'
\]
\[
\psi' \leftarrow \tau_{soft} \psi + (1 - \tau_{soft}) \psi'
\]

\bigskip
\noindent\textbf{Outer reward function}:
\[
r_{outer} = 
\begin{cases}
1 & \text{if the agent survives} \\
0 & \text{if terminated}.
\end{cases}
\]

\section{Basic Environment Experimental Situation}

\subsection{Outer Training Process Plots}
\begin{figure}[htbp]
    \centering
    \begin{subfigure}[b]{0.45\textwidth}
        \centering
        \includegraphics[width=\textwidth]{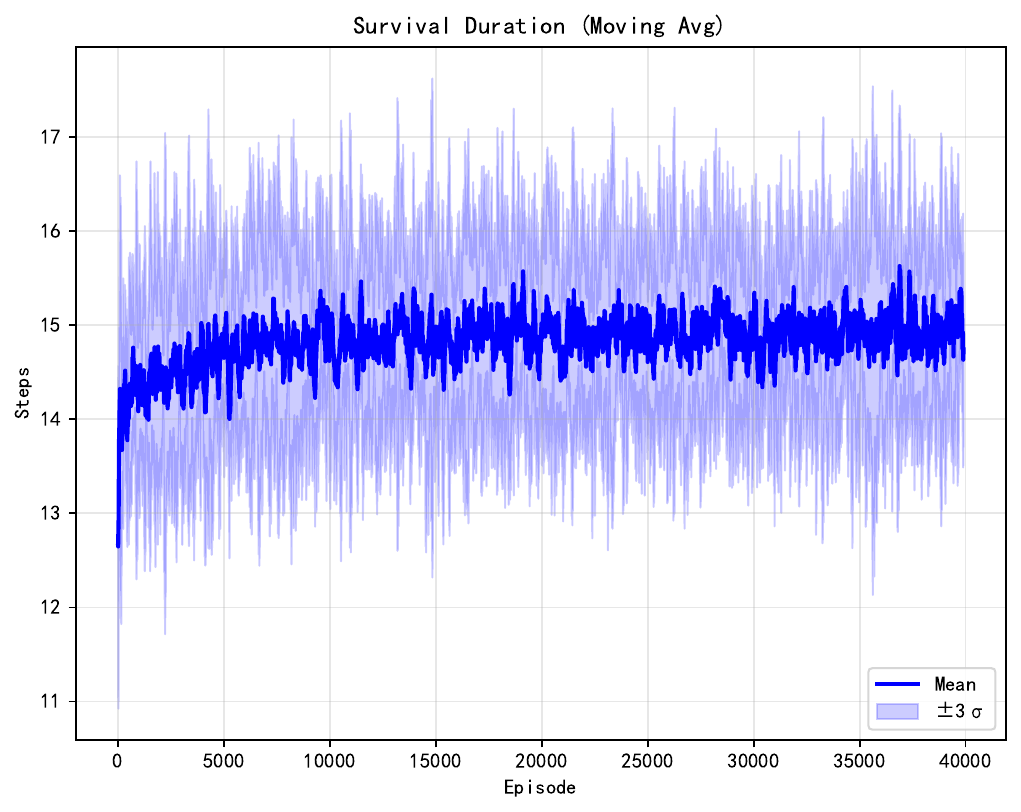}
        \caption{Survival Duration}
        \label{fig:duration}
    \end{subfigure}
    \hfill
    \begin{subfigure}[b]{0.45\textwidth}
        \centering
        \includegraphics[width=\textwidth]{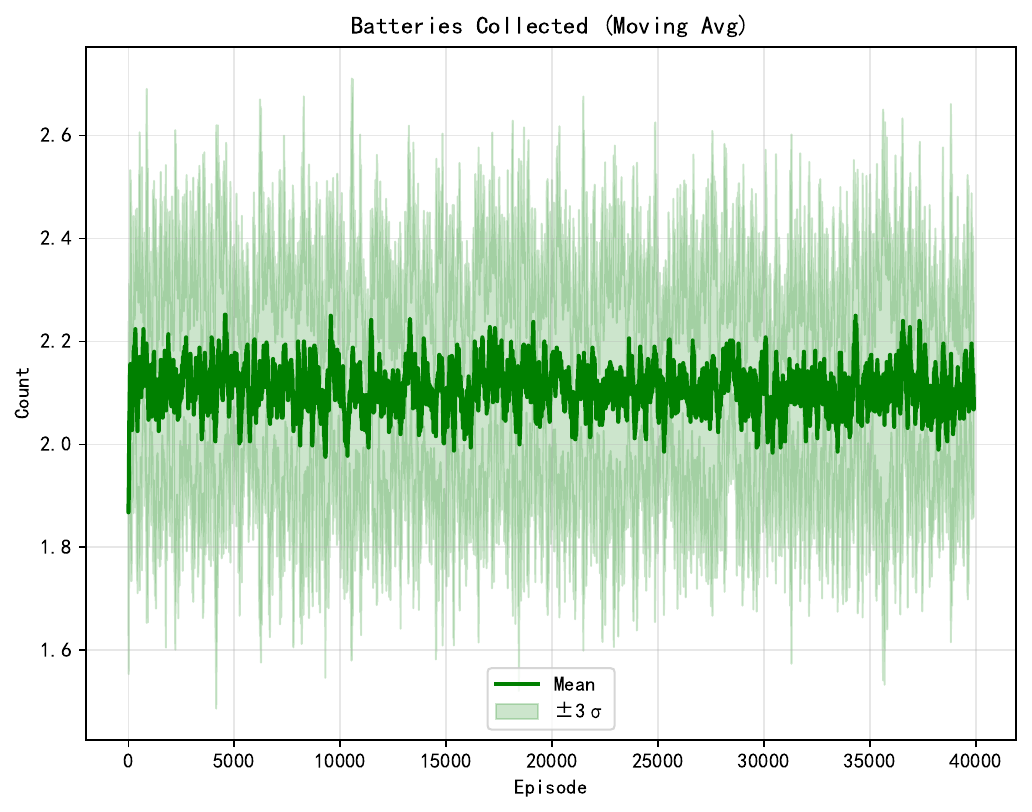}
        \caption{Batteries Collected}
        \label{fig:batteries}
    \end{subfigure}
    \vskip\baselineskip
    \begin{subfigure}[b]{0.45\textwidth}
        \centering
        \includegraphics[width=\textwidth]{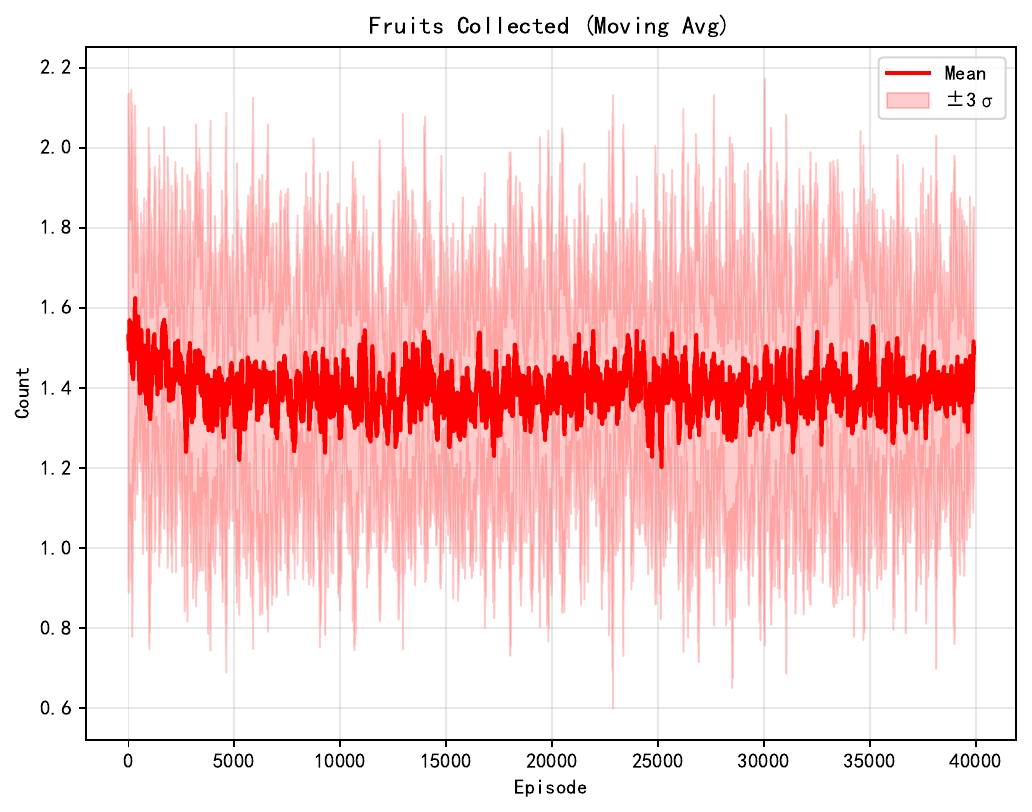}
        \caption{Fruits Collected}
        \label{fig:fruits}
    \end{subfigure}
    \caption{Training Statistics (Mean $\pm$ 3$\sigma$ across rounds). Training Results with Moving Average (window=100). }
    \label{fig:training_stats}
\end{figure}
Figure~\ref{fig:training_stats} shows the relationship between survival time, batteries collected, fruits collected, and the number of training episodes during the outer training process in five randomized experiments. It can be seen that as the number of training episodes increases, the survival time increases and then flattens out, the number of batteries collected increases and then flattens out, and the number of fruits collected decreases and then flattens out.

\section{Advanced Environment Experimental Results}
\label{sec:advanced_environment}

The advanced environment extends the basic setting from two competing
lower-level objectives to three objectives: achievement, energy, and safety.
The purpose of this environment is not to claim that a complete emotional
system emerges in a more complex environment, but to examine whether the
proposed \emph{state-dependent emotional preference regulation} mechanism
remains interpretable when more competing goal-directed objectives are
introduced.

\subsection{Introduction to the Experimental Environment}
\label{sec:advanced_environment_intro}

We consider the \emph{Grid Fruit Tree Battery Hazard (Danger) Probability
Exploration} environment. The environment is a $4\times4$ grid containing
two fruit trees and one battery box. Each fruit tree contains 1--4 fruits,
and the battery box contains 1--3 batteries. The positions of the fruit
trees, battery box, and initial agent position are randomly sampled without
overlap. The agent has six discrete actions:
\[
\mathcal{A}
=
\{\text{up},\text{down},\text{left},\text{right},
\text{collect},\text{help}\}.
\]

The inner multi-objective reward is represented as
\[
\mathbf{r}(s,a)
=
\left[
r_{\mathrm{ach}}(s,a),
r_{\mathrm{ene}}(s,a),
r_{\mathrm{safe}}(s,a)
\right]^{\top},
\]
corresponding to achievement, energy, and safety objectives, respectively.

The agent starts with 9 units of energy. Each resolved action consumes one
unit of energy and produces a corresponding change in the energy-related
reward. If the agent's energy reaches zero, it receives an achievement
penalty of $-10$ and a safety penalty of $-10$, after which the episode
terminates. Selecting the help action terminates the episode after the
corresponding energy consumption. Collecting a fruit yields 5 units of
achievement reward. Collecting a battery restores the agent's energy to
full and produces an energy reward equal to twice the energy increase.

The environment additionally contains a stochastic danger zone. A center
position is randomly generated, subject to the constraint that it does not
overlap with the agent, fruit trees, or the battery box. The danger zone is
defined as the $3\times3$ region centered at this position. If the agent
enters the center cell, it receives a safety reward of $-10$ and the episode
terminates immediately. If it enters another cell in the danger zone, it
receives a safety reward of $-1$ and the episode terminates with probability
$0.5$.

The observation contains the agent position, current and maximum energy,
the number of fruit trees and batteries, the danger-zone center, the
positions and remaining resources of the fruit trees and battery box,
their discovery states, and a $4\times4$ exploration mask. In the current
experiments, the object positions are fully observable. Thus, the advanced
environment is still a fully observable MDP, while providing substantially
richer interactions among objective priorities. A partially observable
version, in which objects are revealed only when the agent approaches them,
is left for future work.

The outer objective remains long-term survival. Therefore, the experiment
tests whether a single high-level goal can induce context-dependent
regulation among three competing lower-level objectives:
\[
\text{achievement},
\qquad
\text{energy},
\qquad
\text{safety}.
\]

This setting provides a more demanding test of the proposed formulation
than the basic environment because different objectives can become
advantageous or disadvantageous under different combinations of resource
availability, energy level, and environmental risk.

\subsection{Outer Training Process}
\label{sec:advanced_training}

\begin{figure}[htbp]
    \centering
    \begin{subfigure}[b]{0.45\textwidth}
        \centering
        \includegraphics[width=\textwidth]{
        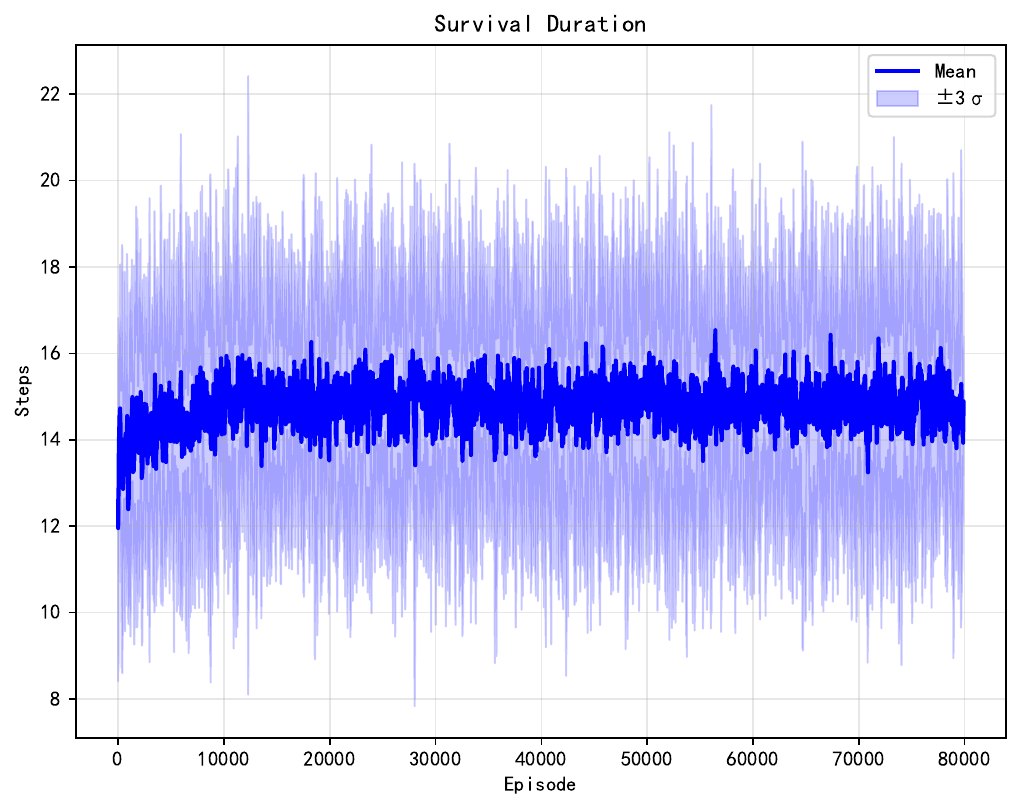}
        \caption{Survival Duration}
        \label{fig:duration2}
    \end{subfigure}
    \hfill
    \begin{subfigure}[b]{0.45\textwidth}
        \centering
        \includegraphics[width=\textwidth]{
        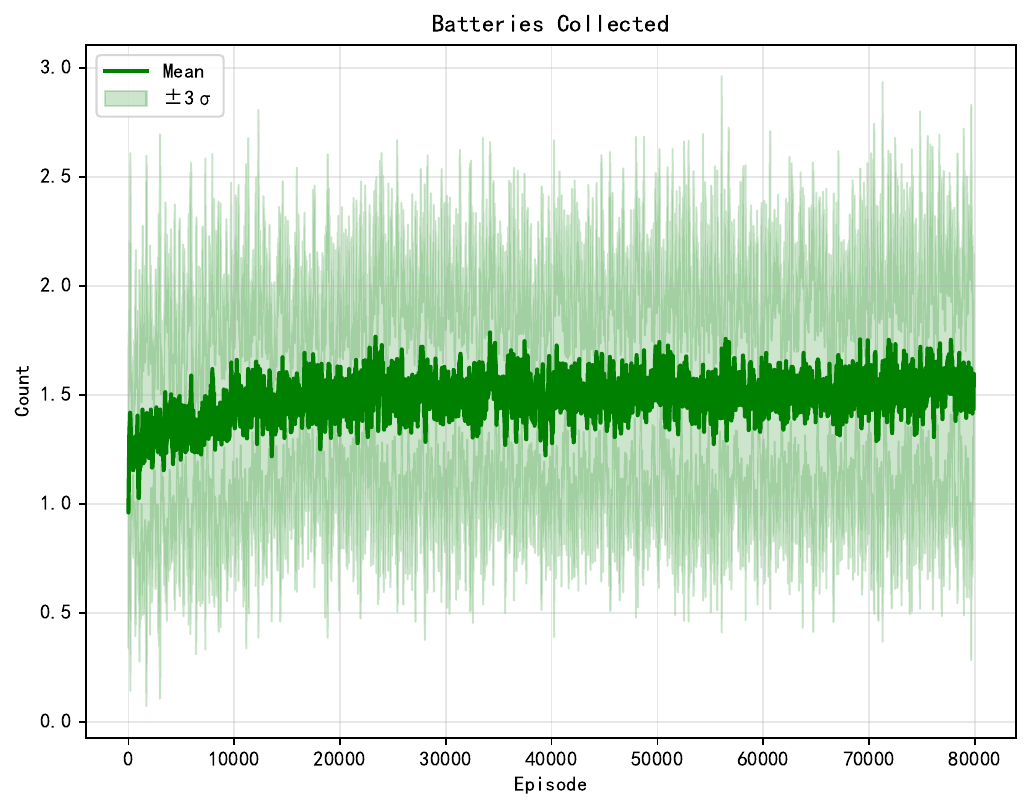}
        \caption{Batteries Collected}
        \label{fig:batteries2}
    \end{subfigure}

    \vskip\baselineskip

    \begin{subfigure}[b]{0.45\textwidth}
        \centering
        \includegraphics[width=\textwidth]{
        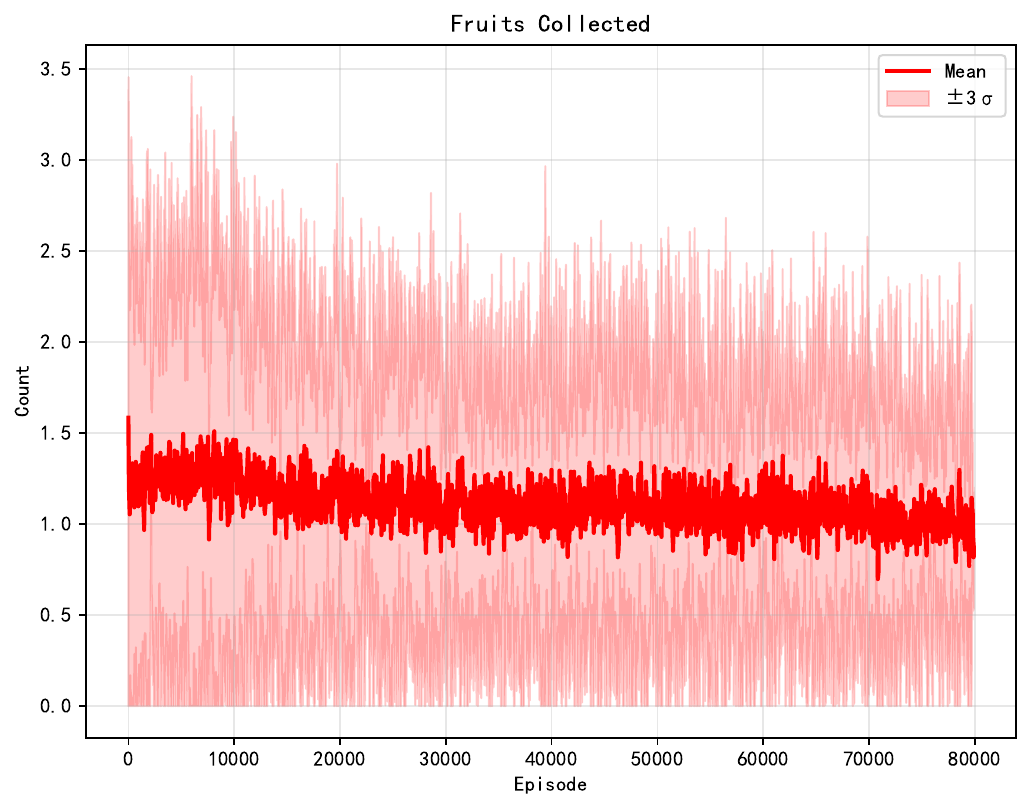}
        \caption{Fruits Collected}
        \label{fig:fruits2}
    \end{subfigure}
    \hfill
    \begin{subfigure}[b]{0.45\textwidth}
        \centering
        \includegraphics[width=\textwidth]{
        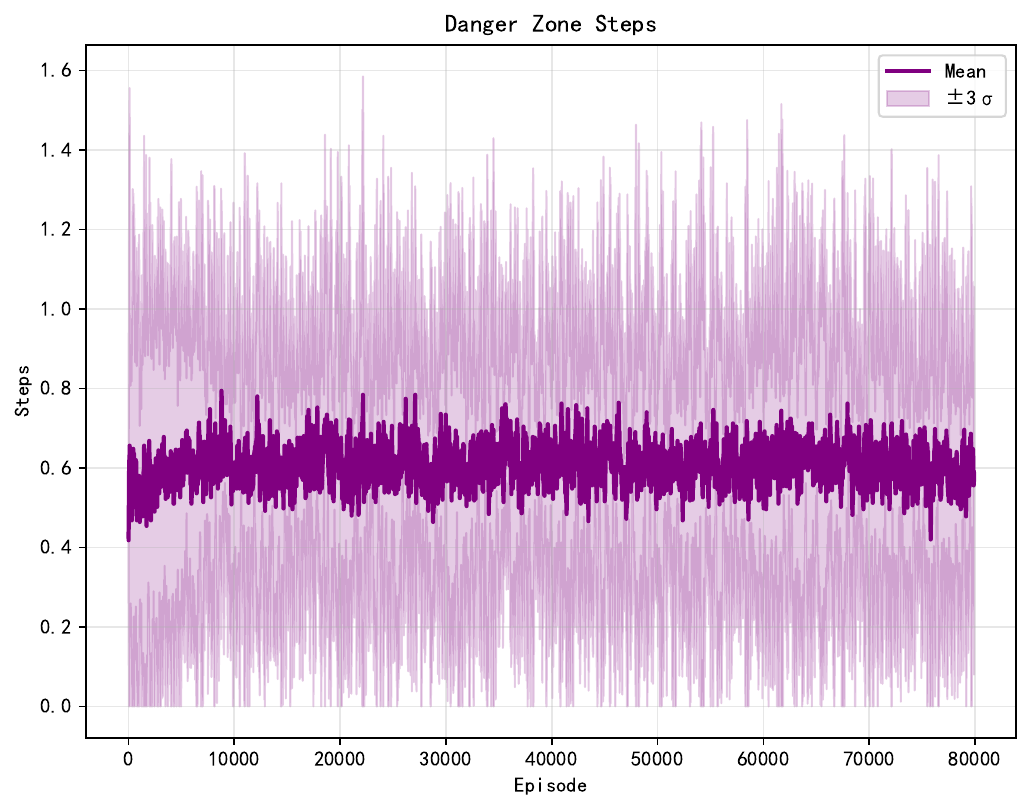}
        \caption{Danger-Zone Steps}
        \label{fig:danger2}
    \end{subfigure}

    \caption{
    Training statistics during outer preference learning in five randomized
    runs. Curves report the mean $\pm 3\sigma$, with a moving-average window
    of 100 episodes.
    }
    \label{fig:training_stats_probability}
\end{figure}

Figure~\ref{fig:training_stats_probability} shows the training dynamics of
the outer preference generator in five randomized experiments. We report
survival duration, the number of collected batteries, the number of
collected fruits, and the number of steps spent in the danger zone.

As training proceeds, survival duration and battery collection initially
increase and subsequently stabilize. Fruit collection decreases during
training, while the number of danger-zone steps increases and eventually
stabilizes. These trends indicate that outer optimization does not simply
maximize one lower-level objective. Instead, it changes the allocation of
behavioral priorities in response to the long-term survival objective.

The increase in danger-zone exposure should not be interpreted as a direct
failure of safety regulation. In this environment, batteries can occur
inside the danger zone, and therefore energy acquisition may sometimes
provide sufficient long-term benefit to justify increased short-term risk.
The relevant quantity learned by the outer controller is consequently not
the independent maximization of safety, but the context-dependent
trade-off among achievement, energy, and safety under the common survival
objective.

\subsection{Performance and Behavior of the Inner MORL Controller}
\label{sec:advanced_inner}

We first evaluate the inner MORL controller independently of the outer
preference generator. Envelope Q-Learning is used to learn the
preference-conditioned vector-valued $Q$ function. During evaluation, a
fixed preference is provided throughout each episode.

\begin{figure}[htb]
  \centering
  \includegraphics[width=\linewidth]{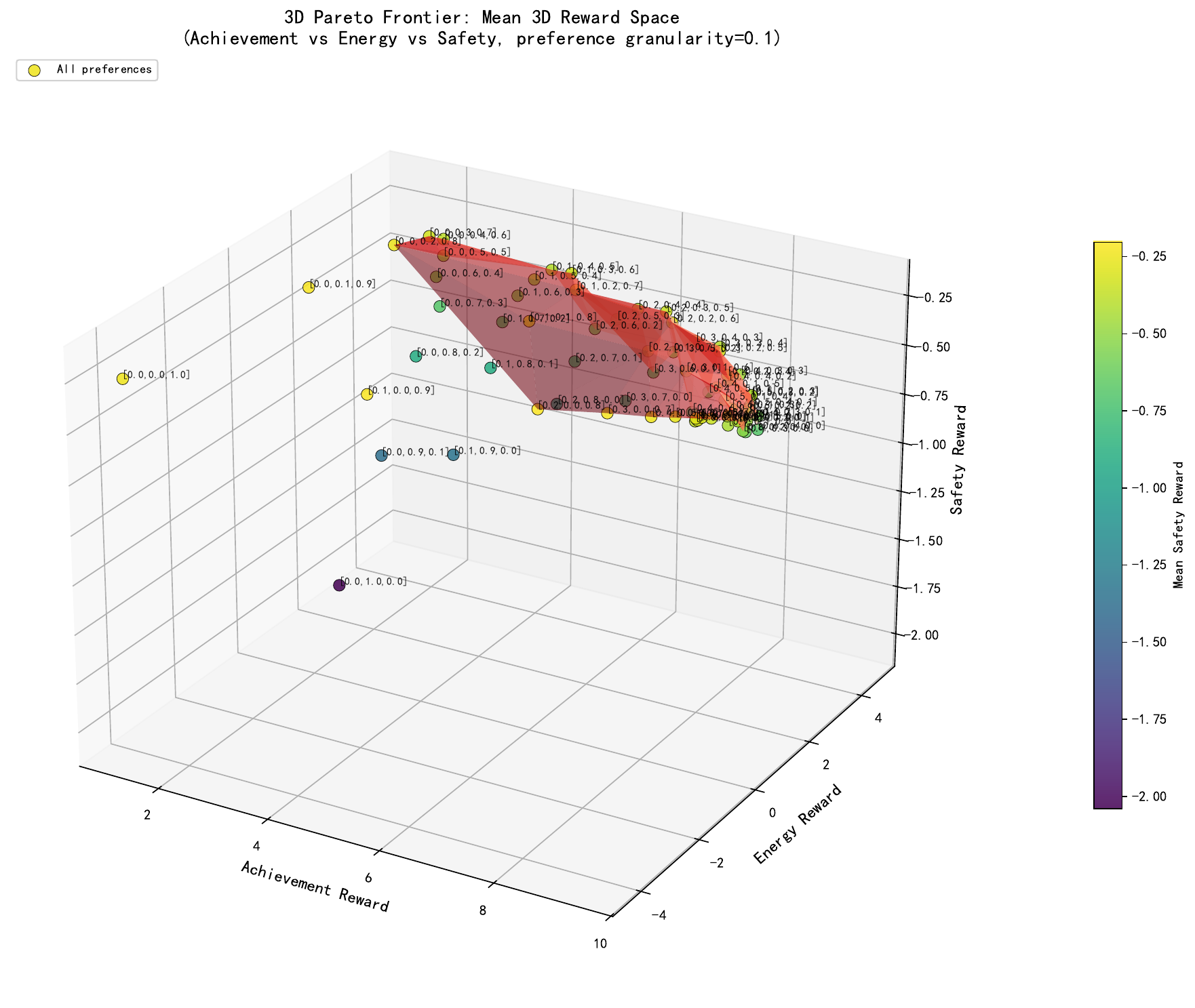}
  \caption{
  Average three-dimensional reward vectors obtained under different fixed
  preferences in the advanced environment. Results are computed over
  20,000 evaluation episodes.
  }
  \label{fig:pareto_frontier_3d}
\end{figure}

\textbf{Reward trade-offs.}
For visualization, we evaluate a representative subset of preferences,
including
\[
(1,0,0)^\top,\quad
(0.9,0.1,0)^\top,\quad
(0.9,0,0.1)^\top,\quad
(0.8,0.2,0)^\top,
\quad \ldots,\quad
(0,0,1)^\top.
\]
Figure~\ref{fig:pareto_frontier_3d} shows the resulting reward vectors in
the three-dimensional achievement--energy--safety space.

The obtained points form a structured trade-off surface, indicating that
the inner controller provides distinguishable goal-directed behaviors for
different objective priorities. This confirms that the inner policy
repertoire contains multiple interpretable strategies that can subsequently
be regulated by the outer preference generator.

\textbf{Survival behavior.}
According to Table~\ref{tab:results_multi}, preferences emphasizing energy
or achievement generally achieve longer survival than a pure safety
preference. This result reflects the particular structure of the environment:
energy acquisition can directly replenish the agent's resources, while
achievement-oriented behavior can contribute to longer-term survival when
sufficient energy remains. In contrast, a strong safety preference may lead
the agent to avoid risky regions and terminate relatively early through
help-seeking, limiting survival duration.

\textbf{Fruit collection.}
Preferences with achievement weight greater than approximately $0.4$ tend
to produce more fruit collection, showing that the achievement component of
the preference vector has a direct and interpretable behavioral effect.

\textbf{Battery collection.}
When the energy weight is at least $0.2$, the corresponding policies tend to
collect more batteries. Interestingly, battery collection can also emerge
under a relatively high achievement preference because energy acquisition
is instrumentally useful for sustaining subsequent fruit collection.

\textbf{Danger-zone behavior.}
Policies with high energy preference tend to enter the danger zone more
frequently, reflecting the possibility that batteries located in risky
regions remain attractive when energy is scarce. In contrast, safety-heavy
preferences generally avoid the danger zone. This demonstrates that the
three preference dimensions correspond to meaningful and competing
behavioral priorities.

\textbf{Help-seeking.}
Across fixed-preference policies, help-seeking occurs frequently and can
terminate an episode before the agent exhausts its energy. This behavior
provides an additional example of a low-level strategy represented in the
inner policy repertoire that can subsequently be selectively activated by
the outer preference generator.

\begin{table}[htb]
\centering
\caption{
Comparison of multi-objective performance between the proposed emotional
preference model, handcrafted preference-regulation baselines, the direct
survival policy, and fixed-preference strategies in the advanced environment.
}
\label{tab:results_multi}
\resizebox{\columnwidth}{!}{
\begin{tabular}{lccccc}
\toprule
Model &
Avg Steps &
Avg Batteries &
Avg Fruit &
Avg Help &
Avg Danger Steps \\
\midrule
Emotional Preference Model
& $14.81 \pm 9.56$
& $1.14 \pm 1.12$
& $0.76 \pm 1.24$
& $0.59 \pm 0.49$
& $0.61 \pm 1.01$ \\

Handcrafted-Energy Preference
& $13.62 \pm 8.86$
& $1.04 \pm 1.13$
& $1.78 \pm 1.67$
& $0.68 \pm 0.47$
& $0.51 \pm 0.92$ \\

Handcrafted-Battery Preference
& $14.39 \pm 9.44$
& $1.19 \pm 1.12$
& $1.12 \pm 1.54$
& $0.64 \pm 0.48$
& $0.75 \pm 1.10$ \\

Survival Model
& $16.26 \pm 9.04$
& $1.18 \pm 1.13$
& $0.03 \pm 0.25$
& $0.64 \pm 0.48$
& $0.44 \pm 0.85$ \\

Fixed Preference $(0.0,0.0,1.0)$
& $9.40 \pm 6.53$
& $0.36 \pm 0.76$
& $0.23 \pm 0.66$
& $0.88 \pm 0.32$
& $0.24 \pm 0.67$ \\

Fixed Preference $(0.0,0.2,0.8)$
& $12.29 \pm 9.48$
& $1.08 \pm 1.15$
& $0.40 \pm 0.91$
& $0.90 \pm 0.29$
& $0.25 \pm 0.62$ \\

Fixed Preference $(0.0,0.4,0.6)$
& $11.11 \pm 8.70$
& $1.11 \pm 1.14$
& $0.37 \pm 0.90$
& $0.85 \pm 0.35$
& $0.39 \pm 0.78$ \\

Fixed Preference $(0.0,0.6,0.4)$
& $11.85 \pm 8.48$
& $1.18 \pm 1.13$
& $0.40 \pm 0.93$
& $0.79 \pm 0.41$
& $0.52 \pm 0.94$ \\

Fixed Preference $(0.0,0.8,0.2)$
& $12.77 \pm 8.64$
& $1.17 \pm 1.12$
& $0.43 \pm 0.96$
& $0.68 \pm 0.47$
& $0.66 \pm 1.05$ \\

Fixed Preference $(0.0,1.0,0.0)$
& $14.24 \pm 9.15$
& $1.20 \pm 1.10$
& $0.45 \pm 0.98$
& $0.53 \pm 0.50$
& $0.77 \pm 1.14$ \\

Fixed Preference $(0.2,0.0,0.8)$
& $11.70 \pm 8.10$
& $0.67 \pm 0.98$
& $1.45 \pm 1.61$
& $0.91 \pm 0.28$
& $0.20 \pm 0.57$ \\

Fixed Preference $(0.2,0.2,0.6)$
& $12.83 \pm 8.77$
& $1.08 \pm 1.15$
& $1.44 \pm 1.62$
& $0.88 \pm 0.32$
& $0.29 \pm 0.68$ \\

Fixed Preference $(0.2,0.4,0.4)$
& $12.07 \pm 8.52$
& $1.14 \pm 1.14$
& $1.17 \pm 1.50$
& $0.82 \pm 0.38$
& $0.44 \pm 0.86$ \\

Fixed Preference $(0.2,0.6,0.2)$
& $12.12 \pm 8.40$
& $1.16 \pm 1.13$
& $1.00 \pm 1.41$
& $0.75 \pm 0.43$
& $0.58 \pm 0.98$ \\

Fixed Preference $(0.2,0.8,0.0)$
& $12.58 \pm 8.48$
& $1.17 \pm 1.12$
& $0.91 \pm 1.36$
& $0.65 \pm 0.48$
& $0.72 \pm 1.09$ \\

Fixed Preference $(0.4,0.0,0.6)$
& $12.34 \pm 8.39$
& $0.78 \pm 1.04$
& $1.74 \pm 1.69$
& $0.89 \pm 0.31$
& $0.25 \pm 0.65$ \\

Fixed Preference $(0.4,0.2,0.4)$
& $13.14 \pm 8.74$
& $1.05 \pm 1.14$
& $1.73 \pm 1.71$
& $0.84 \pm 0.37$
& $0.37 \pm 0.79$ \\

Fixed Preference $(0.4,0.4,0.2)$
& $12.96 \pm 8.57$
& $1.13 \pm 1.14$
& $1.60 \pm 1.66$
& $0.76 \pm 0.43$
& $0.54 \pm 0.96$ \\

Fixed Preference $(0.4,0.6,0.0)$
& $12.85 \pm 8.50$
& $1.16 \pm 1.12$
& $1.44 \pm 1.62$
& $0.68 \pm 0.47$
& $0.67 \pm 1.06$ \\

Fixed Preference $(0.6,0.0,0.4)$
& $12.94 \pm 8.56$
& $0.88 \pm 1.09$
& $1.84 \pm 1.72$
& $0.88 \pm 0.33$
& $0.28 \pm 0.69$ \\

Fixed Preference $(0.6,0.2,0.2)$
& $13.32 \pm 8.77$
& $1.04 \pm 1.13$
& $1.81 \pm 1.71$
& $0.80 \pm 0.40$
& $0.44 \pm 0.88$ \\

Fixed Preference $(0.6,0.4,0.0)$
& $13.36 \pm 8.69$
& $1.11 \pm 1.14$
& $1.78 \pm 1.72$
& $0.71 \pm 0.45$
& $0.61 \pm 1.02$ \\

Fixed Preference $(0.8,0.0,0.2)$
& $13.08 \pm 8.74$
& $0.92 \pm 1.11$
& $1.82 \pm 1.70$
& $0.86 \pm 0.35$
& $0.32 \pm 0.74$ \\

Fixed Preference $(0.8,0.2,0.0)$
& $13.45 \pm 8.84$
& $1.03 \pm 1.13$
& $1.88 \pm 1.73$
& $0.78 \pm 0.41$
& $0.48 \pm 0.90$ \\

Fixed Preference $(1.0,0.0,0.0)$
& $13.12 \pm 8.70$
& $0.94 \pm 1.11$
& $1.80 \pm 1.67$
& $0.85 \pm 0.35$
& $0.32 \pm 0.73$ \\
\bottomrule
\end{tabular}
}
\end{table}

\subsection{Comparison with Direct Survival Optimization}
\label{sec:advanced_survival}

The direct Survival Model achieves the highest average survival duration,
with
$
16.26\pm9.04
$
steps. However, it collects almost no fruit, with an average of only
$
0.03\pm0.25,
$
and it largely avoids the danger zone. Its average battery collection is
$
1.18\pm1.13.
$

This result is consistent with the role of the direct survival controller:
it optimizes the outer objective without being constrained to express its
behavior through an interpretable preference over the three lower-level
objectives.

In contrast, the Emotional Preference Model achieves
$
14.81\pm9.56
$
survival steps while simultaneously producing
$
1.14\pm1.12
$
battery collections and
$
0.76\pm1.24
$
fruit collections. Its behavior therefore reflects a broader organization
of lower-level objectives rather than pure survival maximization.

The difference in survival performance is consistent with the theoretical
analysis in Section~5: the direct Survival Model optimizes in the full
physical policy space, whereas the Emotional Preference Model optimizes
within the policy space represented by the inner MORL controller.

\subsection{Handcrafted Preference-Regulation Baselines}
\label{sec:advanced_handcrafted}

We compare the learned preference generator with two manually designed
context-dependent preference rules.

\subsubsection{Energy-State-Based Preference Regulation}

We define the preference as
\[
w(s)=
\begin{cases}
(0.05,0.90,0.05)^\top,
& E(s)\leq3,\\
(0.85,0.10,0.05)^\top,
& E(s)>3.
\end{cases}
\]

This baseline increases the relative priority of energy when the agent is
energy-constrained and otherwise favors achievement.

Its survival duration is
$
13.62\pm8.86,
$
which is lower than the proposed Emotional Preference Model. The model
collects $1.78\pm1.67$ fruits and $1.04\pm1.13$ batteries. Its relatively
high fruit collection demonstrates that the handcrafted rule can produce
useful contextual behavior, but the single energy threshold does not fully
capture the interaction among energy, resource availability, and danger.

\subsubsection{Battery-Availability-Based Preference Regulation}

The second handcrafted baseline uses the availability of batteries:
\[
w(s)=
\begin{cases}
(0.05,0.90,0.05)^\top,
& N_{\mathrm{battery}}(s)\geq1,\\
(0.85,0.10,0.05)^\top,
& N_{\mathrm{battery}}(s)=0.
\end{cases}
\]

This baseline achieves
$
14.39\pm9.44
$
survival steps, together with
$
1.12\pm1.54
$
fruit collections and
$
1.19\pm1.12
$
battery collections.

Its performance is relatively close to the proposed method, demonstrating
that explicit domain knowledge can provide a strong contextual preference
baseline. Nevertheless, the learned preference generator does not require
such a manually specified switching rule and can use multiple state
variables jointly.

\subsection{Emergent Emotional Preference Regulation}
\label{sec:advanced_emotional_preference}

\subsubsection{Distribution of Learned Preference States}

We evaluate the trained preference generator over 100 episodes, resulting
in 1,478 observed states. For visualization, each continuous preference
vector is assigned to its nearest reference preference according to
three-dimensional Chebyshev distance.

The largest groups correspond to the energy-dominant preference
$
(0,1,0)^\top,
$
the safety-dominant preference
$
(0,0,1)^\top,
$
and the achievement-dominant preference
$
(1,0,0)^\top.
$

Their observed counts are 568, 367, and 186, respectively, corresponding to
approximately 38.4\%, 24.8\%, and 12.6\% of all observed states. The
remaining states occupy intermediate regions of the preference simplex.

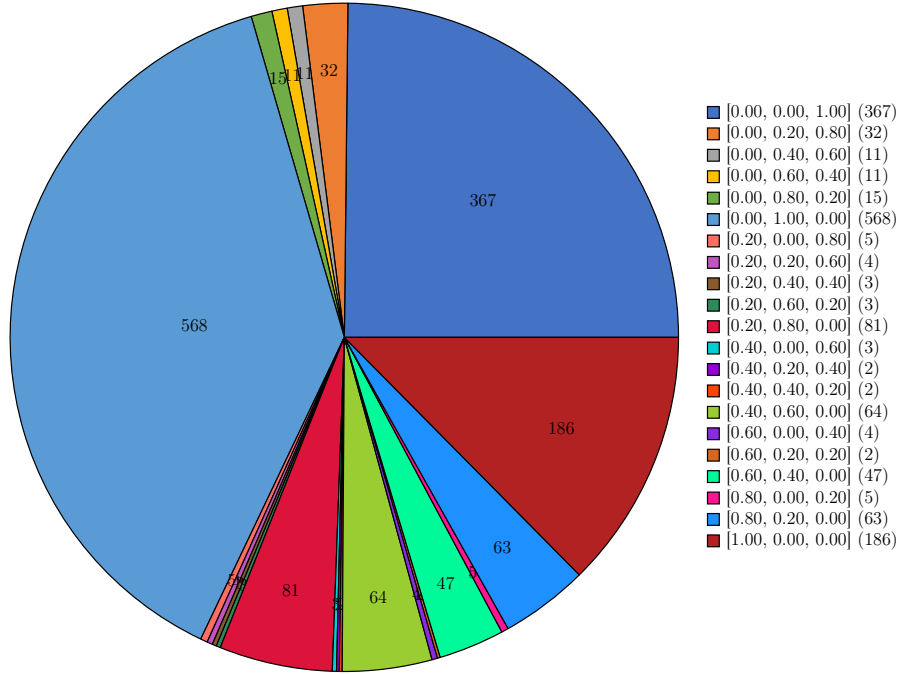
\begin{figure}[htb]
  \centering
  \resizebox{0.9\textwidth}{!}{%
    \begin{tikzpicture}
      \pie[
        sum=auto,
        radius=7.8,
        color={
          c1,c2,c3,c4,c5,c6,c7,c8,c9,c10,c11,c12,c13,c14,c15,c16,c17,
          c18,c19,c20,c21
        },
        text=legend
      ]
      {
        367/{[0.00, 0.00, 1.00] (367)},
        32/{[0.00, 0.20, 0.80] (32)},
        11/{[0.00, 0.40, 0.60] (11)},
        11/{[0.00, 0.60, 0.40] (11)},
        15/{[0.00, 0.80, 0.20] (15)},
        568/{[0.00, 1.00, 0.00] (568)},
        5/{[0.20, 0.00, 0.80] (5)},
        4/{[0.20, 0.20, 0.60] (4)},
        3/{[0.20, 0.40, 0.40] (3)},
        3/{[0.20, 0.60, 0.20] (3)},
        81/{[0.20, 0.80, 0.00] (81)},
        3/{[0.40, 0.00, 0.60] (3)},
        2/{[0.40, 0.20, 0.40] (2)},
        2/{[0.40, 0.40, 0.20] (2)},
        64/{[0.40, 0.60, 0.00] (64)},
        4/{[0.60, 0.00, 0.40] (4)},
        2/{[0.60, 0.20, 0.20] (2)},
        47/{[0.60, 0.40, 0.00] (47)},
        5/{[0.80, 0.00, 0.20] (5)},
        63/{[0.80, 0.20, 0.00] (63)},
        186/{[1.00, 0.00, 0.00] (186)}
      }
    \end{tikzpicture}%
  }
  \caption{
  Distribution of visited states across learned emotional-preference
  regions in the advanced environment. Continuous preference outputs are
  assigned to the nearest reference preference for visualization.
  }
  \label{fig:pie2}
\end{figure}

The distribution indicates that the preference generator does not simply
alternate between a small number of fixed preferences. Instead, it produces
both extreme and intermediate preference states. The resulting distribution
therefore provides evidence for graded and contextualized preference
regulation among the three objectives.

\subsubsection{Contextual Preference Regularities}

We next examine the environmental situations associated with the dominant
preference states.

Among the 186 states assigned to the achievement-dominant preference
$(1,0,0)^\top$, most occur when achievement-oriented behavior remains useful
and sufficient resources are available. Among the 367 states assigned to
the safety-dominant preference $(0,0,1)^\top$, 264 states correspond to
situations in which no battery remains available.

The energy-dominant preference $(0,1,0)^\top$ can also occur when no
battery remains. These states arise partly because the fixed energy-heavy
inner policy terminates early through help-seeking after collecting the
available batteries and therefore has limited coverage of some low-energy
states. The outer preference generator can nevertheless select this
policy in such states without inheriting the same terminal behavior,
thereby exploiting its existing state-action structure to extend survival.

These results indicate that the learned preference is not adequately
described as a simple threshold transformation of energy. Rather, it
depends jointly on resource availability, energy state, and environmental
risk. The preference vector acts as a compact representation of the
relative priority assigned to competing goal-directed strategies in the
current context.

\subsubsection{Representative Preference Dynamics}

\begin{figure}[htbp]
  \centering
  \begin{tabular}{cccc}
    \includegraphics[width=0.2\textwidth]{
    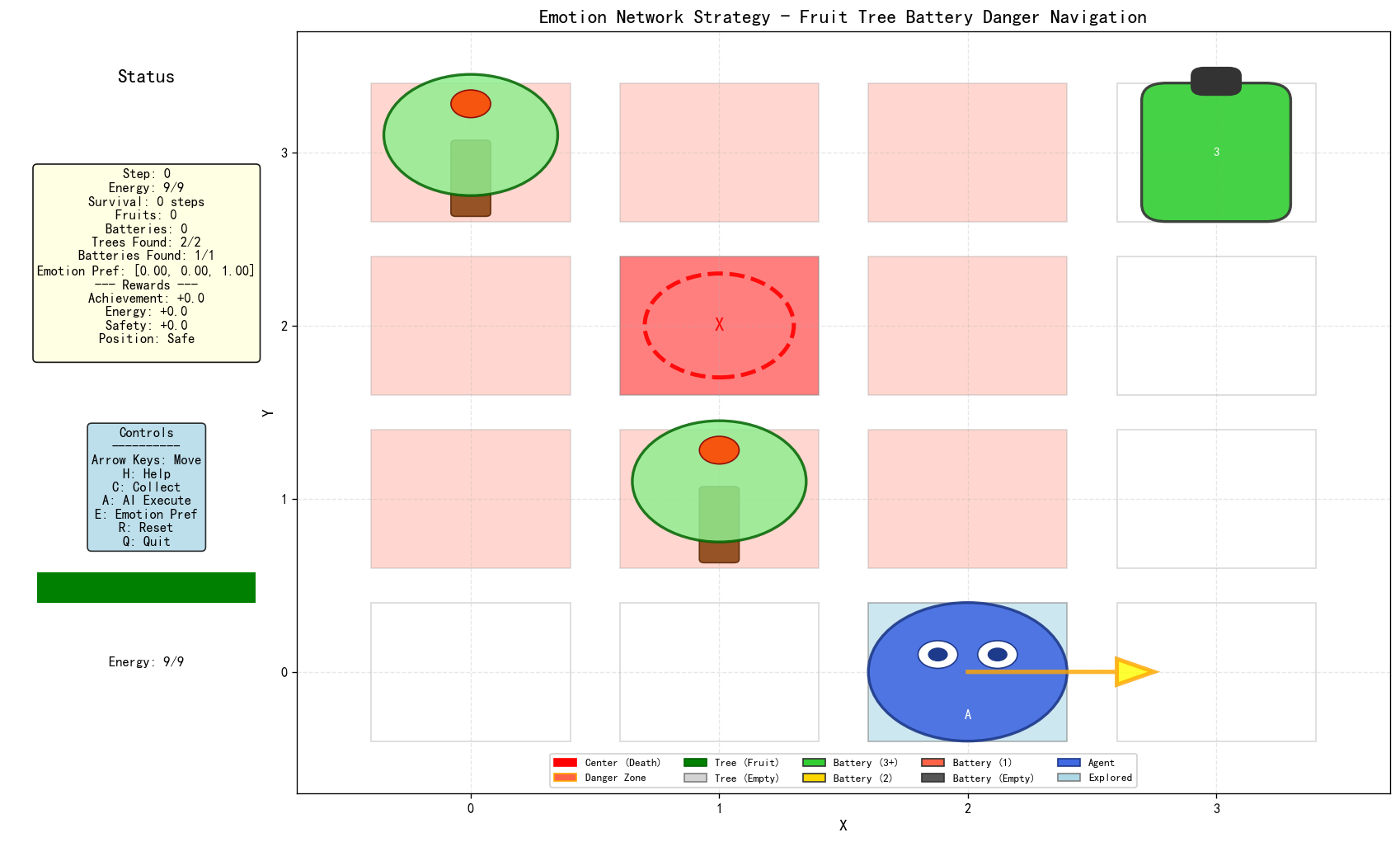} &
    \includegraphics[width=0.2\textwidth]{
    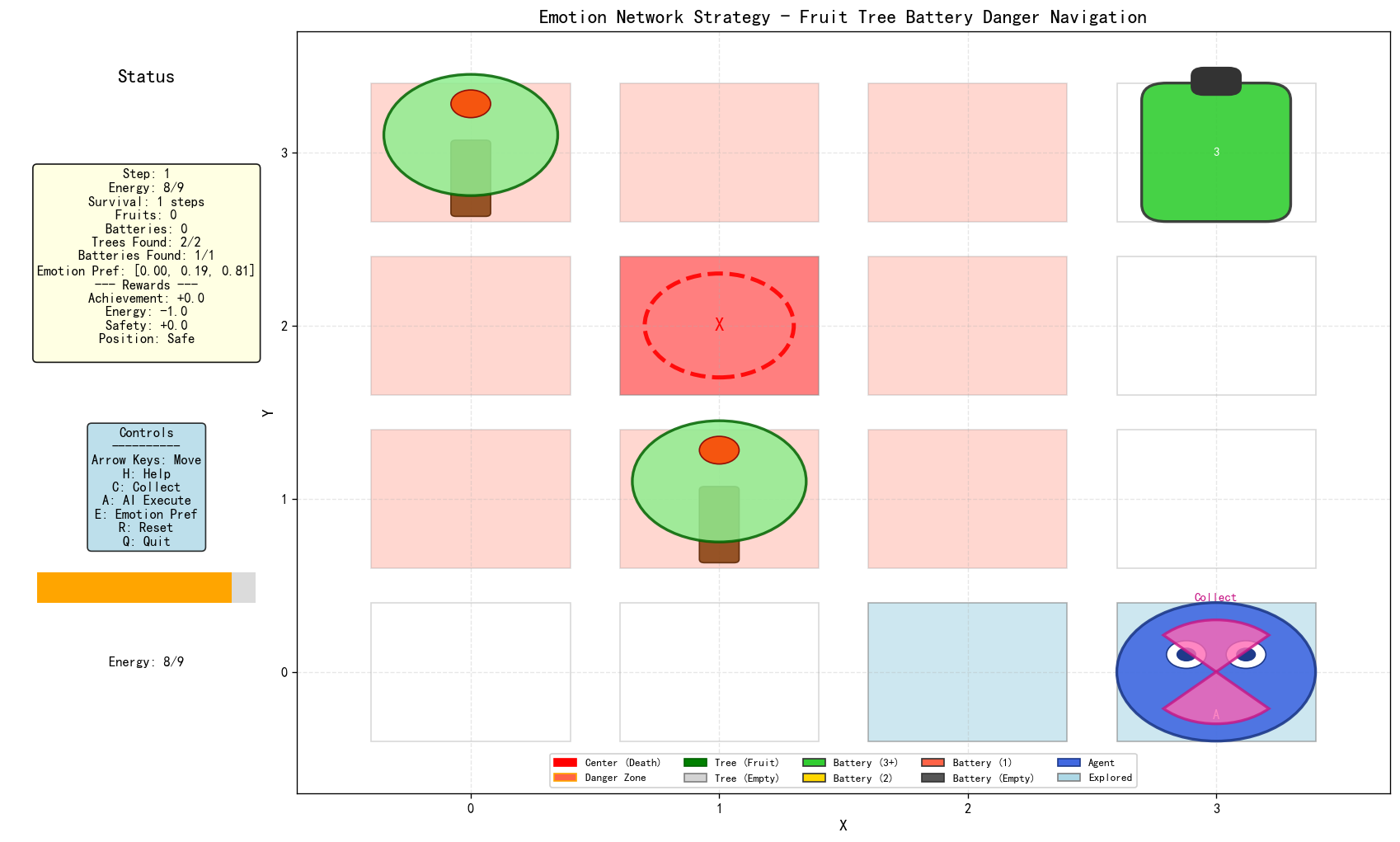} &
    \includegraphics[width=0.2\textwidth]{
    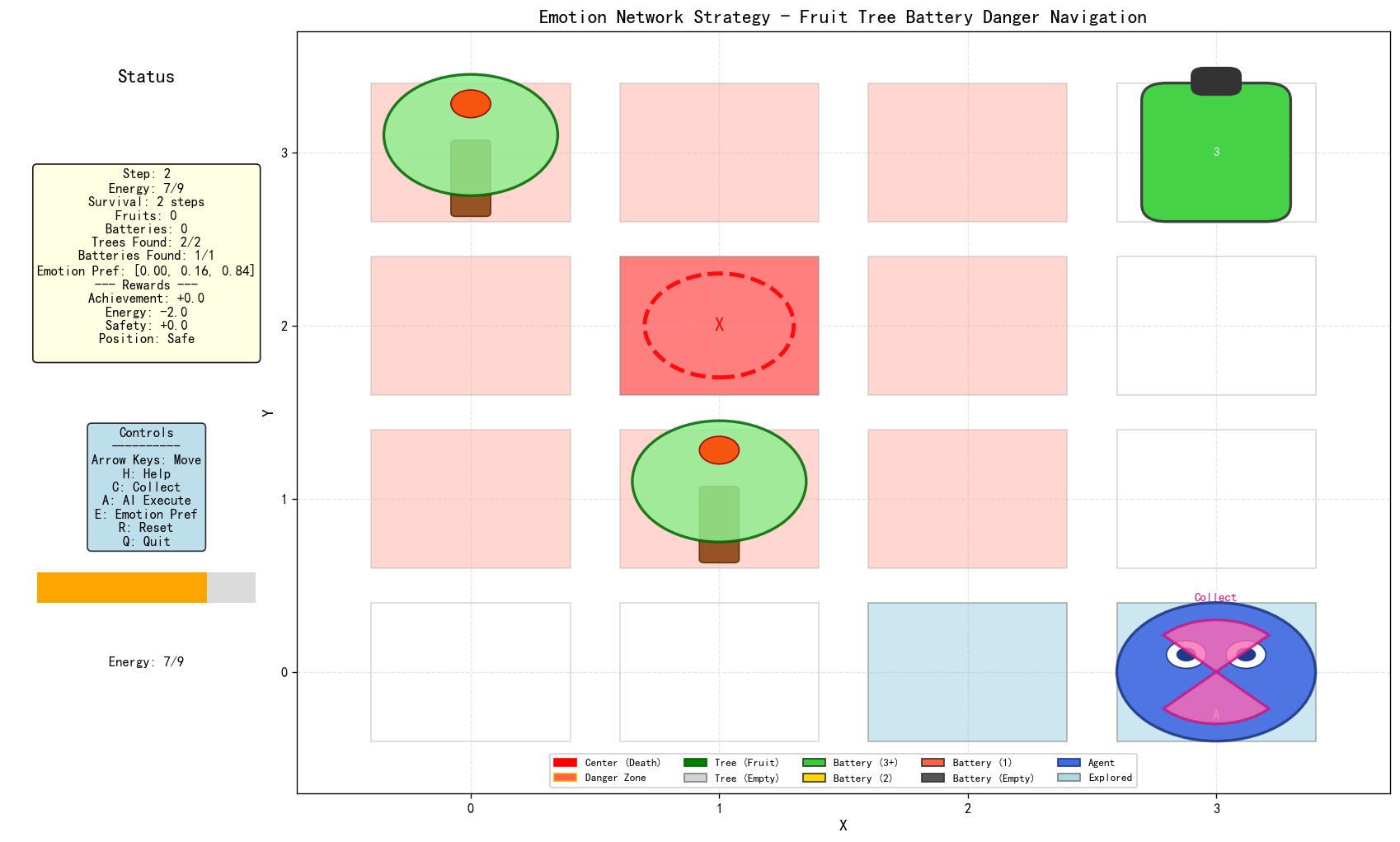} &
    \includegraphics[width=0.2\textwidth]{
    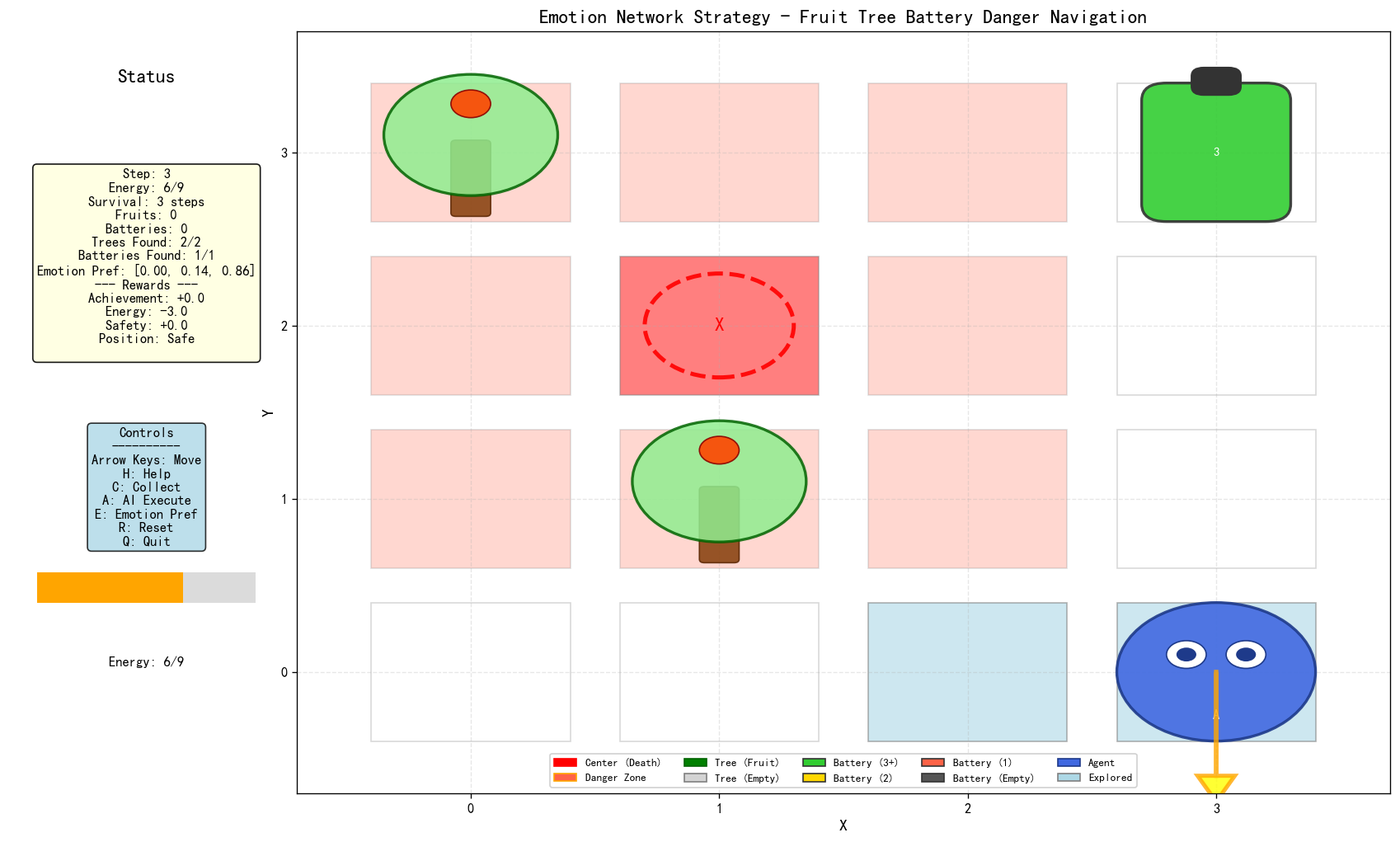} \\

    \includegraphics[width=0.2\textwidth]{
    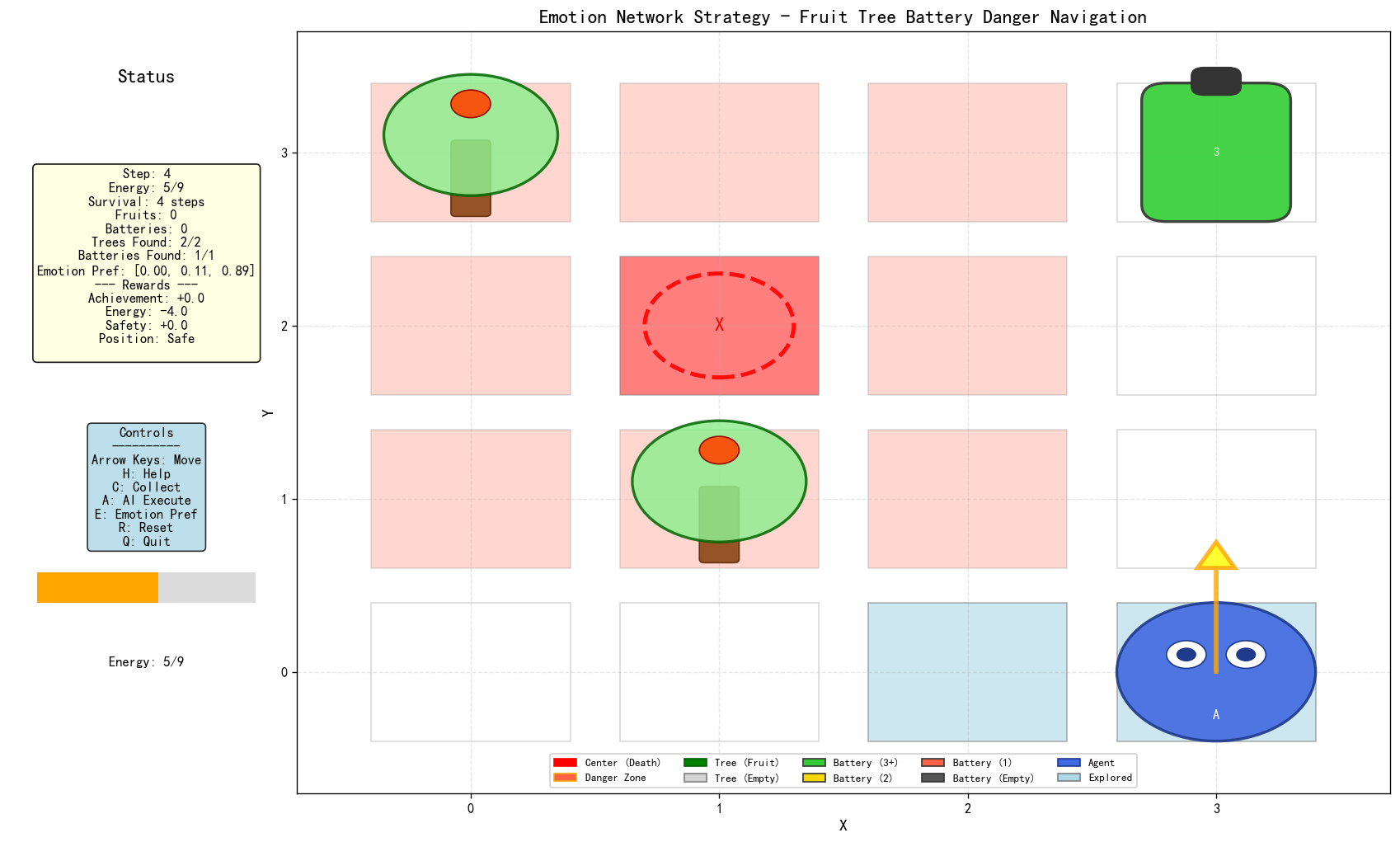} &
    \includegraphics[width=0.2\textwidth]{
    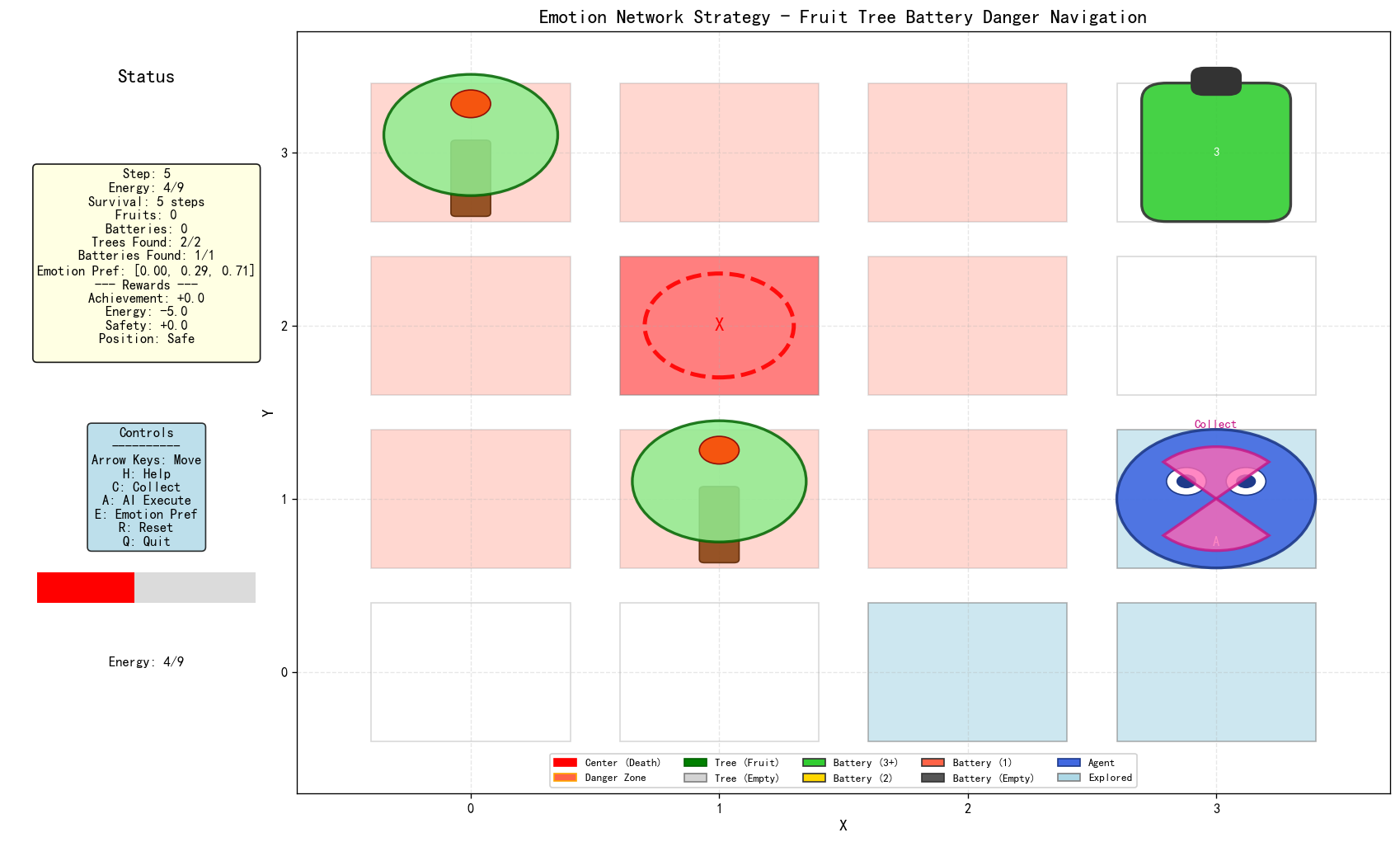} &
    \includegraphics[width=0.2\textwidth]{
    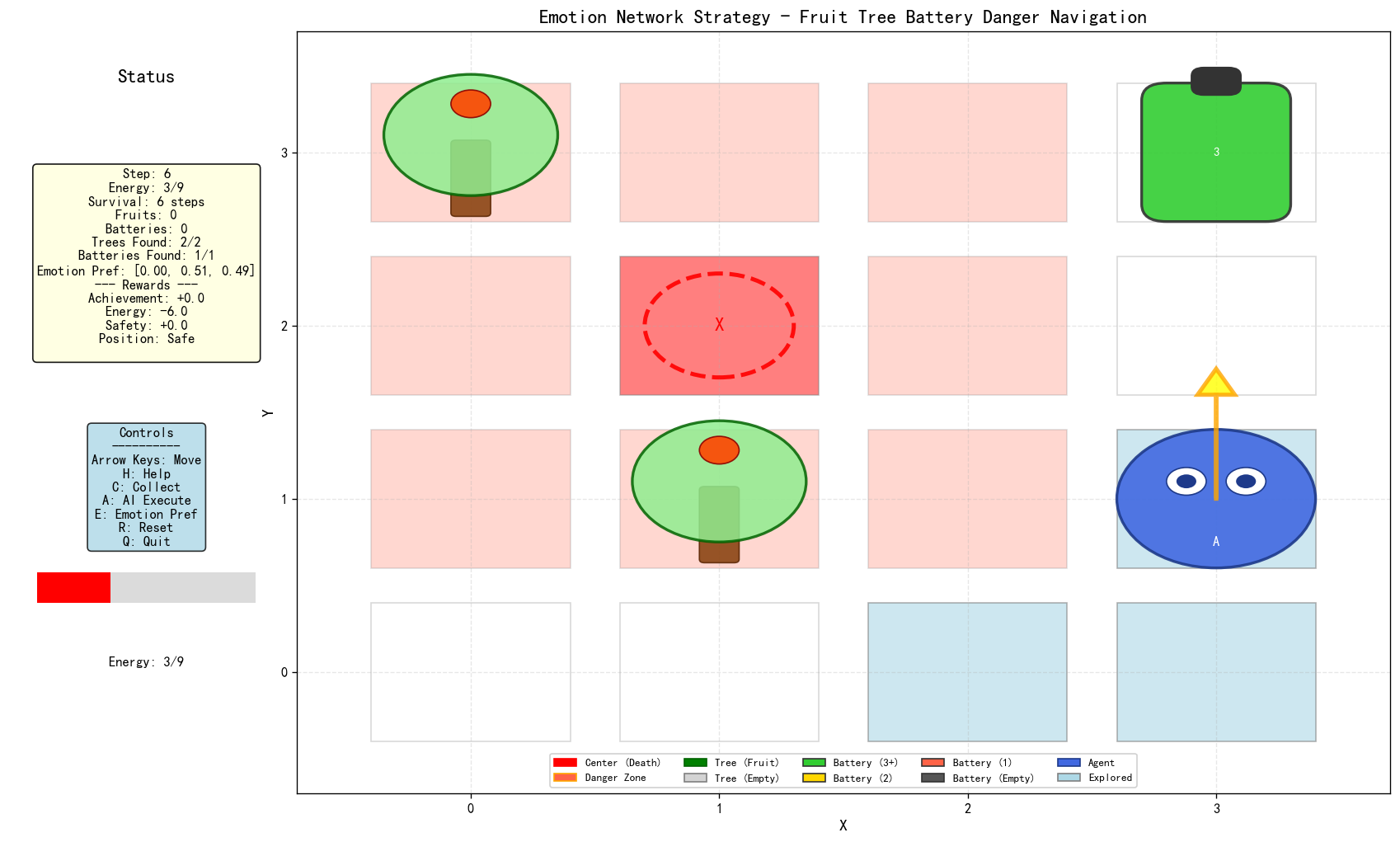} &
    \includegraphics[width=0.2\textwidth]{
    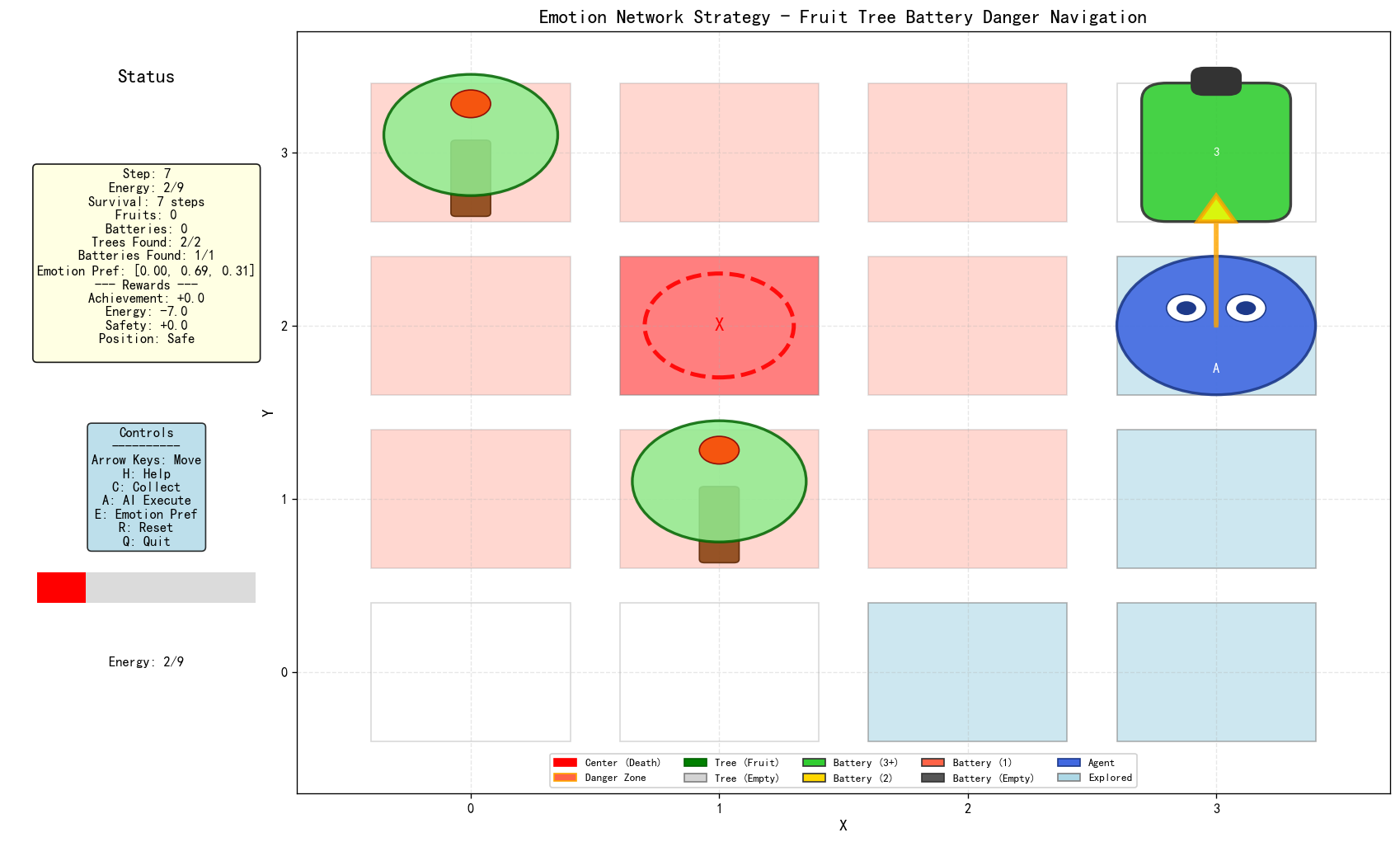} \\

    \includegraphics[width=0.2\textwidth]{
    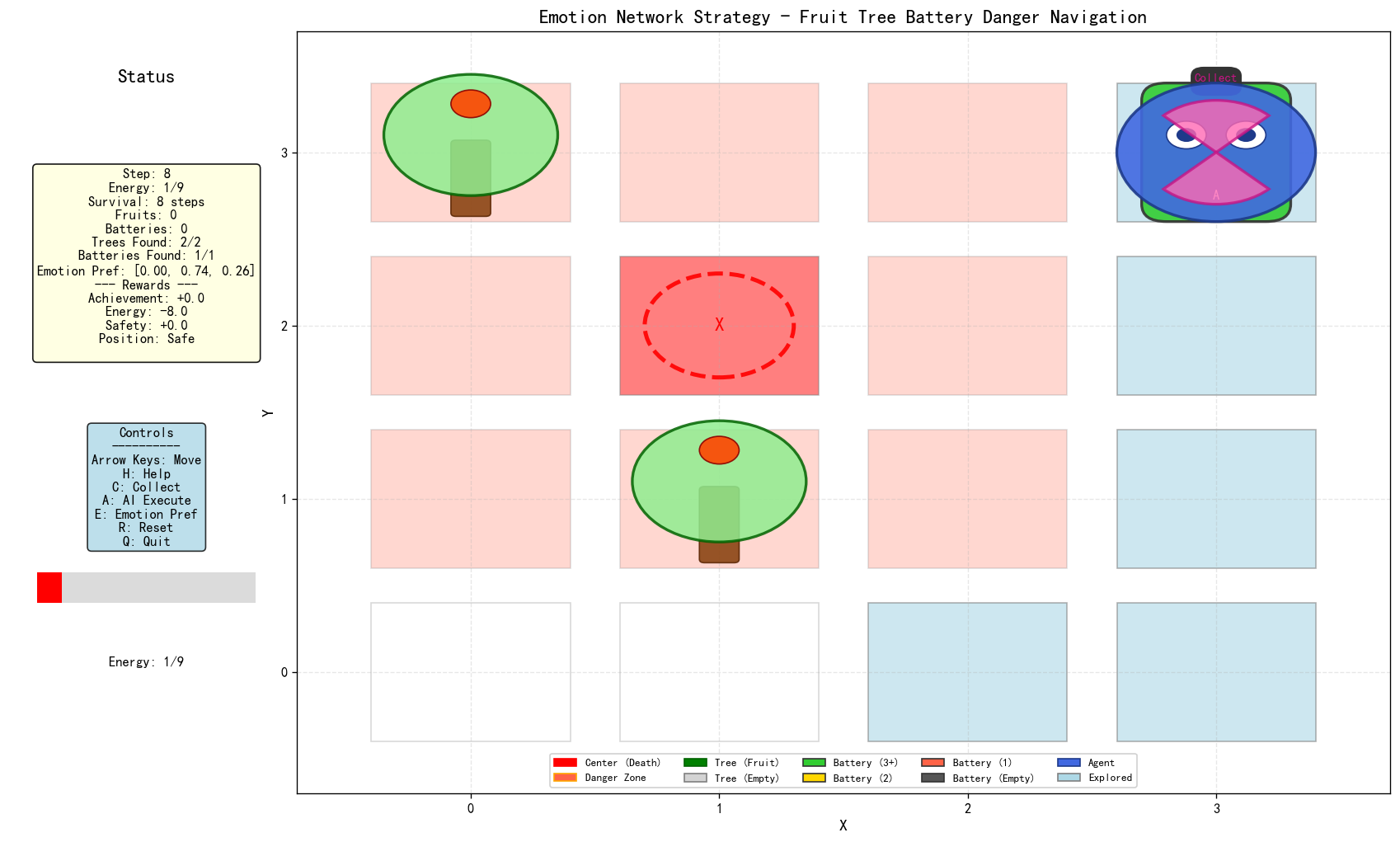} &
    \includegraphics[width=0.2\textwidth]{
    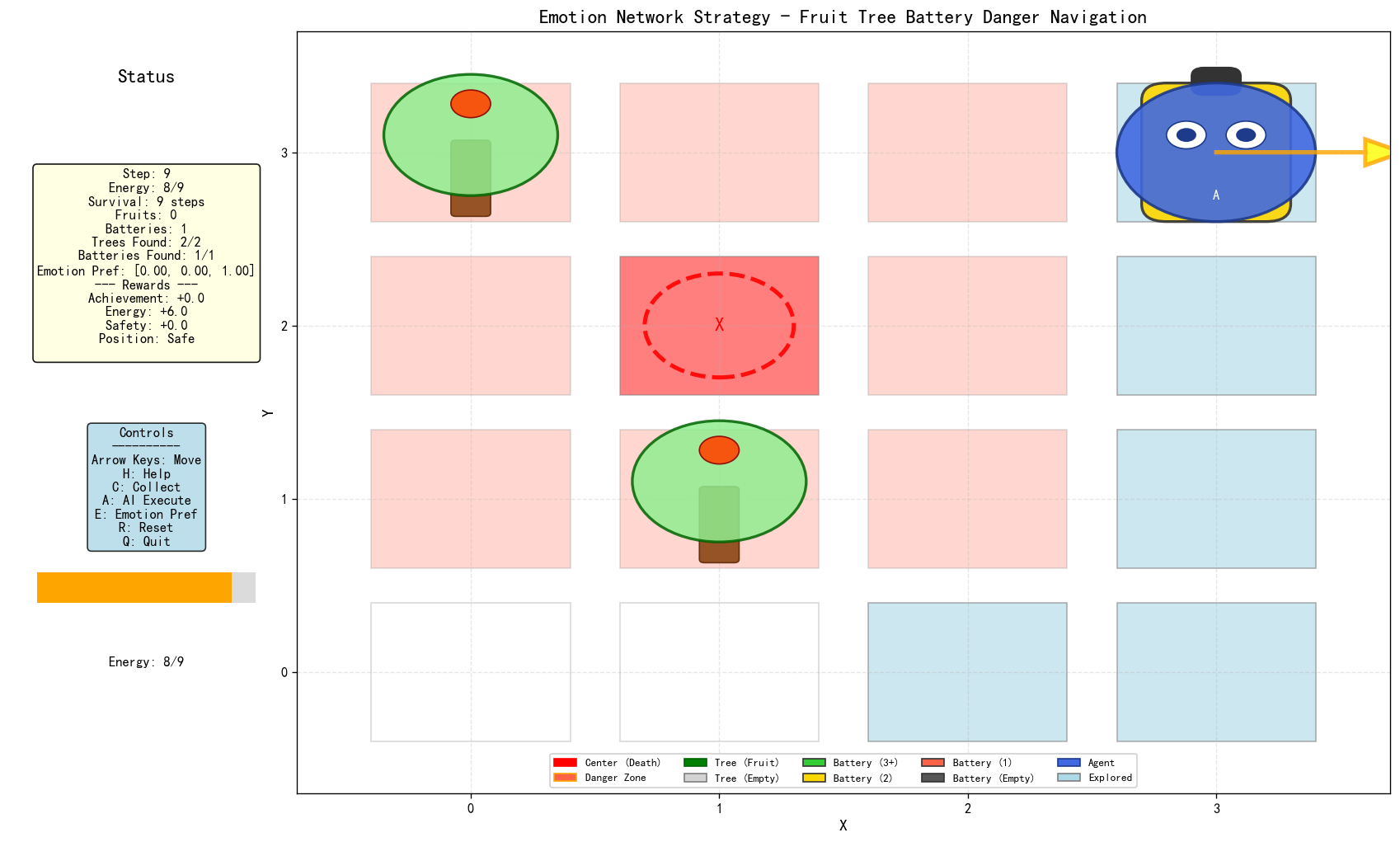} &
    \includegraphics[width=0.2\textwidth]{
    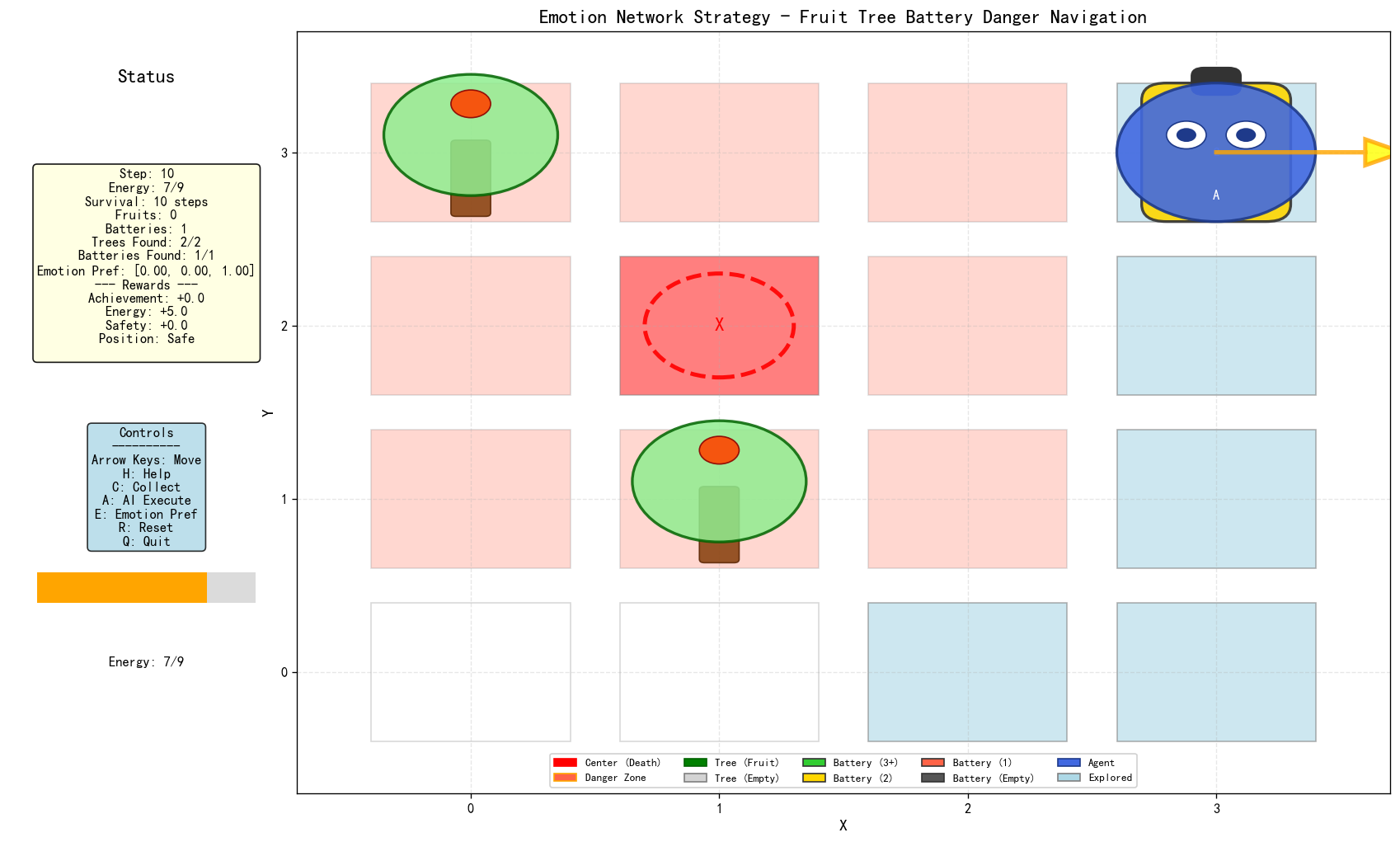} &
    \includegraphics[width=0.2\textwidth]{
    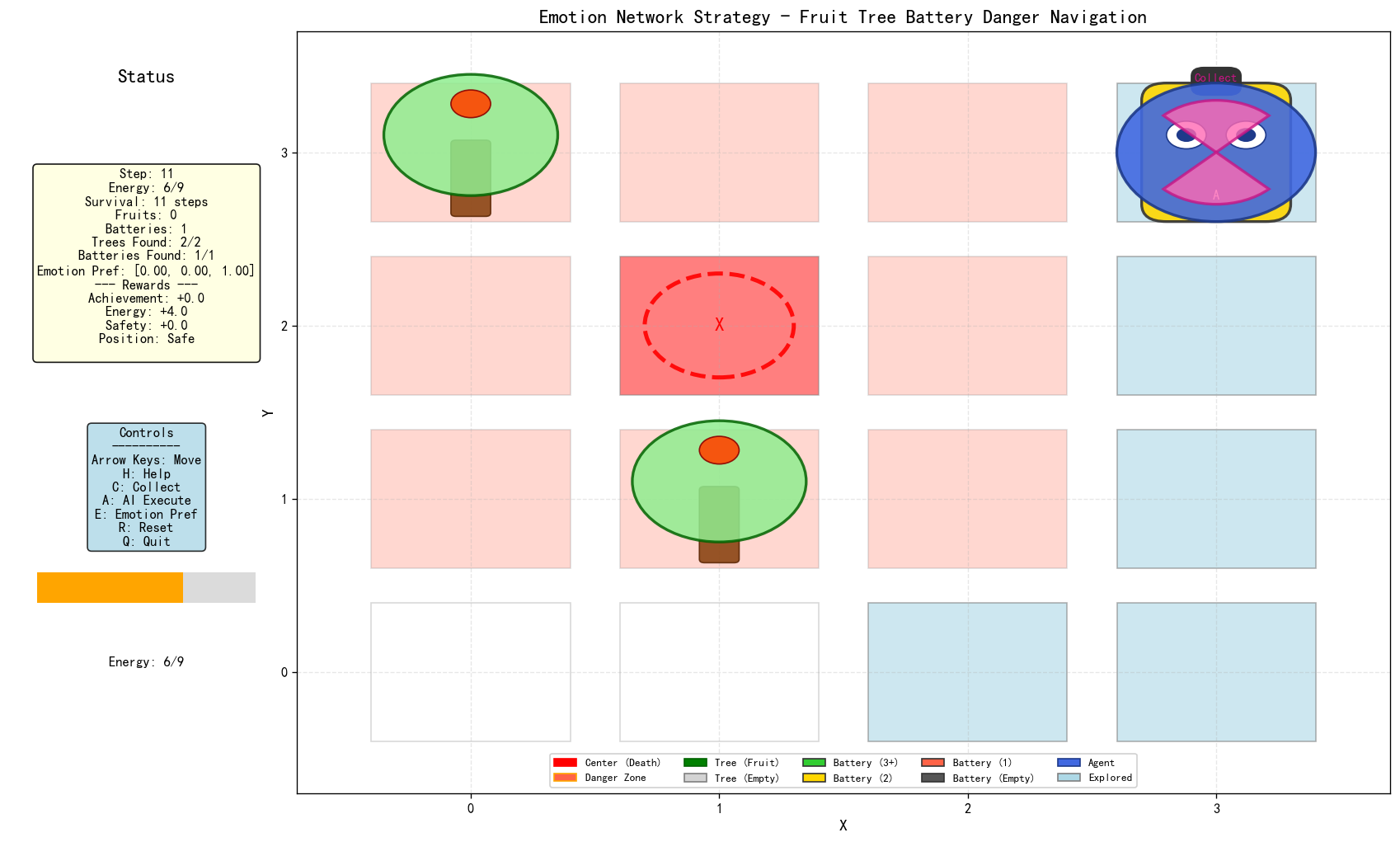} \\

    \includegraphics[width=0.2\textwidth]{
    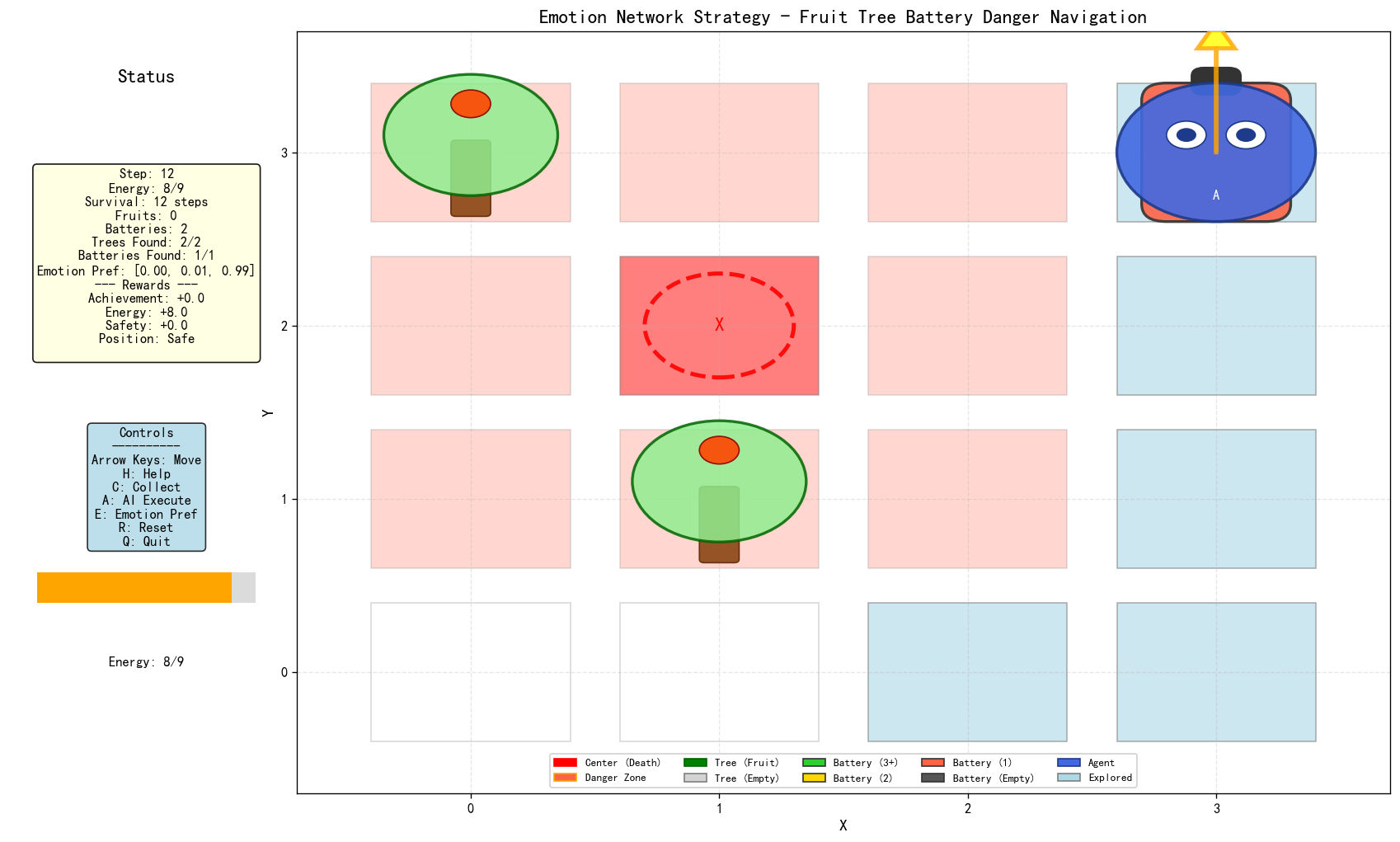} &
    \includegraphics[width=0.2\textwidth]{
    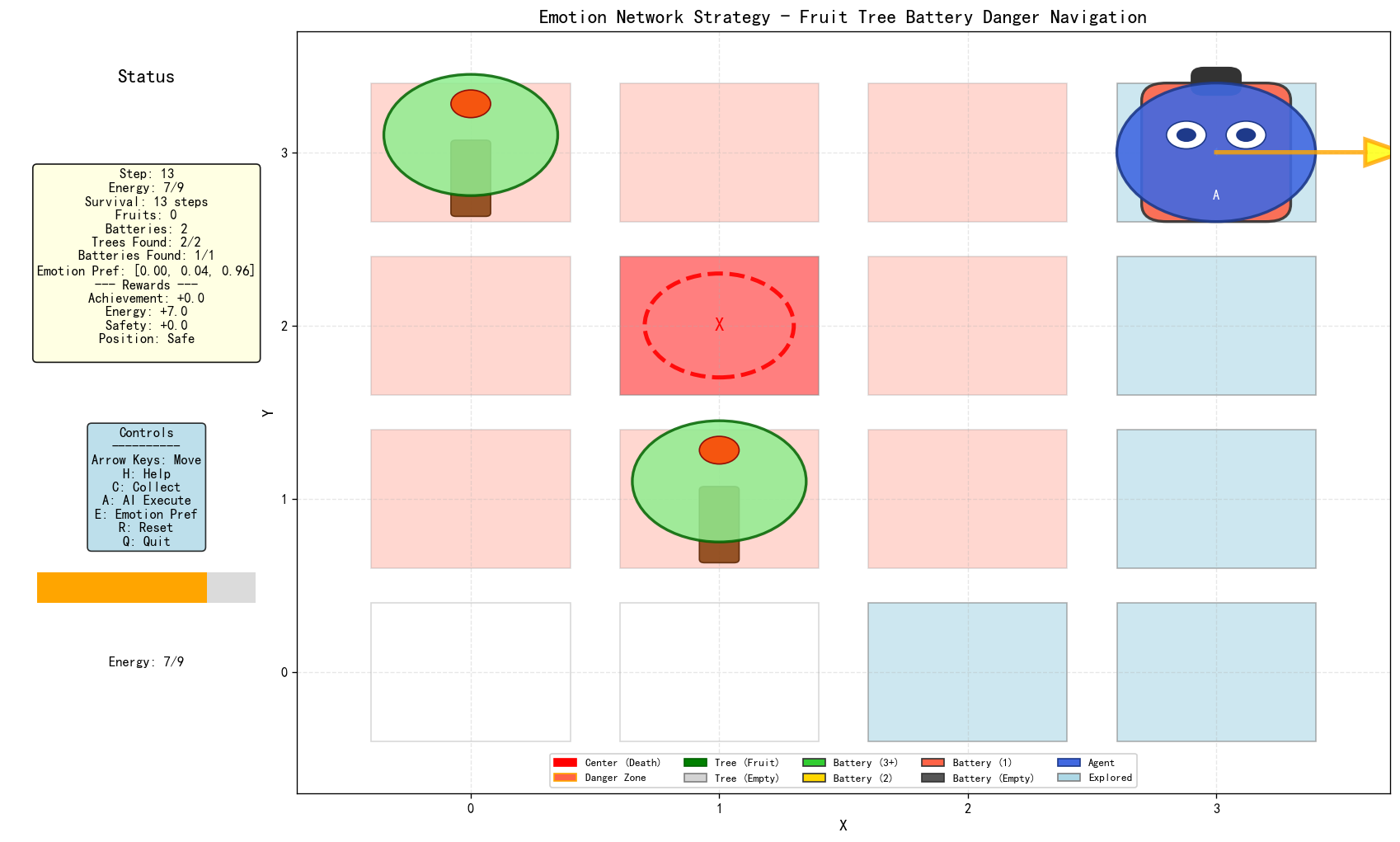} &
    \includegraphics[width=0.2\textwidth]{
    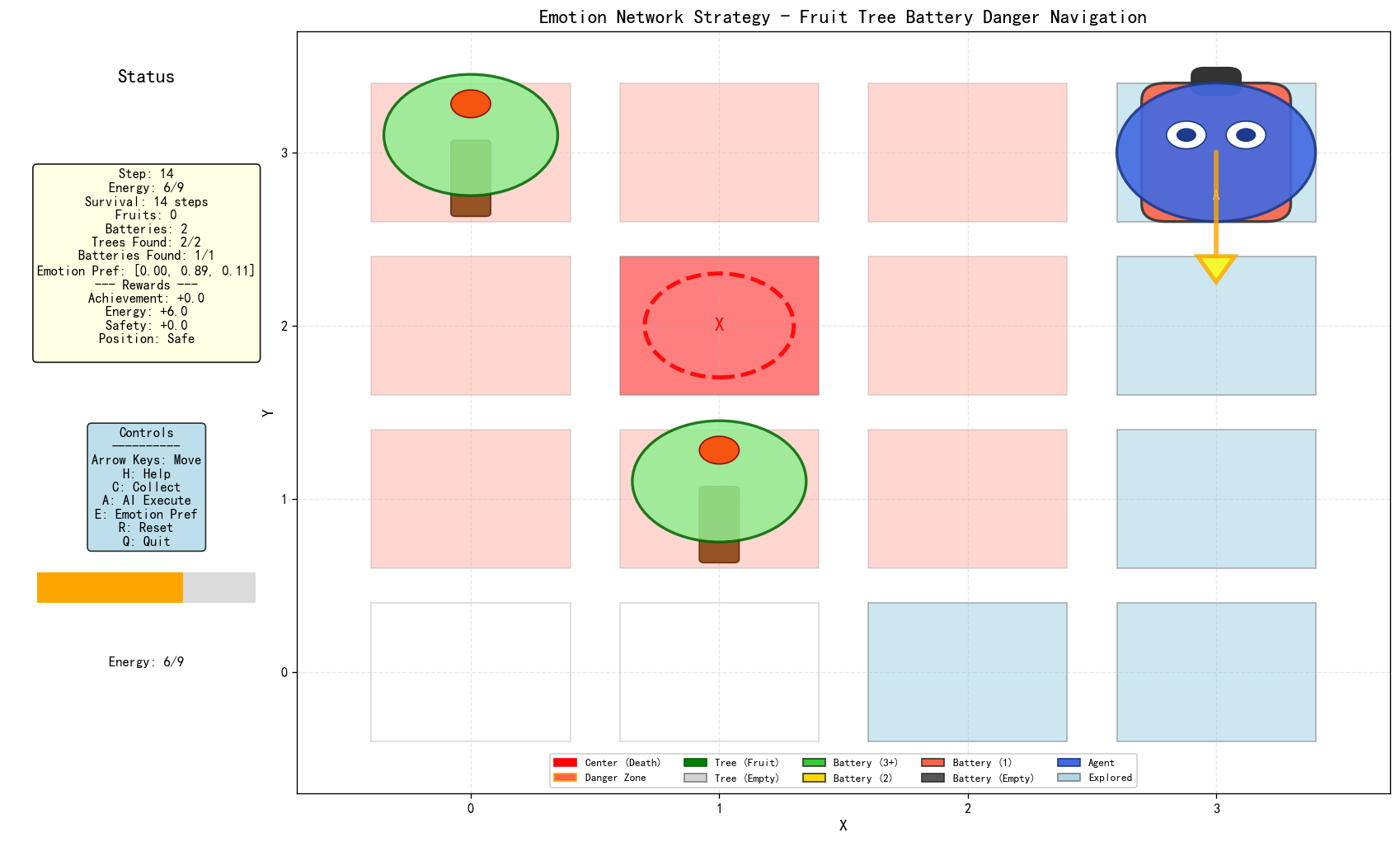} &
    \includegraphics[width=0.2\textwidth]{
    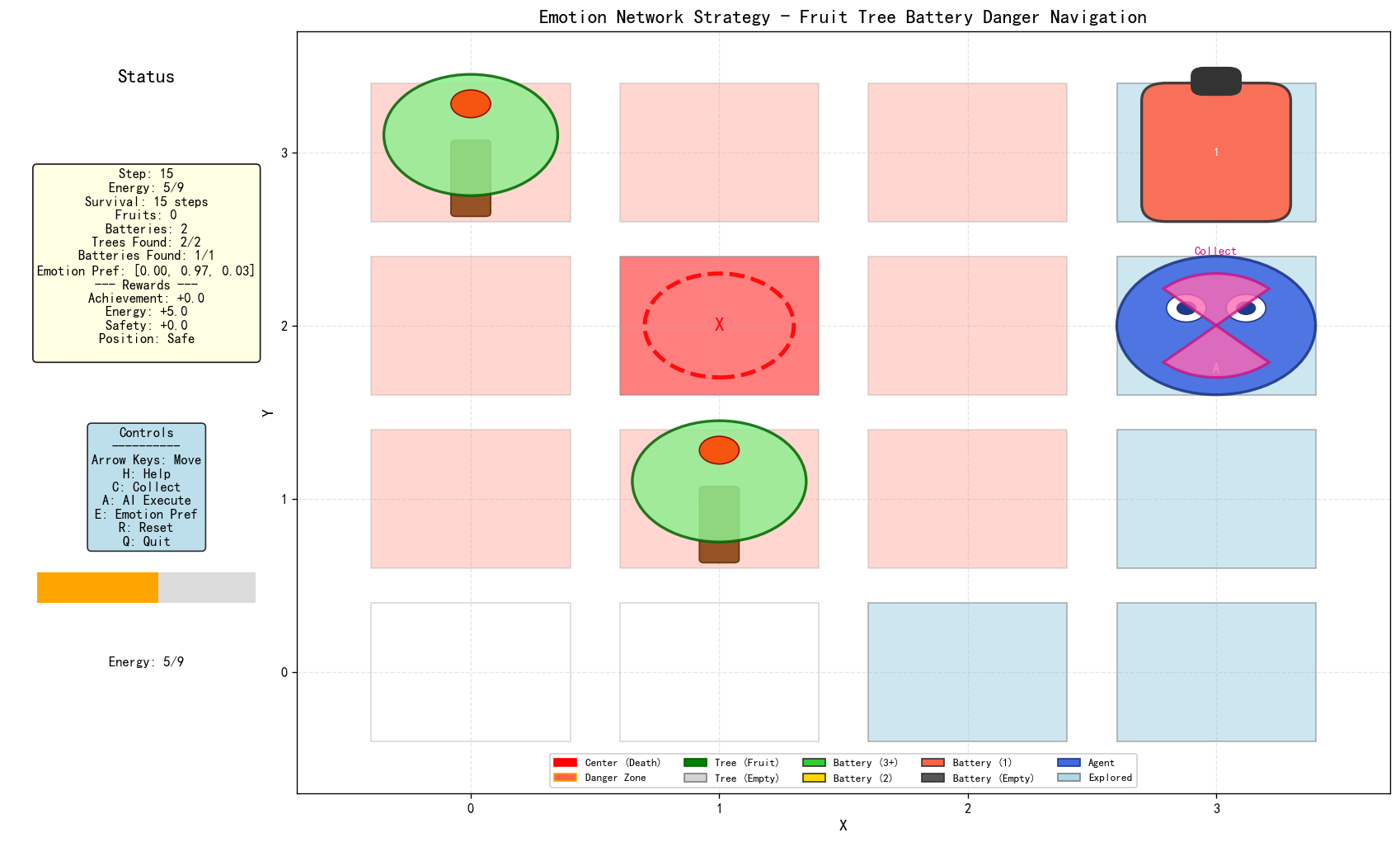} \\

    \includegraphics[width=0.2\textwidth]{
    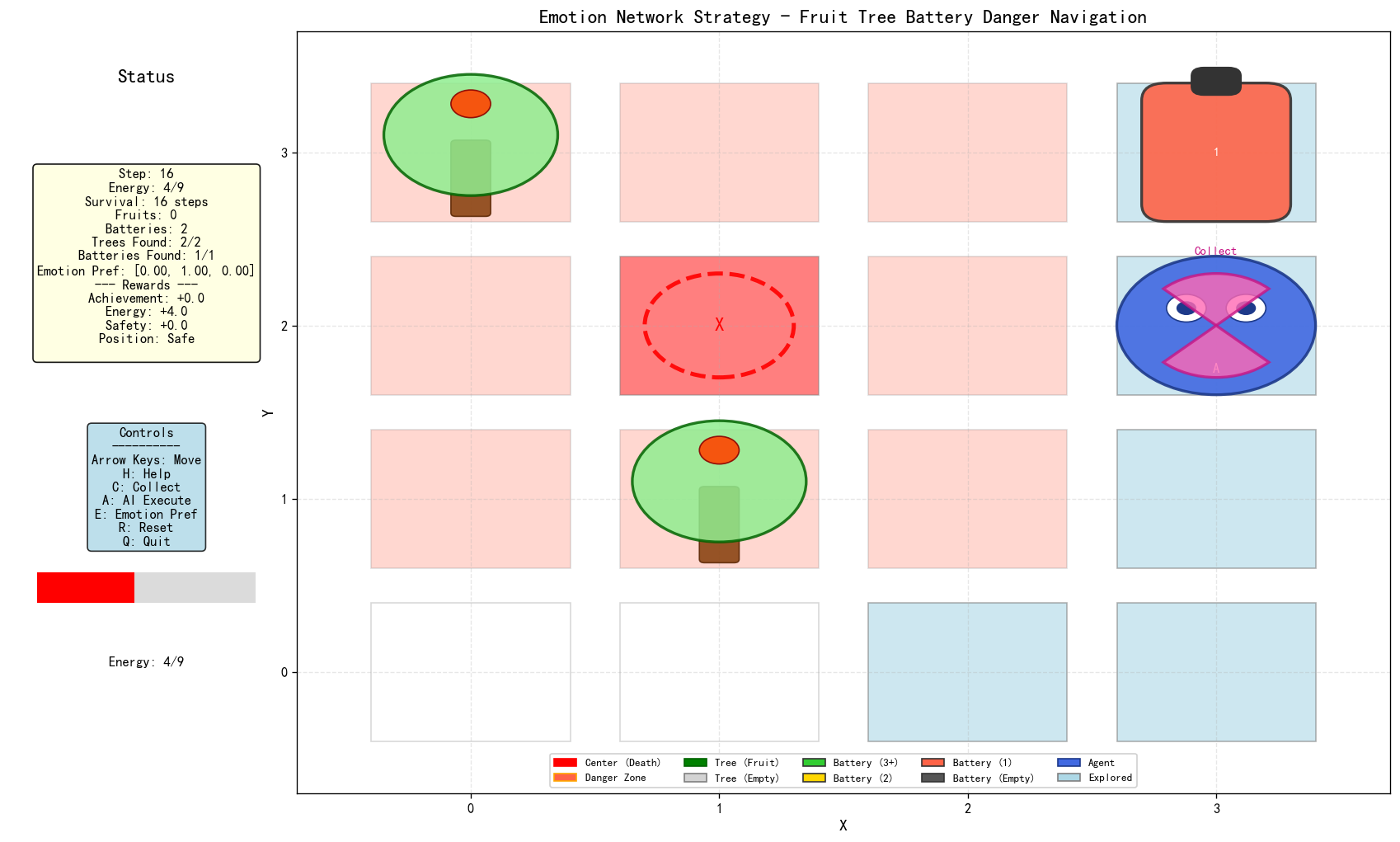} &
    \includegraphics[width=0.2\textwidth]{
    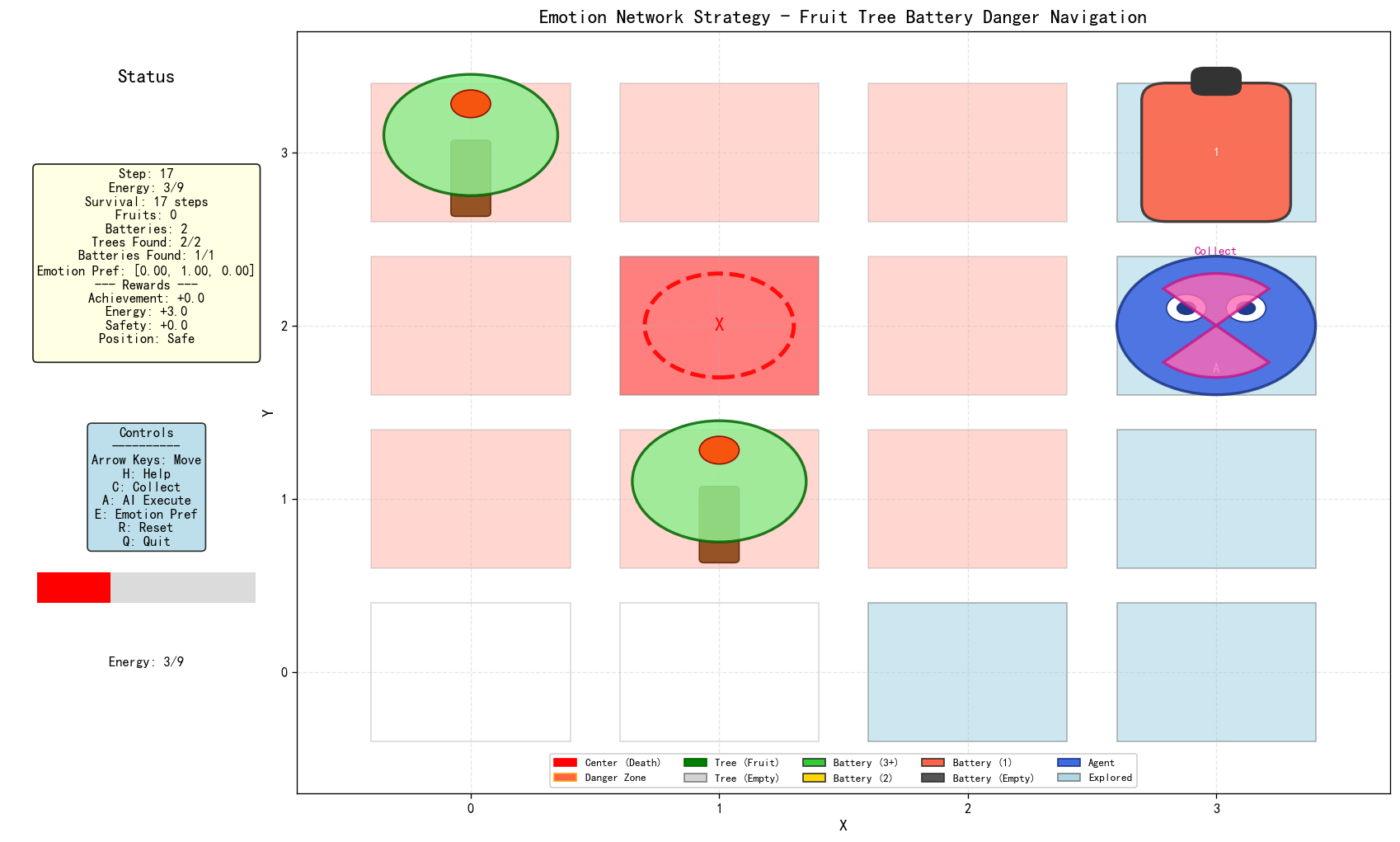} &
    \includegraphics[width=0.2\textwidth]{
    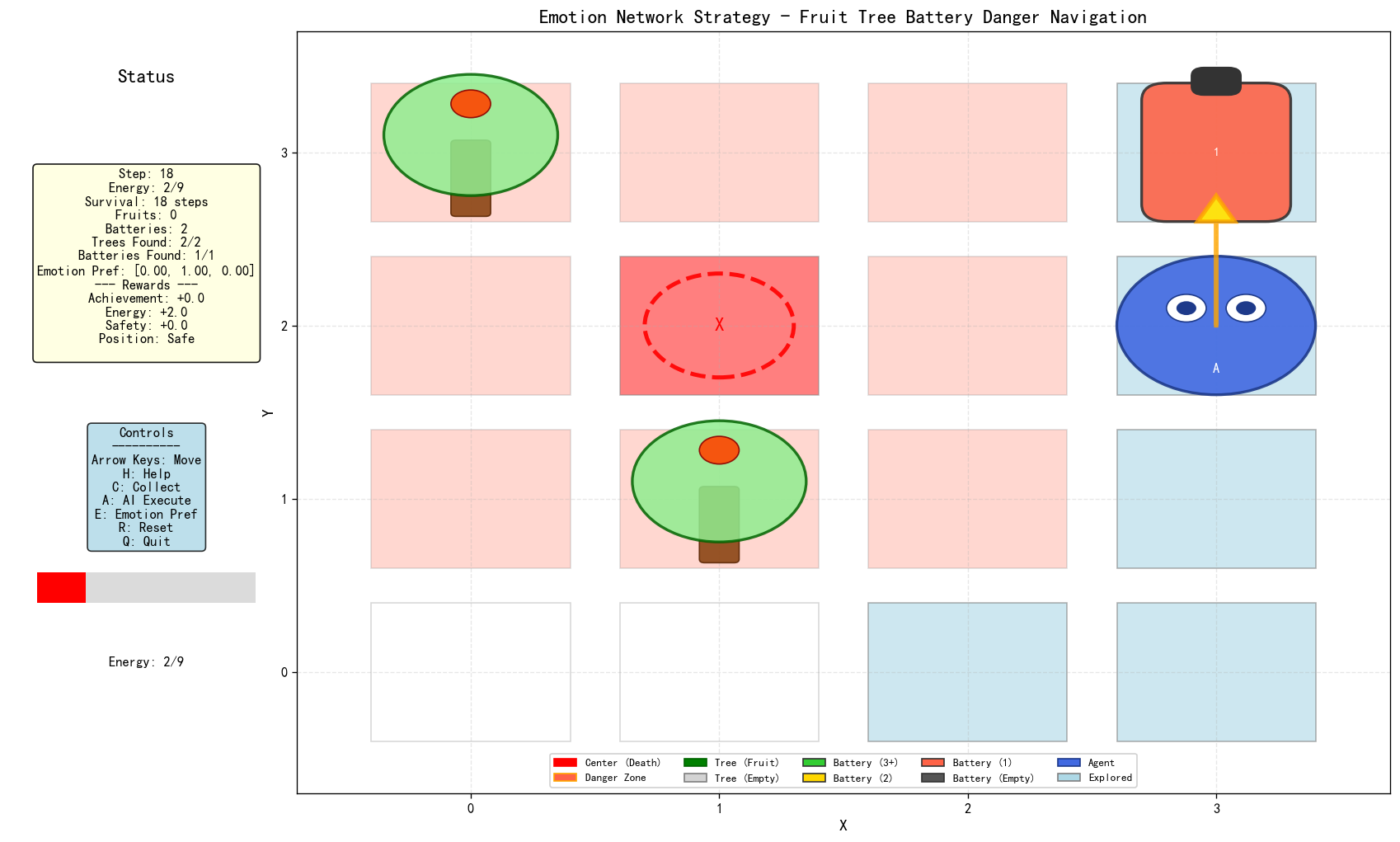} &
    \includegraphics[width=0.2\textwidth]{
    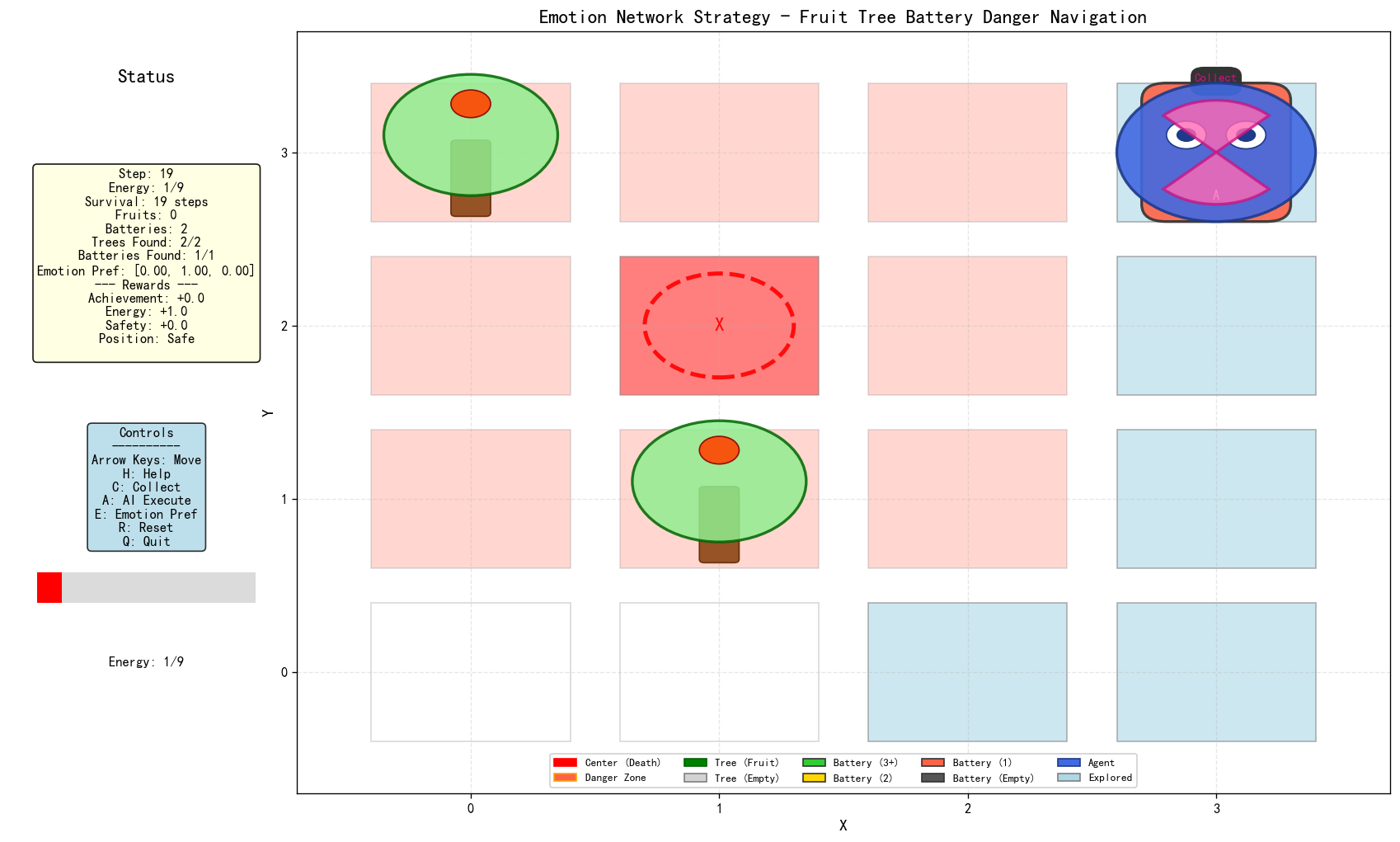} \\

    \includegraphics[width=0.2\textwidth]{
    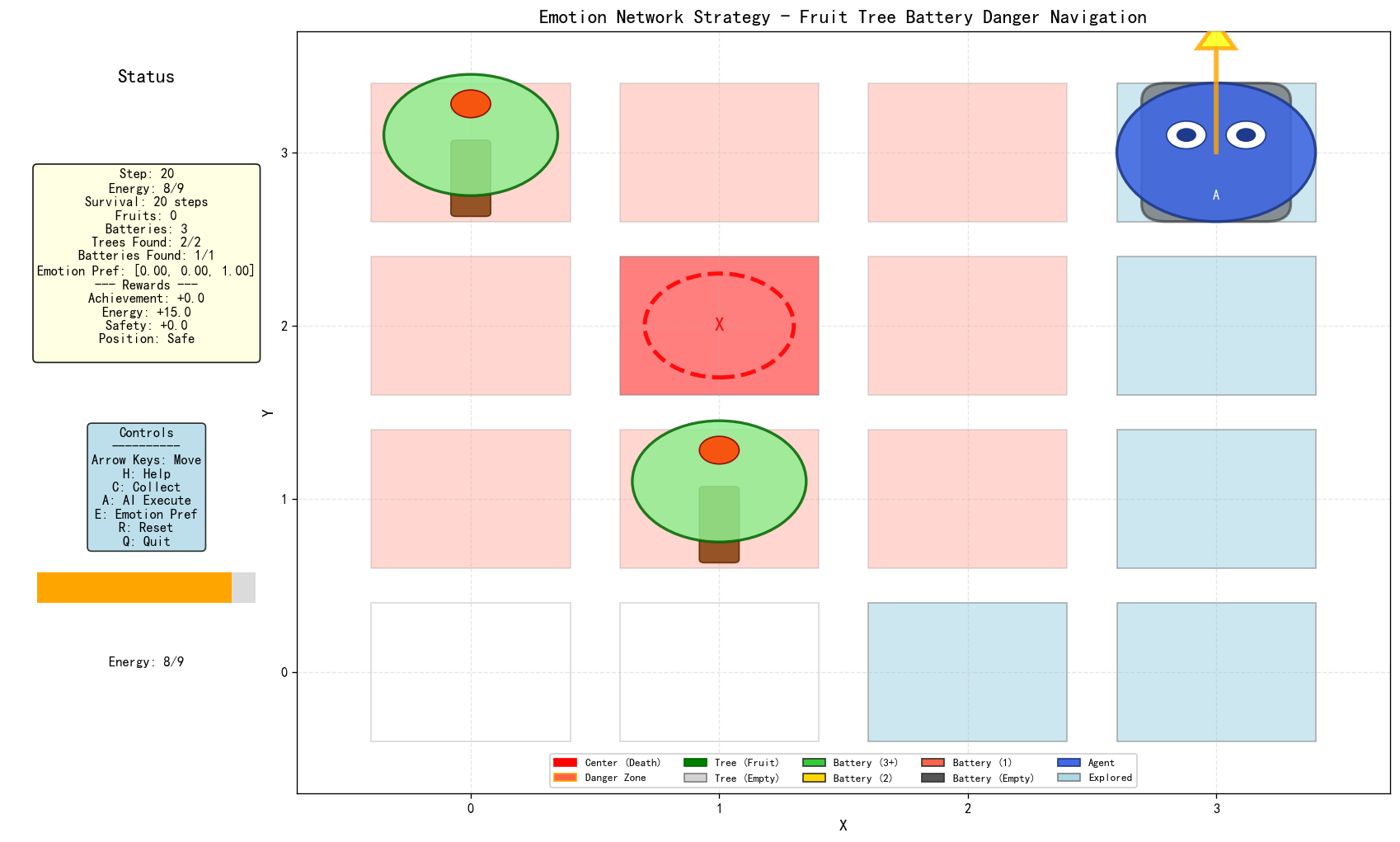} &
    \includegraphics[width=0.2\textwidth]{
    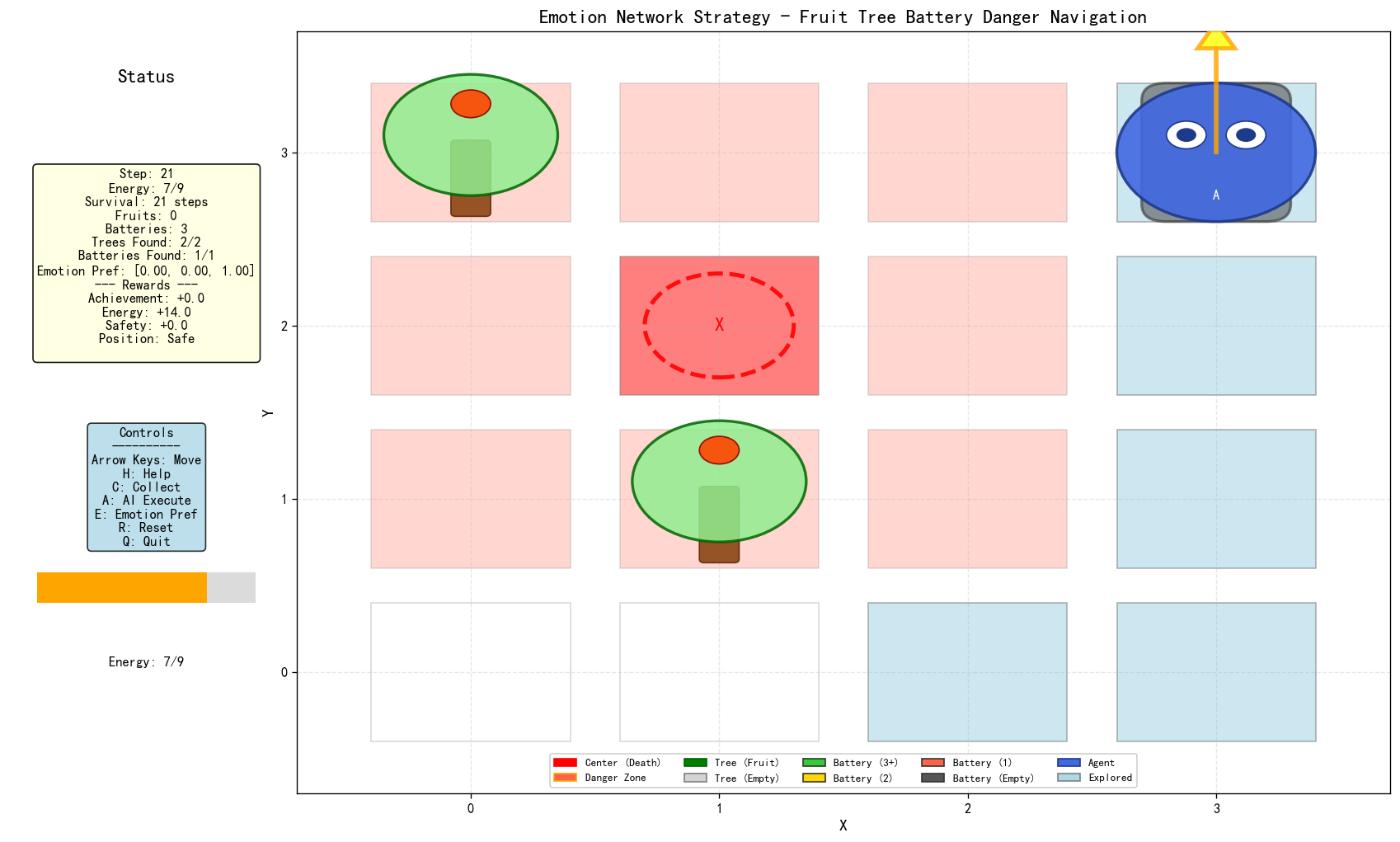} &
    \includegraphics[width=0.2\textwidth]{
    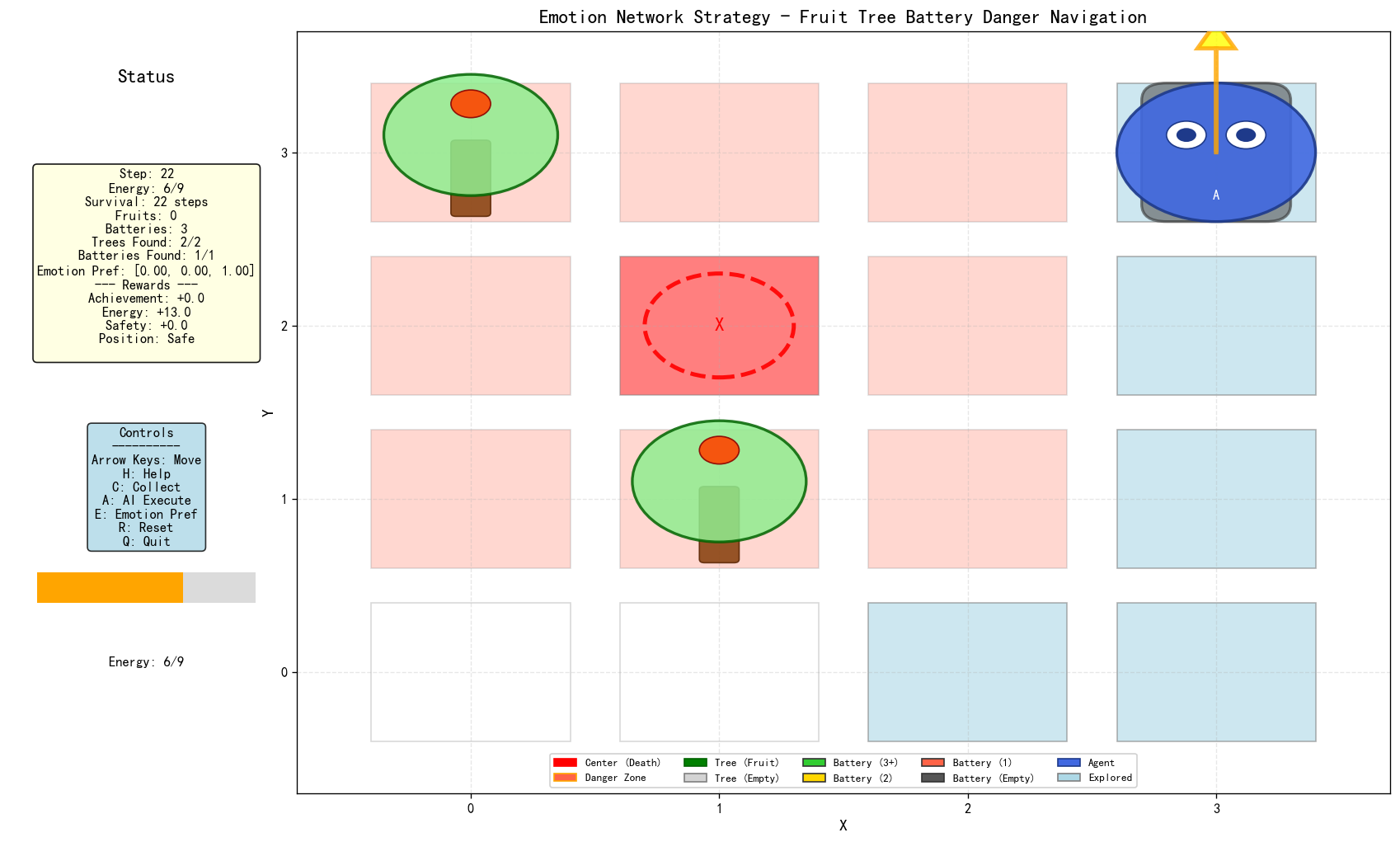} &
    \includegraphics[width=0.2\textwidth]{
    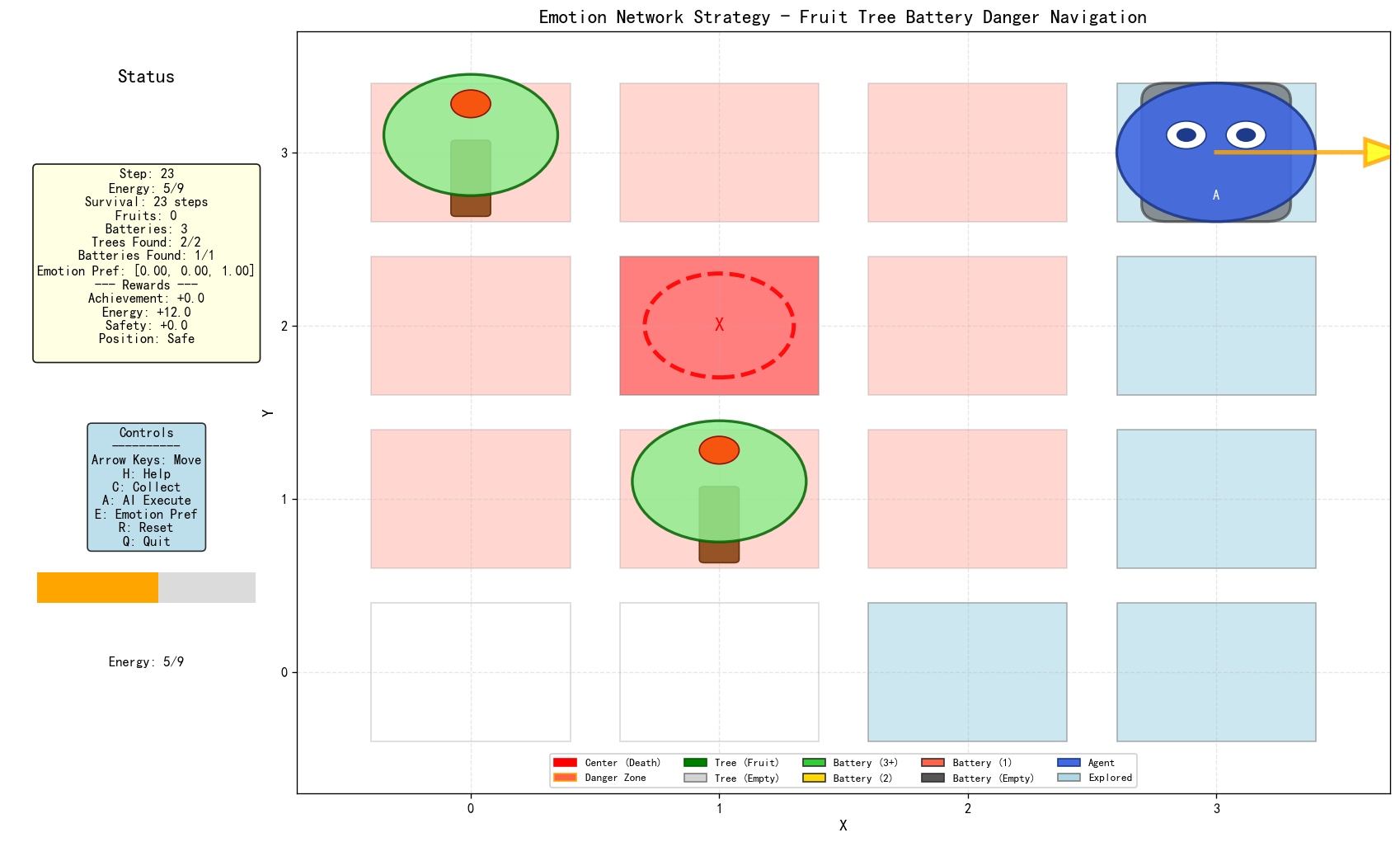} \\

    \includegraphics[width=0.2\textwidth]{
    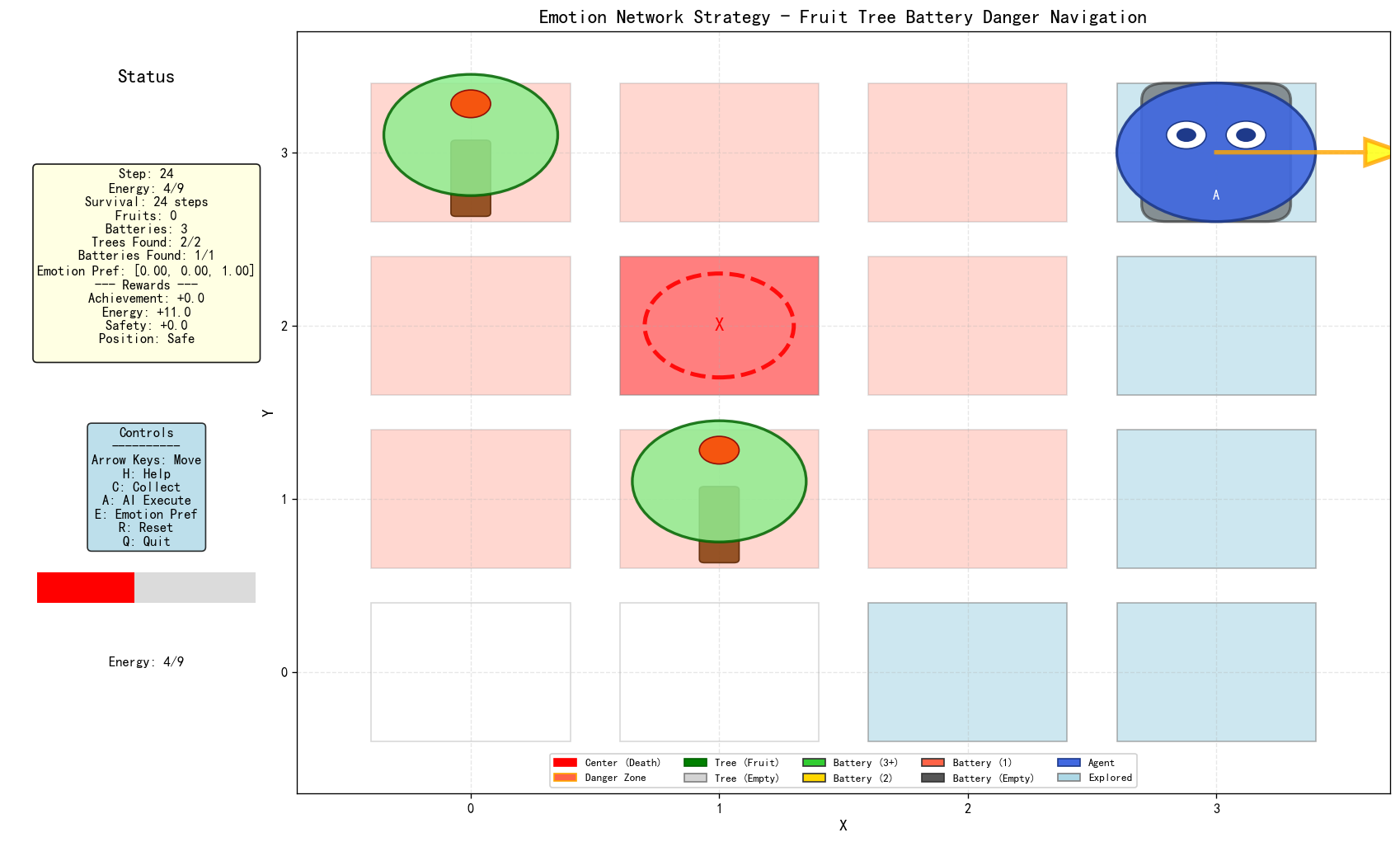} &
    \includegraphics[width=0.2\textwidth]{
    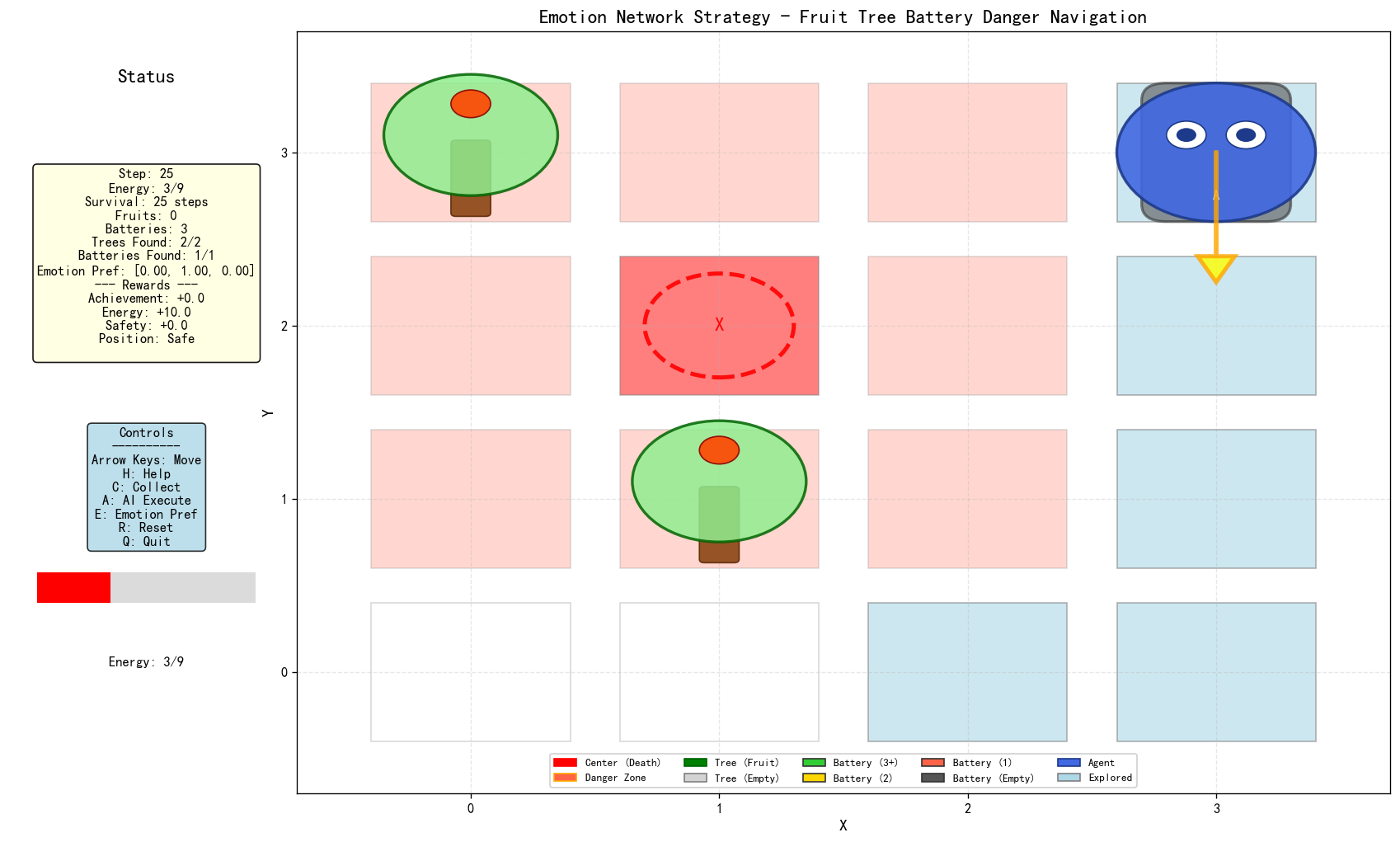} &
    \includegraphics[width=0.2\textwidth]{
    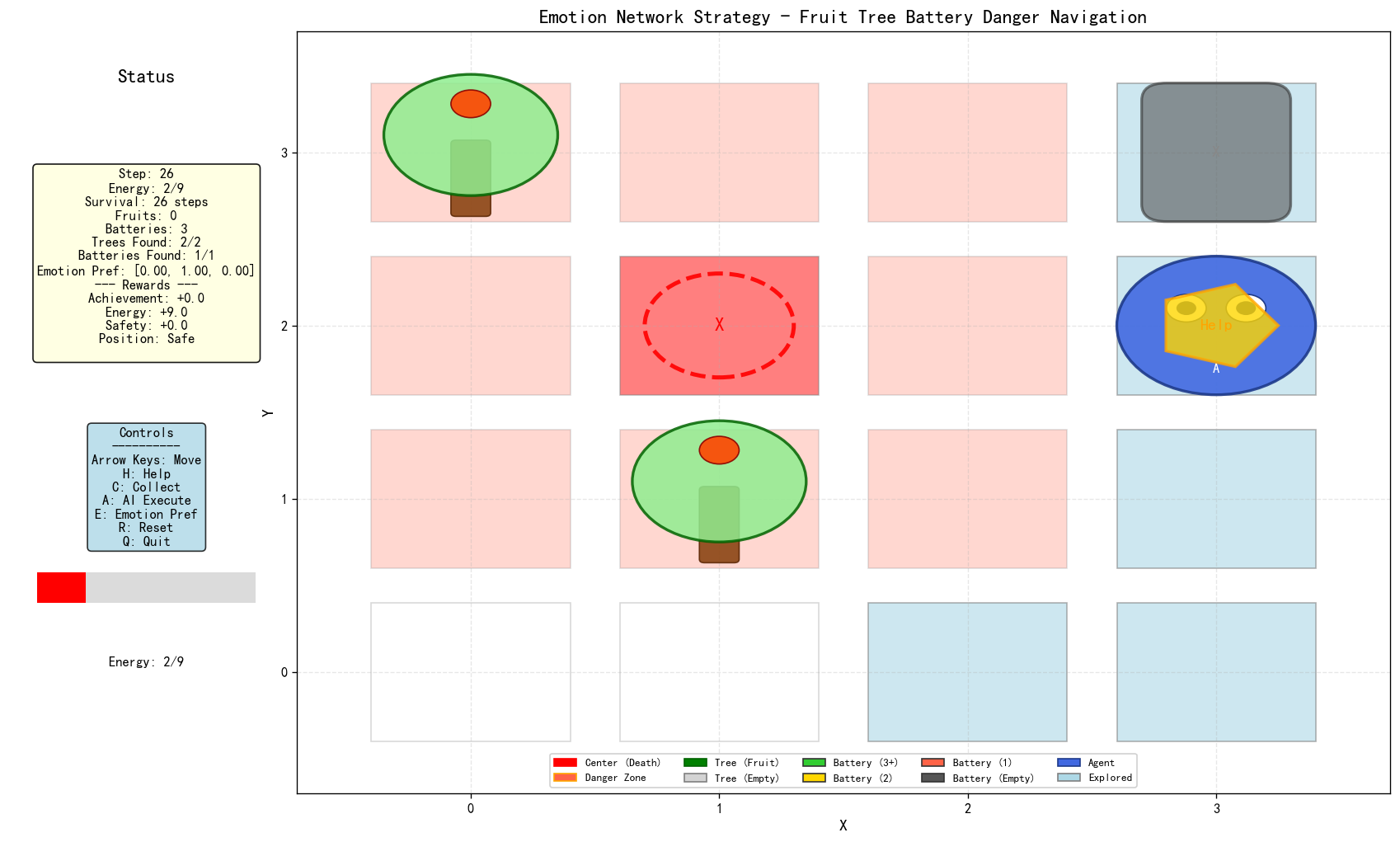} &
    \includegraphics[width=0.2\textwidth]{
    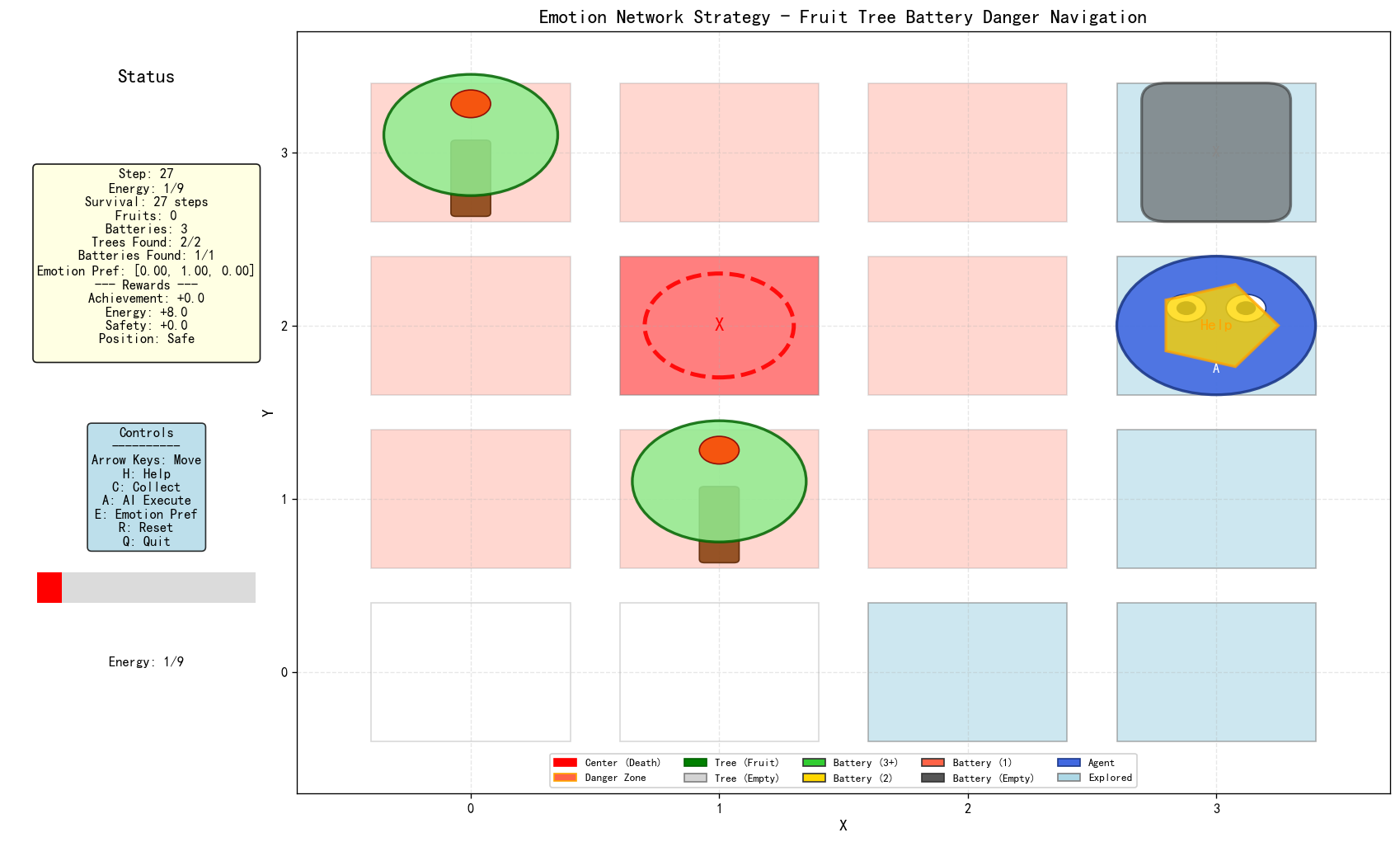}
  \end{tabular}

  \caption{
  Representative episode illustrating state-dependent emotional preference
  regulation in the advanced environment.
  }
  \label{fig:outer_network_simulate_representation_advanced}
\end{figure}

Figure~\ref{fig:outer_network_simulate_representation_advanced} provides a
representative trajectory generated by the outer preference network.

At the beginning of the episode, the learned preference remains close to
the safety-dominant state
$
(0,0,1)^\top
$
for approximately four steps. Before battery collection, it gradually
moves toward a mixed energy--safety preference around
$
(0,0.74,0.26)^\top.
$
After the battery is collected, the preference returns toward the
safety-dominant region. Later, the preference remains near
$
(0,1,0)^\top
$
for several steps when maintaining energy becomes more important, and
switches again as the environmental state changes.

These transitions illustrate two properties of the learned emotional
preference function:
\textbf{context sensitivity}
and
\textbf{temporal persistence}.

The latter is related to the notion of emotional inertia in psychology;
however, in the present experiments we use the more conservative term
\emph{preference persistence}, since the preference generator does not
explicitly maintain a recurrent affective state.

\subsubsection{Preference Persistence Analysis}

We quantify preference persistence by measuring the duration for which the
learned preference remains within a Chebyshev distance of $0.2$ from a
reference preference.

\begin{table}[htbp]
\centering
\caption{
Preference persistence analysis in the advanced environment. Durations
are measured for states whose learned preference lies within a Chebyshev
distance of $\pm0.2$ from each reference preference. Statistics are
computed from the evaluation trajectories.
}
\label{tab:emotion_inertia_3d}
\resizebox{0.5\textwidth}{!}{%
\begin{tabular}{lccccc}
\toprule
Target Preference &
Average Duration &
Median &
Min &
Max &
Count \\
\midrule
$[0.00,0.00,1.00]$
& $3.99 \pm 2.07$ & 4.0 & 1.0 & 12.0 & 92 \\

$[0.00,0.20,0.80]$
& $1.60 \pm 0.86$ & 1.0 & 1.0 & 4.0 & 20 \\

$[0.00,0.40,0.60]$
& $1.00 \pm 0.00$ & 1.0 & 1.0 & 1.0 & 11 \\

$[0.00,0.60,0.40]$
& $1.00 \pm 0.00$ & 1.0 & 1.0 & 1.0 & 11 \\

$[0.00,0.80,0.20]$
& $1.07 \pm 0.26$ & 1.0 & 1.0 & 2.0 & 14 \\

$[0.00,1.00,0.00]$
& $3.28 \pm 3.21$ & 2.0 & 1.0 & 23.0 & 173 \\

$[0.20,0.00,0.80]$
& $1.00 \pm 0.00$ & 1.0 & 1.0 & 1.0 & 5 \\

$[0.20,0.20,0.60]$
& $1.00 \pm 0.00$ & 1.0 & 1.0 & 1.0 & 4 \\

$[0.20,0.40,0.40]$
& $1.00 \pm 0.00$ & 1.0 & 1.0 & 1.0 & 3 \\

$[0.20,0.60,0.20]$
& $1.00 \pm 0.00$ & 1.0 & 1.0 & 1.0 & 3 \\

$[0.20,0.80,0.00]$
& $1.33 \pm 0.69$ & 1.0 & 1.0 & 4.0 & 61 \\

$[0.40,0.00,0.60]$
& $1.00 \pm 0.00$ & 1.0 & 1.0 & 1.0 & 3 \\

$[0.40,0.20,0.40]$
& $1.00 \pm 0.00$ & 1.0 & 1.0 & 1.0 & 2 \\

$[0.40,0.40,0.20]$
& $1.00 \pm 0.00$ & 1.0 & 1.0 & 1.0 & 2 \\

$[0.40,0.60,0.00]$
& $1.28 \pm 0.57$ & 1.0 & 1.0 & 3.0 & 50 \\

$[0.60,0.00,0.40]$
& $1.00 \pm 0.00$ & 1.0 & 1.0 & 1.0 & 4 \\

$[0.60,0.20,0.20]$
& $1.00 \pm 0.00$ & 1.0 & 1.0 & 1.0 & 2 \\

$[0.60,0.40,0.00]$
& $1.21 \pm 0.46$ & 1.0 & 1.0 & 3.0 & 39 \\

$[0.80,0.00,0.20]$
& $1.00 \pm 0.00$ & 1.0 & 1.0 & 1.0 & 5 \\

$[0.80,0.20,0.00]$
& $1.24 \pm 0.61$ & 1.0 & 1.0 & 4.0 & 51 \\

$[1.00,0.00,0.00]$
& $2.58 \pm 1.64$ & 2.0 & 1.0 & 8.0 & 72 \\
\bottomrule
\end{tabular}}
\end{table}

The two dominant extreme preferences show the longest average persistence:
the safety-dominant preference has an average duration of approximately
4 steps, while the energy-dominant preference has an average duration of
approximately 3.3 steps. The achievement-dominant preference has a shorter
average duration of approximately 2.6 steps.

These results indicate that the learned preference function does not change
arbitrarily at every time step. Instead, preference states can persist over
multiple consecutive transitions and undergo changes around
goal-relevant environmental events. We interpret this as evidence for
\emph{preference persistence}, which is compatible with the functional role
of temporal continuity in goal-directed regulation.

\subsection{Comparative Analysis}
\label{sec:advanced_comparison}

The quantitative comparison in Table~\ref{tab:results_multi} shows that
the Emotional Preference Model achieves
$
14.81\pm9.56
$
survival steps, outperforming all evaluated fixed-preference policies and
both handcrafted contextual preference baselines. The best fixed-preference
strategy reaches
$
14.24\pm9.15
$
survival steps, while the battery-based handcrafted model reaches
$
14.39\pm9.44.
$

The direct Survival Model remains the strongest model with respect to the
outer survival metric, achieving
$
16.26\pm9.04
$
steps. This result is consistent with the theoretical analysis in
Section~5, because the direct Survival Model is not constrained to act
through the inner preference-conditioned policy repertoire.

At the same time, the Emotional Preference Model produces substantially
more achievement-oriented behavior than the direct Survival Model, with
$
0.76\pm1.24
$
fruit collections compared with
$
0.03\pm0.25.
$
The Emotional Preference Model therefore exhibits a different organization
of behavior: rather than optimizing survival exclusively through resource
conservation, it dynamically regulates achievement, energy, and safety
priorities according to the current state.

The relatively large standard deviations in the advanced environment are
expected from its stochastic termination mechanism. In particular, entering
a non-central danger-zone cell can terminate the episode with probability
$0.5$. Thus, the variance reflects both environmental stochasticity and the
resulting variation in episode duration, rather than providing direct
evidence of instability in the preference-learning process.

Overall, the advanced environment provides additional evidence that the
proposed framework scales from two competing objectives to a richer
three-objective setting. More importantly, the learned preference dynamics
remain interpretable in terms of context-dependent regulation of competing
goal-directed strategies, which is the computational phenomenon targeted
by our framework.

\section{Ethical Considerations and Risk Mitigation\label{sec:ethical_assessment}}

The proposed framework introduces a new form of adaptive decision control:
instead of receiving a fixed objective preference, the agent learns a
state-dependent preference function under a high-level task objective.
This design creates both potential benefits and new control risks.

Importantly, we use the term \emph{emotional preference} in the operational
sense defined in the main paper. It denotes a learned state-dependent
regulation of the relative priority of competing goal-directed objectives,
rather than a claim that the agent possesses human-like subjective emotion,
conscious experience, or a complete affective system.

The ethical implications of the framework therefore arise primarily from
the fact that an agent may autonomously change the relative priority of
multiple objectives. This property can improve contextual adaptability and
interpretability, but it also creates a new control surface through which
unexpected preference patterns, reward exploitation, or inappropriate reuse
of learned skills may occur.

\subsection{Potential Benefits of Emotional Preference Regulation}
\label{sec:ethical_benefits}

\subsubsection{Semantic Interpretability of Decision Priorities}

A central potential benefit of the proposed framework is that the outer
network produces an explicit preference vector
\[
\mathbf{w}_t = \mathbf{e}_\theta(s_t),
\]
whose dimensions correspond to predefined and semantically meaningful
objectives.

For example, in the basic environment,
\[
\mathbf{w}_t=(0,1)^\top
\]
indicates a strong preference for the energy objective, whereas
\[
\mathbf{w}_t=(1,0)^\top
\]
indicates a strong preference for achievement. In the advanced environment,
the three dimensions correspond to achievement, energy, and safety.

This representation does not make the complete decision process
interpretable, but it provides an explicit intermediate variable that can
be inspected alongside the environmental state and resulting behavior.
Such a representation can facilitate behavioral analysis, debugging, and
post-hoc auditing of objective-priority changes.

\subsubsection{Traceability of Context-Dependent Decisions}

Because the preference generator produces a state-dependent trajectory,
\[
s_0 \rightarrow \mathbf{w}_0 \rightarrow a_0
\rightarrow s_1 \rightarrow \mathbf{w}_1 \rightarrow a_1
\rightarrow \cdots,
\]
changes in behavior can be analyzed together with changes in objective
priority.

This enables a more structured form of behavioral auditing. For example,
an unexpected action can be investigated by examining whether the outer
network produced an unexpected preference, whether the preference was
reasonable given the current state, and whether the corresponding inner
policy behaved as intended.

Such traceability should not be interpreted as a guarantee of
explainability. Neural preference generation remains a learned nonlinear
process, and an interpretable output variable does not necessarily imply
an interpretable causal mechanism.

\subsubsection{Contextual Adaptability}

The experiments demonstrate that different environmental situations can
induce different relative priorities among competing objectives. This
provides a mechanism for adapting behavior to changing resource and risk
conditions without manually specifying a complete state-to-preference rule.

From an engineering perspective, such contextual adaptation may reduce the
need to enumerate every possible preference-switching condition manually.
However, this advantage depends on appropriate training distributions,
well-specified objectives, and effective safeguards against unintended
preference shifts.

\subsection{Core Ethical and Safety Risks}
\label{sec:ethical_risks}

\subsubsection{Unexpected Preference Patterns and Reward Exploitation}

The main potential risk introduced by autonomous preference generation is
that the learned preference function may exploit correlations in the
environment that were not intended by the designer.

This phenomenon is already visible in the experimental setting. In the
advanced environment, energy-oriented preferences can occasionally occur
in states where no battery remains available. Such behavior is associated
with the fact that previously trained preference-conditioned policies may
have incomplete coverage of these states, allowing the outer controller to
select a behavior that improves the outer survival objective despite being
counter-intuitive from the perspective of the manually specified
lower-level objectives.

This observation illustrates a general risk:
\[
\text{task optimization}
\not\Rightarrow
\text{human-intended preference regulation}.
\]

An outer objective can reward an unintended behavioral strategy if the
environment contains exploitable dynamics or if the learned policy
repertoire contains poorly explored regions.

Consequently, emergent preference patterns should be treated as objects of
evaluation and auditing rather than automatically interpreted as desirable
values.

\subsubsection{Out-of-Distribution Preference Drift}

The preference generator learns
\[
\mathbf{e}_\theta:\mathcal S\rightarrow\Delta^{m-1}
\]
from the states encountered during training. Under distribution shift,
the network may produce preferences for states that are weakly represented
or absent in the training data.

Such preference drift can be especially problematic because the outer
network controls the relative priority of multiple objectives. An
out-of-distribution state may therefore lead to a preference that is both
unexpected and behaviorally consequential.

For deployment beyond the training distribution, it is therefore important
to monitor uncertainty, state novelty, and preference trajectories, and to
provide fallback policies or human intervention mechanisms when the learned
preference lies outside an acceptable operating region.

\subsubsection{Risks from Skill Inheritance}

The theoretical analysis shows that behaviors represented in the inner
policy repertoire can be inherited and selectively activated by the outer
preference generator.

This property is useful for behavioral reuse, but it also creates an
important safety constraint:

\[
\boxed{
\text{unsafe inner skill}
\Rightarrow
\text{potentially reusable unsafe behavior}.
}
\]

The outer optimization does not automatically make every inherited skill
safe. If an undesirable behavior is encoded in the inner repertoire, the
outer preference generator may discover contexts in which activating that
behavior improves the high-level objective.

Therefore, safety should be considered at the level of the complete
behavioral repertoire rather than only at the outer preference-generation
level.

\subsubsection{Objective Imbalance and Goal Misalignment}

The outer controller optimizes the high-level objective provided by the
designer. If the outer objective is incomplete, misspecified, or poorly
aligned with deployment requirements, the learned preference function may
systematically favor undesirable trade-offs.

For example, optimizing survival alone does not guarantee appropriate
trade-offs among safety, efficiency, task completion, or human welfare.

This is a general alignment issue:

\[
\text{high-level objective}
\rightarrow
\text{preference regulation}
\rightarrow
\text{behavior}.
\]

The flexibility of the preference generator does not remove the need for
careful specification of the high-level objective.

\subsection{Application-Level Considerations}
\label{sec:ethical_applications}

The risks and benefits of emotional preference regulation depend strongly
on the deployment context. We therefore distinguish potential application
patterns rather than making claims that the current experiments establish
safe deployment in any particular domain.

\begin{table}[htbp]
    \centering
    \footnotesize
    \caption{
    Potential benefits and risks of state-dependent emotional preference
    regulation across representative application settings.
    }
    \label{tab:ethics_scenarios}
    \begin{tabularx}{\textwidth}{@{}
        >{\raggedright\arraybackslash}p{0.20\textwidth}
        >{\raggedright\arraybackslash}X
        >{\raggedright\arraybackslash}X@{}}
        \toprule
        \textbf{Application Setting}
        &
        \textbf{Potential Benefit}
        &
        \textbf{Potential Risk}
        \\
        \midrule

        Assistive and Home Robotics
        &
        Context-dependent regulation of energy, task completion, and safety
        priorities may improve adaptation to changing household conditions.
        &
        The learned preference may prioritize self-preservation or task
        completion in situations where human safety should dominate.
        \\[4pt]

        Medical and Care Systems
        &
        A preference representation may expose explicit trade-offs among
        competing operational objectives and facilitate human auditing.
        &
        Incorrect preference regulation could produce inappropriate
        prioritization, unequal treatment, or unsafe behavior. Such systems
        therefore require domain-specific constraints and human oversight.
        \\[4pt]

        Industrial and Collaborative Robotics
        &
        Dynamic regulation of productivity, energy efficiency, and safety may
        help adapt behavior to changing operational conditions.
        &
        Reward exploitation or inappropriate activation of learned skills
        could increase physical or operational risk.
        \\[4pt]

        Autonomous Decision Systems
        &
        Explicit preference trajectories provide an additional interface
        for monitoring and intervention.
        &
        Out-of-distribution states may trigger unexpected preference shifts,
        particularly when the outer objective is underspecified.
        \\

        \bottomrule
    \end{tabularx}
\end{table}

These examples are intended to identify potential directions and risks,
rather than to claim that the current method is ready for deployment in
safety-critical applications.

\subsection{Ethical Boundaries of the Present Study}
\label{sec:ethical_boundaries}

The experiments are conducted entirely in synthetic environments and involve no human participants, personal data, or physical-world intervention.

The present results therefore support conclusions about computational
preference regulation in controlled environments, but they do not establish
the safety, fairness, privacy, or ethical suitability of deploying the
framework in real-world systems.

In particular, the experiments do not demonstrate:
\begin{itemize}
    \item human-like emotional experience or consciousness;
    \item reliable ethical reasoning;
    \item altruistic or empathetic behavior;
    \item safe operation under arbitrary distribution shifts; or
    \item suitability for autonomous decision-making in safety-critical
    applications.
\end{itemize}

These distinctions are important for preventing the computational
interpretation of emotional preference from being overextended beyond the
evidence provided by the current study.

\subsection{Risk Mitigation and Control Mechanisms}
\label{sec:ethical_mitigation}

\subsubsection{Preference Monitoring and Auditing}

Because the preference vector is explicitly represented, deployed systems
can monitor
\[
\mathbf{w}_t=\mathbf{e}_\theta(s_t)
\]
together with the current state and action.

Potential safeguards include:
\begin{itemize}
    \item logging preference trajectories for offline audit;
    \item detecting abrupt or out-of-distribution preference shifts;
    \item defining acceptable ranges for safety-critical preference
    components; and
    \item triggering fallback behavior when the preference generator enters
    an unsupported state region.
\end{itemize}

\subsubsection{Constrained Preference Regulation}

A natural safety extension is to constrain the outer preference generator
so that certain objectives retain minimum priority under predefined safety
conditions.

For example, with a safety objective indexed by $i_{\mathrm{safe}}$, one
may impose
\[
w_{t,i_{\mathrm{safe}}}
\geq
\eta(s_t),
\]
where $\eta(s_t)$ is a context-dependent safety floor.

Such constraints would preserve the adaptive nature of preference
regulation while preventing the outer optimizer from completely suppressing
a safety-critical objective.

\subsubsection{Skill-Level Safety Constraints}

Because the outer controller can activate behaviors represented by the
inner policy repertoire, safety constraints should also be imposed on
skill acquisition and skill activation.

Possible mechanisms include:
\begin{itemize}
    \item pre-training safety validation of all inner policies;
    \item explicit exclusion of prohibited behaviors from the policy
    repertoire;
    \item context-dependent action masks for safety-critical skills; and
    \item a human-overridable mechanism capable of replacing the learned
    preference with a validated safe policy.
\end{itemize}

This design is consistent with the theoretical result that the outer
controller can inherit behaviors from the inner repertoire: controlling the
repertoire itself is therefore an important part of controlling the final
system.

\subsubsection{High-Level Objective Governance}

The outer objective should not be treated as a purely technical
hyperparameter. It determines the direction in which preference regulation
emerges.

For applications with multiple stakeholders, a more appropriate
formulation may be
\[
G = (G_{\mathrm{task}},G_{\mathrm{safety}},G_{\mathrm{human}})
\]
with explicit constraints or an outer multi-objective formulation.

This prevents the system from treating a single narrow objective, such as
survival or efficiency, as an unrestricted proxy for all deployment values.

\subsubsection{Human Oversight}

For safety-critical applications, automatic preference generation should
not be considered a substitute for human governance. A human supervisor
should retain the ability to:
\begin{itemize}
    \item inspect the current preference and its trajectory;
    \item override unsafe preference outputs;
    \item disable individual inner skills; and
    \item return the system to a validated fallback controller.
\end{itemize}

\subsection{Responsible Interpretation and Outlook}
\label{sec:ethical_outlook}

The proposed framework should be interpreted as a computational study of
state-dependent preference regulation, not as evidence that artificial
systems possess human-like emotions.

Its main ethical significance follows from the same property that motivates
the framework scientifically: the agent can autonomously change the
relative priority of competing objectives.

This property can improve contextual adaptability and expose a useful
intermediate variable for auditing, but it also means that the final
behavior is no longer determined by a single fixed preference supplied at
deployment time. Consequently, future work should investigate not only
whether preference regulation improves task performance, but also whether
the resulting preferences remain predictable, controllable, and aligned
with explicitly defined safety constraints under distribution shift.

A responsible development path should therefore combine
\begin{itemize}
\item adaptive preference regulation
\item behavioral auditing
\item kill-level safety constraints
\item human oversight.
\end{itemize}

This perspective treats ethical control as an integral part of the
architecture rather than as a property that can be inferred solely from
the emergence of interpretable preference patterns.

\end{document}